\documentclass{article}

\usepackage{neurips_2024}
\usepackage{enumitem}

\usepackage[utf8]{inputenc} 
\usepackage[T1]{fontenc}    
\usepackage{hyperref}       
\usepackage{url}            
\usepackage{booktabs}       
\usepackage{amsfonts}       
\usepackage{nicefrac}       
\usepackage{microtype}      
\usepackage{xcolor}         
\usepackage{amsmath}
\usepackage{multirow}
\usepackage{graphicx}
\usepackage{subcaption}
\usepackage{xspace}
\usepackage{amsmath,amsfonts,amssymb}
\usepackage{wrapfig} 

\title{\name: Layer Fused Decoding to Exploit External \\ Knowledge in Retrieval-Augmented Generation}
\newcommand{\name}{\textsc{LFD}\xspace}
\usepackage{bm}
\author{
 \textbf{Yang Sun\textsuperscript{1}\thanks{Work done during an internship at Tencent.}},
 \textbf{Zhiyong Xie\textsuperscript{1}},
 \textbf{Lixin Zou\textsuperscript{1}\thanks{Corresponding author.}},
 \textbf{Dan Luo\textsuperscript{2}},
 \textbf{Min Tang\textsuperscript{3}},
 \textbf{Xiangyu Zhao\textsuperscript{4}}, \\
 \textbf{Yunwei Zhao\textsuperscript{5}},
 \textbf{Xixun Lin\textsuperscript{6}},
 \textbf{Yanxiong Lu\textsuperscript{7}},
 \textbf{Chenliang Li\textsuperscript{1}},
\\
 \textsuperscript{1}Key Laboratory of Aerospace Information Security and Trusted Computing, Ministry of Education, \\ School of Cyber Science and 
Engineering, Wuhan University \\ 
 \textsuperscript{2}Lehigh University, \textsuperscript{3}Monash University , \textsuperscript{4}City University of Hong Kong,
 \textsuperscript{5}CNCERT/CC, \\ \textsuperscript{6}Institute of Information Engineering, Chinese Academy of Sciences \\
 \textsuperscript{7}Search Team, WeChat, Tencent Inc. \\
 \{sunyang419, xzyong, zoulixin, cllee\}@whu.edu.cn, \\ dal417@lehigh.edu, min.tang@monash.edu, xy.zhao@cityu.edu.hk, \\zhaoyw@cert.org.cn, linxixun@iie.ac.cn, alanlu@tencent.com
}

\makeatletter
\@submissionfalse
\makeatother

\begin{document}

\maketitle

\begin{abstract}
Retrieval-augmented generation~(RAG) incorporates external knowledge into large language models~(LLMs), improving their adaptability to downstream tasks and enabling information updates. 
Surprisingly, recent empirical evidence demonstrates that injecting noise into retrieved relevant documents paradoxically facilitates exploitation of external knowledge and improves generation quality. 
Although counterintuitive and challenging to apply in practice, this phenomenon enables granular control and rigorous analysis of how LLMs integrate external knowledge.  
Therefore, in this paper, we intervene on noise injection and establish a layer-specific functional demarcation within the LLM: shallow layers specialize in local context modeling, intermediate layers focus on integrating long-range external factual knowledge, and deeper layers primarily rely on parametric internal knowledge.
Building on this insight, we propose Layer Fused Decoding~(\name), a simple decoding strategy that directly combines representations from an intermediate layer with final-layer decoding outputs to fully exploit the external factual knowledge.
To identify the optimal intermediate layer, we introduce an internal knowledge score (IKS) criterion that selects the layer with the lowest IKS value in the latter half of layers.
Experimental results across multiple benchmarks demonstrate that \name helps RAG systems more effectively surface retrieved context knowledge with minimal cost. 

\end{abstract}

\section{Introductions}

Retrieval-Augmented Generation (RAG) empowers large language models (LLMs) by dynamically integrating external knowledge during inference, enabling precise adaptation to knowledge-intensive tasks and rapidly evolving domains~\cite{DBLP:conf/icml/BorgeaudMHCRM0L22,DBLP:conf/icml/GuuLTPC20,DBLP:conf/nips/LewisPPPKGKLYR020}.
As a cornerstone of context-aware generation, RAG has been widely deployed in real-world applications, including recommendation systems~\cite{recom-1, recom-2, recom-3} and search engines~\cite{search-engine-1, search-engine-2}.
The broad applicability has spurred extensive optimization efforts on dynamic knowledge integration, including reranking strategies~\cite{rankrag, G-rag} to prioritize relevance, adaptive retrieval mechanisms~\cite{selfrag, flare} to minimize redundancy, and graph-based architectures~\cite{g-retriever, graphrag, kgqa} to model inter-document semantic relationships.

Despite these advances, LLMs might underutilize accurate external contexts, disproportionately favoring internal parametric knowledge during generation~\cite{redepe, ragtruth}. 
This overreliance risks propagating outdated information or hallucinations, undermining the trustworthiness of RAG systems. 
Surprisingly, recent studies reveal a paradoxical phenomenon: \textbf{injecting noise---random documents or tokens---to retrieved contexts that already contain answer-relevant snippets can improve the generation accuracy}~\cite{DBLP:conf/sigir/CuconasuTSFCMTS24, DBLP:journals/corr/abs-2502-10634}. 
While this noise-injection approach is simple and effective, its underlying influence on LLM remains unclear.
Furthermore, long contexts containing noise documents create computational overhead.  
Therefore, it is important to design more principled strategies that can achieve similar benefits without incurring excessive cost. 

\begin{figure}[t]
    \centering
    \includegraphics[width=0.85\textwidth]{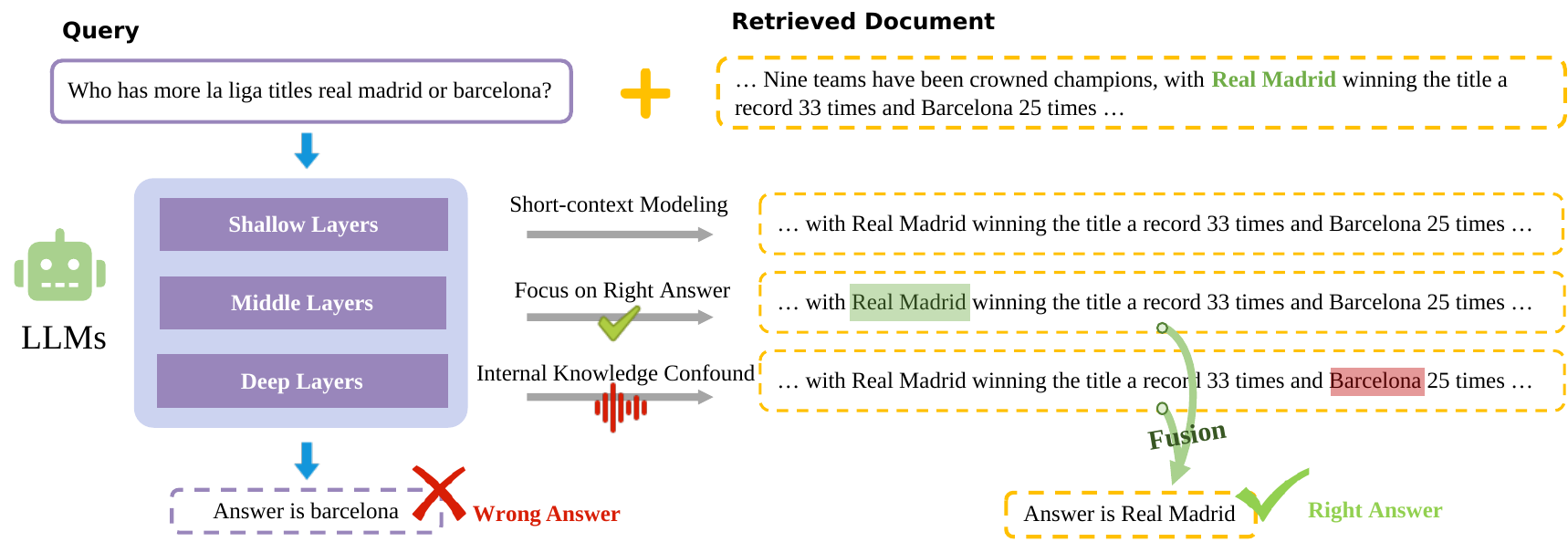}
    \caption{
    Illustration for layer-wise behavior in LLMs for RAG.
    Given a query and retrieved documents with the correct answer (``Real Madrid''), shallow layers capture local context, middle layers focus on answer-relevant content, while deep layers may over-rely on internal knowledge and hallucinate (e.g., ``Barcelona''). Our proposal, \name fuses middle-layer signals into the final output to preserve external knowledge and improve accuracy.}
    \label{fig:intro_example}
\end{figure}

This phenomenon enables more granular control and rigorous analysis of how LLMs integrate external knowledge. 
To investigate the underlying mechanisms, we study layer-wise external knowledge exploitation by measuring the divergence of ablating answer-determining context, i.e., the specific text segment within retrieved documents that directly supports the correct answer to a query. 
By intervening on injecting noise and measuring its impact on divergence patterns, we identify the relative importance of different layers in exploiting external knowledge. 
Empirically, we find that noise amplifies the contribution of answer-determining context in middle layers, highlighting their critical role in integrating long-range external information.
To further support this observation, we compare attention distributions across heads and layers with versus without answer-determining context. The analysis shows that attention differences peak in middle layers but decline in later layers, signaling a transition from external knowledge reliance to internal parametric knowledge utilization.
Following these observations, we propose a functional categorization of LLM layers: (1) shallow layers for short-context modeling, (2) intermediate layers for external knowledge integration, and (3) deeper layers for internal knowledge transformation. Therefore, when retrieved context already contains the correct answer, excessive dependence on internal knowledge in later layers introduces confounding effects, reducing generation accuracy, as visualized in Figure~\ref{fig:intro_example} (left).

Based on this insight, we propose a simple decoding method, Layer Fusing Decoding (\name), which enhances access to external factual knowledge without introducing additional noise overhead. 
The core idea is illustrated in the right panel of Figure~\ref{fig:intro_example}. 
\name fuses representations from the long-term context retrieval layer, where external knowledge is most effectively integrated, directly into the final decoding layer to maximize factual grounding.
To identify the appropriate layer for fusion, we track the model’s reliance on internal knowledge by measuring changes in hidden states across transformer feed-forward network (FFN) layers, where model knowledge is primarily stored~\cite{fnn-are-key-value-memories, knowledge-neurons}.
Specifically, we select the layer exhibiting minimal internal knowledge influence from the latter half of the model's layers. This criterion ensures that \name captures the externally grounded signal before it is overridden by parametric knowledge in later layers.
Importantly, \name operates at inference time, requiring no post-hoc fine-tuning or architectural modifications, making it easily integrable into existing LLM pipelines.
Finally, extensive empirical validation across diverse model architectures and datasets demonstrates that \name delivers competitive performance relative to noise-based approaches, while incurring significantly lower computational overhead.
\section{Preliminary}
\vspace{-5pt}
\subsection{Formulation of Retrieval-augmented Generation}
\vspace{-5pt}

In RAG systems, the generator $\mathcal{G}$ (typically a LLM) is expected to produce accurate and well-grounded responses based on retrieved documents $D=\{d_1, d_2, ..., d_\lambda\}$.
To quantify this capability, we define $A=\{a_1,...,a_\lambda\}$ as the set of key information extracted from $D$ that is necessary for generating an accurate answer.
The performance of the RAG system can be evaluated by measuring the inclusion rate of $A$ in its output response $r$, which reflects the model's ability to fully utilize valuable documents.
To produce the final response, the system first encodes the query $q$ and documents $D$ into a structured prompt through an instruction template $\mathcal{T}$, which instantiates the prompt $P=\mathcal{T}(q,D)$,
then the generator processes this prompt to produce the final response $r$.
To optimize the generation process, the generator $\mathcal{G}$ aims to ensure that all answers contained within $A$ are included in the generator’s output.
The accurate answer generated by $\mathcal{G}$ can be formalized as:
\begin{small}
\begin{align*}
    r=\mathcal{G}(q,D), \ \ \text{s.t.} \ \ \forall a_i \in A, \mathcal{I}(r, a_i)=\text{True},
\end{align*}
\end{small}
\noindent where $\mathcal{I}(r, a_i)=\text{True}$ means the answer $a_i$ is included in $r$.
\vspace{-5pt}
\subsection{External Knowledge Intervention in RAG}\label{causal}
\vspace{-5pt}

The counterintuitive effectiveness of noise in RAG systems motivates a deeper investigation into layer-wise  behavior in LLMs. 
To this end, we conduct an empirical study that contrasts layer-wise representation dynamics under two controlled interventions: (1) ablation of the answer-determining context, and (2) injection of varying levels of noise into the retrieved documents. 
This differential analysis reveals how noise modulates the model’s internal information flow, amplifying the influence of external knowledge in middle layers while mitigating the model’s tendency to over-rely on internal parametric memory. Our findings highlight key transformation layers where noise injection helps reduce context-dependent fragility, providing a foundation for our proposed decoding strategy.



\paragraph{Experimental Setup}


We simulate noisy level by adding \(k\) irrelevant Wikipedia documents \(N_k = \{n_1, \ldots, n_k\}\) to each prompt input~\cite{DBLP:conf/sigir/CuconasuTSFCMTS24}. 
To analyze how external knowledge flows in LLMs, we generate a modified document set \(\hat{D}\), which deletes key information \(A\) from the original documents \(D\). 
Without loss of generality, we assume the retrieval set \( D \) contains a single document, i.e., \( D = \{d_1\} \). 
By varying noise levels \(k\), we analyze how external knowledge impacts different layers of the model by comparing two prompts: the \textbf{original prompt} \(P^k = \mathcal{T}(q, D, N_k)\) (containing key information \(A\)) and the \textbf{modified prompt} \(\hat{P}^k = \mathcal{T} (q, \hat{D}, N_k)\), shown in Figure~\ref{fig:empirical-metric}. 
This section focuses on Llama2-7B and the NQ dataset. Additional results are in Appendix~\ref{appendix-b}.

\begin{wrapfigure}{l}{0.45\textwidth}
    \centering
    \includegraphics[width=\linewidth]{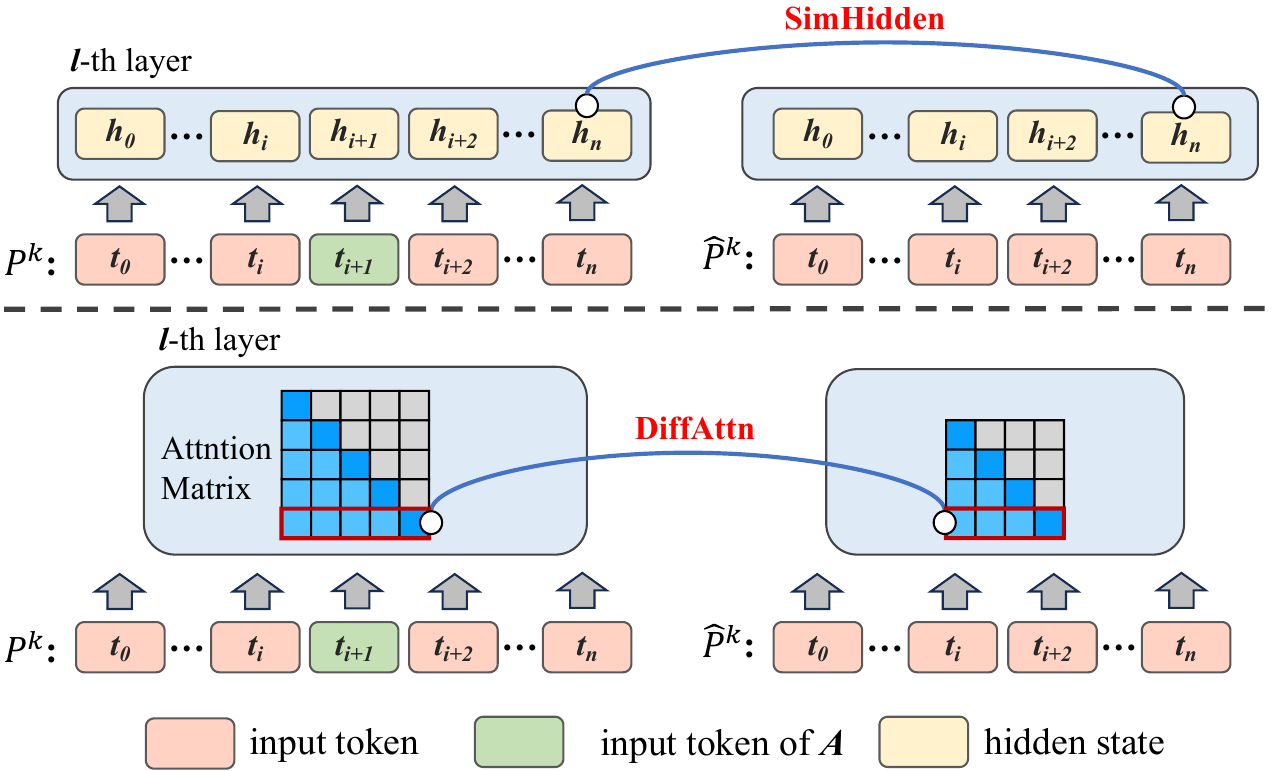}
    \vspace{-5pt}
    \caption{Calculation procedure for SimHidden and DiffAttn metrics.}
    \label{fig:empirical-metric}
    \vspace{-10pt}
\end{wrapfigure}
\paragraph{Quantify External Knowledge's Influence}
To measure how external knowledge influences LLMs, we compare intermediate representations before and after removing critical information \(A\).
For each layer $l \in \{1,~...,~L \}$ in the model, we calculate the cosine similarity between the feed-forward network (FFN) outputs of the original input \(P^k\) and its modified version \(\hat{P}^k\), focusing on the final prompt token as
\begin{small}
\[
\text{SimHidden}_l(P^k, \hat{P}^k) = \frac{\bm{h}_l(P^k) \cdot \bm{h}_l(\hat{P}^k)}{\|\bm{h}_l(P^k)\| \cdot \|\bm{h}_l(\hat{P}^k)\|},
\]
\end{small}
\noindent
where $\bm{h}_l(P)$ denotes the intermediate representation of
layer $l$ under prompt $P$.
This metric reveals how significantly removing \(A\) disrupts the model's contextual processing at each layer. 
A higher score indicates the model's understanding remains consistent even after removing \(A\), while a lower score suggests removing \(A\) plays a critical role in shaping the layer's output. 
By analyzing this metric across varying levels of noise injection (\(k\)), we can assess how different noise perturbation intensities affect the model's reliance on external knowledge, providing insights into how contextual information is integrated across layers.

Additionally, since the divergence of $\bm{h}_l(P^k)$ and $\bm{h}_l(\hat{P}^k)$ tends to accumulate in deeper layers, we further analyze attention patterns before and after the removal of the answer-determining context $A$. 
For each transformer layer $l$, we compute the average Jensen-Shannon Divergence~(JSD)~\cite{JSD} across all attention heads to quantify distributional shifts in attention: 
\begin{small}
\[
\text{DiffAttn}_l (P^k, \hat{P}^k) = \frac{1}{M}\sum_{m=1}^M \text{JSD}\left( \bm{\alpha}_{l,m}(P^k) \parallel \hat{\bm{\alpha}}_{l,m}(\hat{P}^k) \right),
\]  
\end{small}
\noindent
where 
$\bm{\alpha}_{l,m}(P^k)$ denotes the softmax-normalized attention distribution of head $m$ in layer $l$ at the final token position for prompt $P^k$. The variant $\hat{\bm{\alpha}}_{l,m}(\hat{P}^k)$ is computed by filling attention scores corresponding to answer-determining context $A$ with $-\infty$ before softmax normalization, thereby preserving the relative distribution over remaining tokens in modified prompt $\hat{P}^k$.
Larger divergence scores indicate that the external knowledge $A$ significantly influences the model's attention.  
Note that we concentrate on intervention of noise-free models (where $k=0$), we compare the original prompts $P^{0}$ with their intentionally altered counterparts $\hat{P}^{0}$.

\begin{figure}[t]
    \centering
    \begin{subfigure}[b]{0.42\textwidth}
    \vspace{-20pt}
        \includegraphics[width=\linewidth]{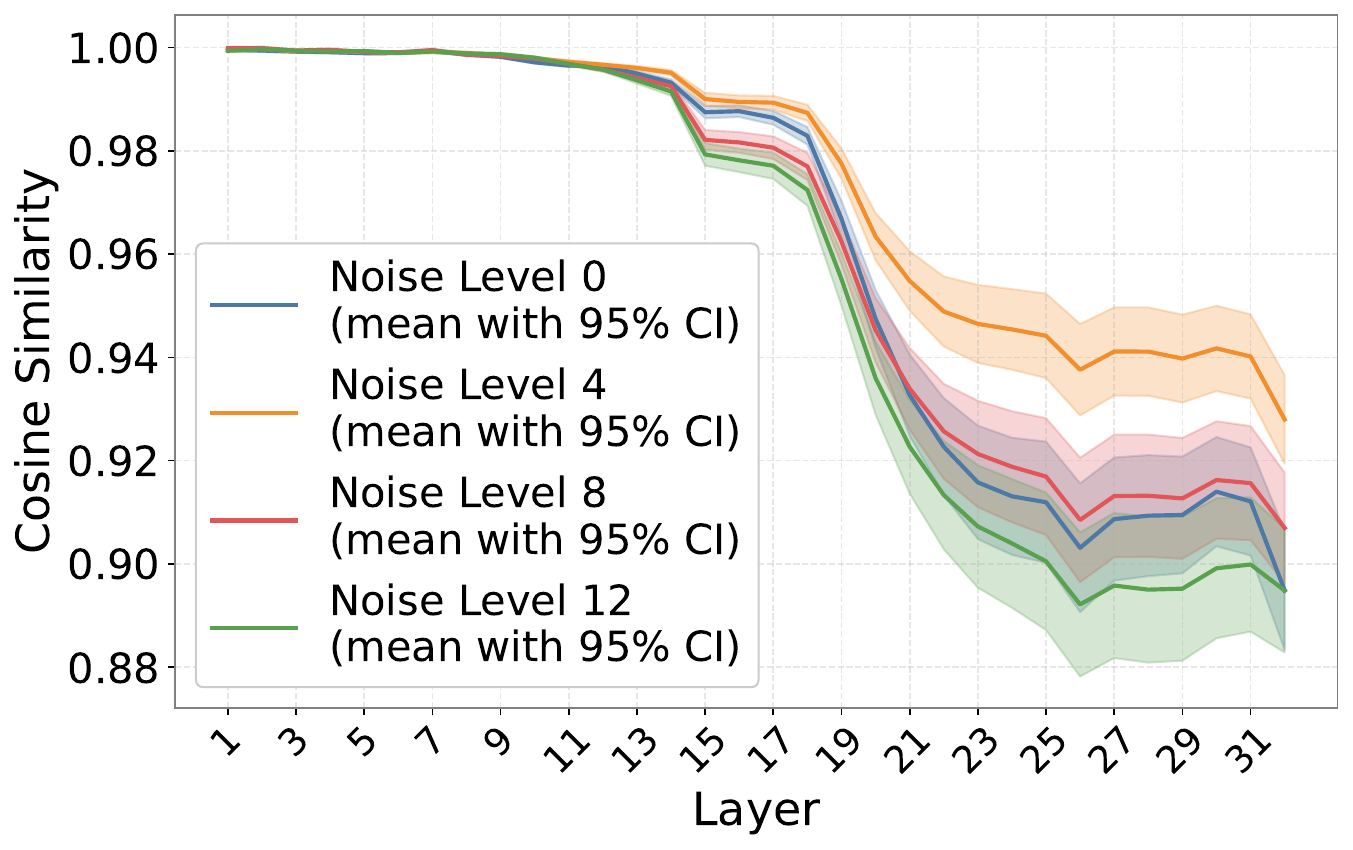}
        \caption{SimHidden (Smaller is better).}
    \end{subfigure}
    \hfill
    \begin{subfigure}[b]{0.42\textwidth}
    \vspace{-20pt}
        \includegraphics[width=\linewidth]{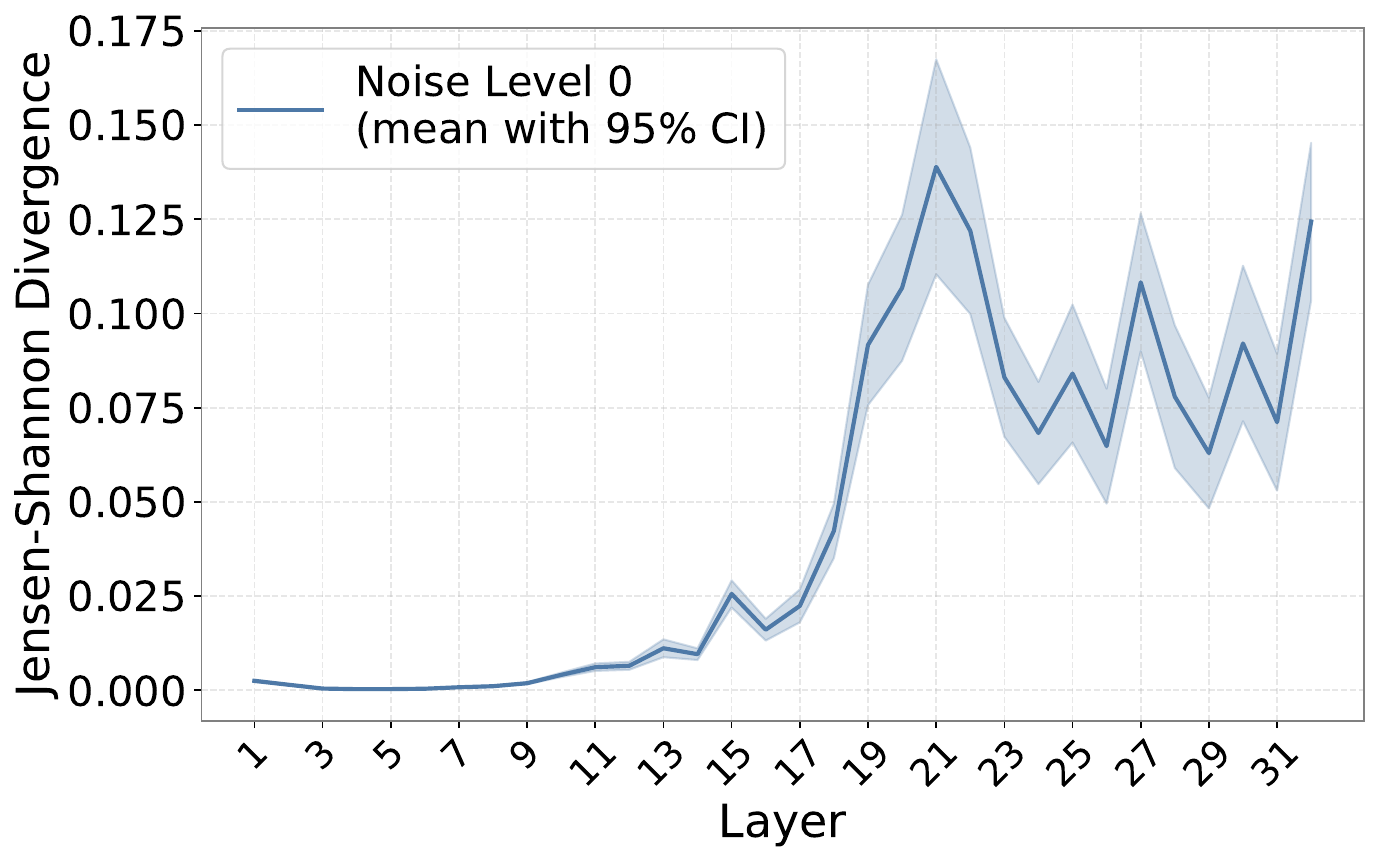}
        \caption{DiffAttn (Larger is better).}
    \end{subfigure}
    \vspace{-5pt}
    \caption{(a)~Average SimHidden scores (with 95\% confidence intervals) across layers under varying noise levels~(0, 4, 8, 12);~(b)~Average DiffAttn scores (with 95\% confidence intervals) across layers when noise level = 0.}
    \label{fig:empirical}
    \vspace{-15pt}
\end{figure}
\paragraph{Analysis and Conclusion} 

The quantified impacts of external knowledge, as depicted in Figure~\ref{fig:empirical}(a-b), lead to the following observations and conclusions: 
\textbf{(1)~The early layers~(1-14)~primarily perform short-context modeling}, which means they focus more on capturing local token relationships rather than integrating global contextual information.
This manifests through two key observations: the hidden state similarity remains relativly high across all noise conditions and the attention divergence also stay constantly low compared to other layers. 
\textbf{(2)~The middle layers~(15-26)~demonstrate long-term context retrieval capabilities}, as evidenced by two complementary patterns: a progressive decline in $\text{SimHidden}_l(P^k, \hat{P}^k)$~(Figure~\ref{fig:empirical}~(a)) and a corresponding increase in $\text{DiffAttn}_l$~(Figure~\ref{fig:empirical}~(b)).
This dual evidence indicates these layers' heightened sensitivity to the removal of $A$ and their capacity for comprehensive global context integration.
Meanwhile, an increasing noise level, especially when $k \geq 8$, leads to greater discrepancies in hidden state similarity, suggesting that a certain amount of noise can enhance the model's focus on $A$, thereby improving answer accuracy.
\textbf{(3)~The deeper layers~(27-32)~exhibit characteristics of parametric knowledge utilization.}
As we can observe, hidden state similarity does not continue to decrease as the model depth increases, instead, it shows a rebound, with SimHidden increasing by a maximum of 0.1.
Meanwhile, attention divergence, after peaking at layer 21, also exhibits a moderate decline, with DiffAttn decreasing by a maximum of approximately 0.06.
This indicates that the role of internal knowledge may be enhanced, as the model may focus more on processing the already captured contextual information rather than continuing to attend to external knowledge.
These observations motivate us to design our own methods to better leverage external knowledge with internal representations, thereby improving the model's performance in RAG systems.

\vspace{-5pt}
\section{Layer Fused Decoding}\label{method}
\vspace{-5pt}
\begin{figure}[t]
    \centering
    \includegraphics[width=0.95\textwidth]{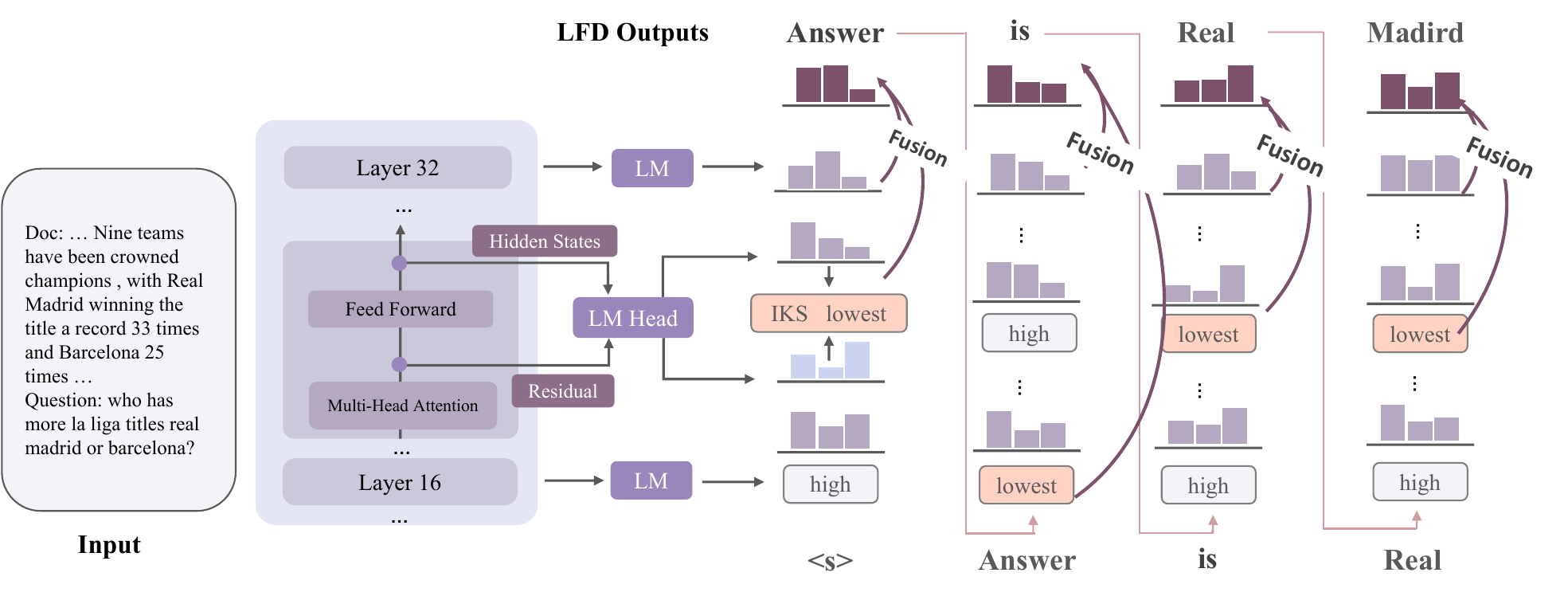}
    \caption{The proposed \name includes two key components: (1) A dynamic layer selection method using IKS to pinpoint the most impactful layers for integrating external retrieval knowledge. (2) A knowledge fusion mechanism that merges external information with the model's predictions after adaptive filtering to ensure alignment with the model's reasoning.}
    \label{fig:overview}
\end{figure}
\paragraph{Analysis and Conclusion}

This section present~\name, a framework designed to improve how external knowledge is integrated into model predictions while retaining accuracy. 
Our approach has two core components: 
(1) A \textbf{dynamic external knowledge layer identification strategy}, which automatically selects the most impactful layer for integrating retrieved context. This selection is guided by Internal Knowledge Scores~(IKS), which measure how strongly each layer reflects the model’s parametric knowledge.
(2) An \textbf{external knowledge fused decoding} mechanism, which merges external knowledge representations with the model’s final output. An adaptive filtering step precedes fusion to ensure the incorporated information complements the model’s reasoning.
Figure~\ref{fig:overview} illustrates the complete workflow of our approach.

\subsection{Dynamic External Knowledge Layer Identification}\label{sec:iks}
Effective integration of retrieval-based knowledge requires identifying layers that are responsive to external context while minimally influenced by internal parametric knowledge. To this end, we quantify the influence of internal knowledge across layers and select candidate layers based on the external information integration.

\paragraph{Internal Knowledge Score}
Recent advances in transformer interpretability reveal that FFN layers function as specialized knowledge repositories in LLMs~\cite{fnn-are-key-value-memories, knowledge-neurons}. To quantify how different layers utilize this internal knowledge, we design the Internal Knowledge Score (IKS), a metric that captures layer-specific knowledge transformations.
Given an input prompt $P$, let $\bm{h}^{\text{in}}_l(P) \in \mathbb{R}^d$ and $\bm{h}^{\text{out}}_l(P) \in \mathbb{R}^d$ denote the input and output activations of the $l$-th FFN layer. 
We project these vectors into the vocabulary space via LogitLens~\cite{DBLP:journals/corr/abs-2303-08112} of LLMs, parameterized by $W_{\text{LM}} \in \mathbb{R}^{d \times |V|}$, as follows:  
\[
\bm{p}^{\text{in}}_l \!=\! \text{softmax}(W_{\text{LM}} \bm{h}^{\text{in}}_l(\!P)), \quad 
\bm{p}^{\text{out}}_l \!=\! \text{softmax}(W_{\text{LM}} \bm{h}^{\text{out}}_l(\!P)).
\]  
The IKS for layer $l$ is defined as the JSD divergence between these distributions as  
\[
\text{IKS}_{l}(P) = \text{JSD}(\bm{p}^{\text{in}}_l \parallel \bm{p}^{\text{out}}_l).
\]  
This divergence quantifies the \textbf{parametric knowledge impact} of FFN layers, where higher IKS indicates greater transformation in the residual stream and stronger reliance on internal knowledge.

\paragraph{External Knowledge Layer Selection}
To identify the optimal layer for leveraging context-derived factual knowledge during inference, we propose two principled criteria for layer selection:
\textbf{(1)~Integration of Late-Stage Layers}: 
We integrate the latter half of the LLM's layers for final decoding. 
This choice is motivated by the observation that early layers predominantly focus on short-context modeling, with limited capacity to capture contextual dependencies. The integration of external knowledge becomes progressively stronger in middle-to-late layers (as analyzed in Section~\ref{causal}). 
\textbf{(2)~Lowest IKS Layer Selection}: Within the subset of late-stage layers, we select the layer exhibiting the lowest IKS score. This layer strikes a balance by retaining sufficient external contextual signals while substantially mitigating distortion of the model’s inherent knowledge representations.
By selecting these layers, we maximize the exploitation of retrieval context before the dominance of internal parametric knowledge obscures external signals. Empirical validation further confirms the efficacy of this strategy~(detailed in Section~\ref{ablation}).

\subsection{External Knowledge Fused Decoding}
We propose an intervention-aware fusion framework that dynamically integrates intermediate representations from layer \( i \) (identified via IKS scoring) with the final layer's predictions. To establish distributional coherence between these complementary knowledge sources, we first compute normalized log-probabilities through \(\log\text{softmax}\) transformation:
\begin{small}
\begin{align*}
\tilde{\bm{p}}^{\text{out}}_i \!=\! \log\!\text{softmax}(\!W_{\text{LM}}\bm{h}^{\text{out}}_i\!(\!P)\!),\quad 
\tilde{\bm{p}}^{\text{out}}_L \!=\! \log\!\text{softmax}(\!W_{\text{LM}}\bm{h}^{\text{final}}_L\!(\!P)\!)
\end{align*} 
\end{small}
where \( W_{\text{LM}}\bm{h}^{\text{out}}_i(P) \) and \( W_{\text{LM}}\bm{h}^{\text{out}}_L(P) \) denote the raw logits from the intervention layer and final layer respectively.
To mitigate noise amplification from early layer predictions while preserving critical external knowledge signals, we implement a dynamic gating mechanism:
\begin{small}
\begin{align*}
\tilde{\bm{f}}_i(t) \!=\! 
\begin{cases} 
\tilde{\bm{p}}^{\text{out}}_i(t) \!+\! \tilde{\bm{p}}^{\text{out}}_L(t),\! &\! \text{if } \tilde{\bm{p}}^{\text{out}}_i(t) \!\geq\! \min\{\!\tau \!\cdot\! \text{max}(\tilde{\bm{p}}^{\text{out}}_L), \text{max-s}(\tilde{\bm{p}}^{\text{out}}_L)\!\} \\
-\infty, & \text{otherwise},
\end{cases}
\end{align*}
\end{small}
where $\text{max}(\tilde{\bm{p}}^{\text{out}}_L)$ and $\text{max-s}(\tilde{\bm{p}}^{\text{out}}_L)$ represent the maximum and \( s \)-th maximum values in final output layer logits \( \tilde{\bm{p}}^{\text{out}}_L \), $\tau = 0.1$. 
The final decoding distribution, derived via normalized fusion \( \bm{f}_i = \text{softmax}(\tilde{\bm{f}}_i) \), enables synergistic knowledge transfer between layers, preserving the final layer’s discriminative capacity to balance the integration of external knowledge with the model’s inherent confidence.

\section{Experiments}
\vspace{-5pt}
\subsection{Setup}\label{sec:setup}
\vspace{-5pt}
\paragraph{Datasets} 
Following the experimental setups of~\cite{DBLP:conf/sigir/CuconasuTSFCMTS24, pandora-noise, adaptive-rag}, we evaluate our approach across following datasets:
\textbf{(1) Natural Questions~(NQ)}~\cite{nq-dataset}, a large-scale QA dataset based on real Google search queries. 
\textbf{(2) RGB}~\cite{rgb-dataset}, a RAG benchmark that evaluates models' ability to utilize retrieved information, focusing on noise robustness, negative rejection, information integration, and counterfactual robustness. 
We use its English test set for evaluation.
\textbf{(3) HotpotQA~(HQA)}~\cite{hotpotqa-dataset}, which requires multi-hop reasoning over multiple documents, featuring both compare and bridge question types: compare questions involve contrasting information from multiple sources, while bridge questions require connecting intermediate facts to reach the answer. We evaluate all methods on their dev set, reporting results under the categories Compare, Bridge, and Total in the main results table.
\textbf{(4) 2WikiMultihopQA~(2WQA)}~\cite{2wiki-dataset}, which presents a more challenging multi-hop QA scenario with four distinct task types: comparison (comparing information), bridge comparison (connecting intermediate facts for comparison), inference (deriving conclusions), and compositional (integrating multiple facts).
We use its dev set for evaluation, reporting separate performance for each question type (denoted as Comapre, Bridge, Inf, Compose and Total) in the experimental results.
\vspace{-10pt}
\paragraph{Baselines} To demonstrate the broad effectiveness, we evaluate it on four widely used language models:  Llama-2-7B-Chat-hf~(Llama2-7B)~\cite{llama2}, Mistral-7B-v0.1~(Mistral-7B)~\cite{mistral7b}, DeepSeek-llm-7B-base~(DeepSeek-7B)~\cite{deepseek}, and Qwen3-8B~\cite{qwen3}.
We evaluate \name against three decoding strategies:
\textbf{(1) Greedy Decoding~(GD)} is the standard autoregressive decoding method that selects the highest-probability token at each generation step.
To further examine the effect of noise, we augment the GD strategy with varying numbers of irrelevant documents~(4, 8, and 12) added to the prompt context. 
The abbreviations GD~(0), GD~(4), GD~(8) and GD~(12), as used in the main results table, refer to greedy decoding with 0, 4, 8, 12 noise documents added to the prompt context, respectively
\textbf{(2) Contrastive Search~(CS)}~\cite{contrastive_search} promotes more comprehensive outputs by balancing response quality and diversity.
\textbf{(3) Decoding by Contrasting Layers}~(DoLA)~\cite{dola} is a contrastive decoding approach that reduces hallucinations by comparing predictions across different model layers. 
\textbf{(4) \name~(Random)} is a variant that randomly selects an intermediate layer to fuse during the final decoding stage.



\vspace{-10pt}
\paragraph{Evaluation Metrics} We use accuracy as our primary evaluation metric. For the NQ and RGB datasets, which include samples with multiple acceptable answer variants (e.g., alternative phrasings of the same concept), we follow the evaluation protocol established in \cite{DBLP:conf/sigir/CuconasuTSFCMTS24, metric-eval-ref-1, metric-eval-ref-2}. A model's response is marked correct if it matches any annotated ground-truth answer.

\vspace{-10pt}
\paragraph{Experimental Setting}
Both \name and DoLA require the implementation of dynamic layer selection strategies.
For LFD, we prioritize layers in the latter half of the architecture (e.g., layers 16–32 for 32-layer models like Llama2-7B and Mistral-7B). 
In contrast, DoLA selects layers across the entire depth (layers 0–32) based on divergence from the final layer’s predictions. 
Similar configurations apply to DeepSeek-7B (30 layers: LFD uses 15–30; DoLA uses 0–30) and Qwen3-8B (36 layers: LFD uses 18–36; DoLA uses 0–36). 
Following DoLA’s convention, we restrict candidates to even-numbered layers within these ranges for efficiency. 
Contrastive Search uses a degeneration penalty $\alpha=0.6$ and top-k candidate size $k=5$, adopting the parameters from~\cite{contrastive_search}.
Since each sample in the aforementioned benchmark datasets is accompanied by multiple retrieved documents, we construct the input context using these documents, supplemented with golden documents, i.e., the ground-truth passages that contain the information necessary to answer the question.
This setup, following prior work~\cite{rag-nlp-tasks, DBLP:conf/sigir/CuconasuTSFCMTS24, frames, mirage}, allows us to evaluate the model’s ability to effectively leverage external knowledge when it is explicitly provided in the input context.


\vspace{-5pt}
\subsection{Main Results}
\vspace{-5pt}
\begin{table}[t]
    \caption{The accuracy performance comparison of different methods on four datasets. \textbf{Bold} values indicate the best performance, while \underline{underlined} values represent the second-best.}
    \centering
    \setlength{\tabcolsep}{4pt}
    \scriptsize
    \begin{tabular}{llcccccccccc}
        \toprule
        & & \multirow{2}{*}{NQ} & \multirow{2}{*}{RGB} & \multicolumn{3}{c}{HotpotQA} & \multicolumn{5}{c}{2WikiMultihopQA} \\
        \cmidrule(lr){5-7} \cmidrule(lr){8-12}
        & & & & Compare & Bridge & Total & Compare & Bridge & Inf & Compose & Total \\
        \hline \hline
        \multirow{8}{*}{Llama2-7B}
         & GD~(0) & 0.5745 & 0.8900 &  0.4755 &  0.4108 &  0.4237 &  0.3095 &  0.2815 &  0.2607 &  0.2928 &  0.2904 \\
         & GD~(4) & 0.5559 & 0.8267 & 0.5084 & 0.4049 & 0.4257 & 0.2559 & 0.2564 & 0.2048 & 0.2405 & 0.2433 \\
         & GD~(8) & \underline{0.5965} & 0.8300 & 0.5091 & 0.4096 & 0.4317 & \textbf{0.3257} & 0.2983 & \underline{0.2789} & 0.2995 & \underline{0.3030} \\
         & GD~(12) & \textbf{0.6383} & 0.8200 & \underline{0.5286} & 0.4236 & \underline{0.4447} & \underline{0.3204} & \textbf{0.3202} & 0.2646 & 0.2943 & 0.3027 \\
         & CS & 0.5775 & 0.8567 & 0.4284 & 0.4049 & 0.4096 & 0.2800 & 0.2462 & 0.2464 & 0.2866 & 0.2712 \\
         & DoLA & 0.5809 & 0.8733 & 0.4438 & 0.3883 & 0.3995 & 0.2738 & 0.2411 & 0.2289 & 0.2592 & 0.2550 \\
         & LFD~(Random) & 0.5928 & \underline{0.8900} & 0.4385 & \underline{0.4295} & 0.4313 & 0.3065 & 0.2845 & 0.2744 & \underline{0.3066} & 0.2978 \\
         & LFD & 0.5949 & \textbf{0.9067} & \textbf{0.5453} & \textbf{0.4295} & \textbf{0.4528} & 0.3174 & \underline{0.3005} & \textbf{0.3043} & \textbf{0.3232} & \textbf{0.3144} \\
        \midrule
        \multirow{8}{*}{Mistral-7B}
         & GD~(0) & 0.6130 & 0.8900 & 0.5864 & 0.5889 & 0.5884 & 0.5518 & 0.5204 & 0.5605 & 0.5337 & 0.5385 \\
         & GD~(4) & 0.5667 & 0.8033 & 0.5494 & 0.5348 & 0.5377 & 0.3681 & 0.3443 & 0.3173 & 0.3296 & 0.3406 \\
         & GD~(8) & 0.5678 & 0.7700 & 0.5326 & 0.5260 & 0.5273 & 0.2963 & 0.2874 & 0.2503 & 0.2582 & 0.2728 \\
         & GD~(12) & 0.5814 & 0.8267 & 0.5440 & 0.5461 & 0.5457 & 0.3098 & 0.2888 & 0.2503 & 0.2655 & 0.2795 \\
         & CS &  0.5598 & 0.7433 & 0.4149 & 0.5117 & 0.4922 & 0.4346 & 0.3873 & 0.4486 & 0.4740 & 0.4423 \\
         & DoLA & 0.6142 & 0.8900 & 0.5931 & 0.5870 & 0.5883 & 0.5366 & 0.5018 & 0.5572 & 0.5239 & 0.5262 \\
         & LFD~(Random) & \underline{0.6270} & \underline{0.9133} & \underline{0.5985} & \underline{0.6016} & \underline{0.6009} & \underline{0.5687} & \underline{0.5317} & \underline{0.5767} & \underline{0.5506} & \underline{0.5541} \\
         & LFD & \textbf{0.6357} & \textbf{0.9367} & \textbf{0.6026} & \textbf{0.6093} & \textbf{0.6079} & \textbf{0.5813} & \textbf{0.5434} & \textbf{0.5982} & \textbf{0.5645} & \textbf{0.5680} \\
        \midrule
        \multirow{8}{*}{DeepSeek-7B}
         & GD~(0) & 0.5250 & 0.8233 & 0.3282 & 0.4033 & 0.3883 & 0.2625 & 0.2414 & 0.2663 & 0.2609 & 0.2609 \\
         & GD~(4) & 0.5216 & 0.8300 & 0.3968 & 0.4437 & 0.4343 & 0.2807 & 0.2939 & 0.2744 & 0.2730 & 0.2796 \\
         & GD~(8) & 0.5226 & \underline{0.8800} & 0.4506 & \underline{0.4579} & \underline{0.4564} & 0.3578 & 0.3603 & 0.3407 & 0.3463 & 0.3515 \\
         & GD~(12) & 0.5204 & \textbf{0.8933} & \underline{0.4573} & \textbf{0.4618} & \textbf{0.4609} & 0.3701 & 0.3705 & 0.3349 & 0.3614 & 0.3622 \\
         & CS & 0.5322 & 0.8000 & 0.4028 & 0.4584 & 0.4472 & \underline{0.3929} & \underline{0.3833} & \underline{0.3888} & \underline{0.4077} & \underline{0.3964} \\
         & DoLA & 0.3755 & 0.5033 & 0.3490 & 0.3003 & 0.3100 & 0.3039 & 0.3118 & 0.2484 & 0.2594 & 0.2803 \\
         & LFD~(Random) & 0.4858 & 0.6700 & 0.3847 & 0.3734 & 0.3757 & 0.3466 & 0.3563 & 0.3277 & 0.3320 & 0.3403 \\
         & LFD & \textbf{0.5412} & 0.8267 & \textbf{0.4801} & 0.4466 & 0.4533 & \textbf{0.4492} & \textbf{0.4227} & \textbf{0.4194} & \textbf{0.4223} & \textbf{0.4285} \\
        \midrule
        \multirow{6}{*}{Qwen3-8B}
         & GD~(0) & 0.7318 & \underline{0.9571} & 0.6960 & 0.6708 & 0.6759 & \underline{0.6637} & \underline{0.6335} & 0.5897 & 0.6085 & 0.6250 \\
         & GD~(4) & 0.7213 & 0.9467 & 0.6658 & 0.6544 & 0.6567 & 0.6044 & 0.5729 & 0.4948 & 0.5268 & 0.5517 \\
         & GD~(8) & 0.7233 & 0.9500 & 0.6584 & 0.6568 & 0.6571 & 0.6117 & 0.5780 & 0.5078 & 0.5370 & 0.5605 \\
         & GD~(12) & 0.7133 & 0.9533 & 0.6530 & 0.6582 & 0.6571 & 0.6146 & 0.5722 & 0.5052 & 0.5462 & 0.5634 \\
         & CS & 0.7204 & 0.9533 & 0.7081 & 0.6762 & 0.6826 & 0.6468 & 0.6204 & 0.5754 & 0.6015 & 0.6134 \\
         & DoLA & 0.7168 & 0.9533 & 0.7014 & 0.6703 & 0.6766 & 0.6485 & 0.6058 & 0.5650 & 0.5929 & 0.6057 \\
         & LFD~(Random) & \underline{0.7357} & 0.9567 & \underline{0.7108} & \underline{0.6935} & \underline{0.6970} & 0.6627 & 0.6334 & \underline{0.6014} & \underline{0.6136} & \underline{0.6289} \\
         & LFD & \textbf{0.7380} & \textbf{0.9600} & \textbf{0.7182} & \textbf{0.6974} & \textbf{0.7016} & \textbf{0.6663} & \textbf{0.6342} & \textbf{0.6034} & \textbf{0.6148} & \textbf{0.6301} \\
        \bottomrule
    \end{tabular}
    \vspace{-10pt}
    \label{tab:main}
\end{table}

Table~\ref{tab:main} summarizes the accuracy of RAG based on four different models across four QA datasets. From the table, we have the following observations:
(1) \textbf{
\name matches or exceeds the performance of noise-injection strategies.}
We can see our method demonstrates consistent performance gains, ranging from minimal 0.29\%~(Qwen3-8B,~RGB) to maximal 16.76\%~(DeepSeek-7B, 2WikiMultihopQA) improvement.
Injecting noise, i.e., GD~(12), shows performance gains on some datasets and models, with the highest improvement being 10.13\%~(DeepSeek-7B,~2WikiMultihopQA). However, it significantly degrades performance for Mistral-7B and Qwen3-8B across all datasets, leading to a notable reduction in accuracy~(maximum $\Delta = -26.57\,\%$ on Mistral-7B).
(2) \textbf{\name outperforms decoding strategies without noise injection.}
Compared to alternative decoding methods, \name consistently delivers superior performance. While methods like DoLA and CS show strong results in specific cases, e.g., achieving 1.94\% and 13.55\% gains on DeepSeek-7B with the 2WikiMultihopQA dataset, they occasionally underperform even relative to the greedy decoding baseline.
In particular, DoLA shows the largest decline of 30\% on DeepSeek-7B with the RGB dataset, while CS exhibits a maximum drop of 14.57\% on Mistral-7B with the RGB dataset.
These results indicate that both strategies are sensitive to specific model architectures or data characteristics.
\textbf{Additional comparisons of these methods under different noise levels are provided in Appendix~\ref{appendix-c}.}
(3) \textbf{The fusion layer selection plays a critical role for RAG.}
Our dynamic layer selection strategy consistently outperforms the random approach, with an average improvement of 3.11\%, particularly achieving a 15.67\% gain on DeepSeek-7B/RGB, demonstrating its efficacy.

\subsection{Analysis Experiments}\label{ablation}

\begin{table}[!thb]
    \caption{Comparison of accuracy between different layer selection ranges under dynamic layer selection strategy. \textbf{Bold} indicate the best performance, while \underline{underline} represent the second-best.
    LFD[0, 16) and LFD[16, 32) mean selecting layers from the earlier and later half respectively. 
    }
    \centering
    \setlength{\tabcolsep}{4pt}
    \scriptsize
    \begin{tabular}{llcccccccccc}
        \toprule
        & & \multirow{3}{*}{NQ} & \multirow{2}{*}{RGB} & \multicolumn{3}{c}{HotpotQA} & \multicolumn{5}{c}{2WikiMultihopQA} \\
        \cmidrule(lr){5-7} \cmidrule(lr){8-12}
        & & & & Compare & Bridge & Total & Compare & Bridge & Inf & Compose & Total \\
        \hline\hline
        \multirow{3}{*}{Llama2-7B}
        & GD~(0) & 0.5745 & \underline{0.8900} & \underline{0.4755} & \underline{0.4108} & \underline{0.4237} & \underline{0.3095} & \underline{0.2815} & \underline{0.2607} & \underline{0.2928} & \underline{0.2904} \\
        & LFD[0, 16) &  \underline{0.5758} & 0.8833 & 0.4371 & 0.4027 & 0.4096 & 0.2956 & 0.2582 & 0.2529 & 0.2739 & 0.2731 \\
        & LFD[16, 32) & \textbf{0.5949} & \textbf{0.9067} & \textbf{0.5454} & \textbf{0.4295} & \textbf{0.4528} & \textbf{0.3174} & \textbf{0.3005} & \textbf{0.3043} & \textbf{0.3232} & \textbf{0.3144} \\
        \midrule
        \multirow{3}{*}{Mistral-7B}
        & GD~(0) & \underline{0.6130} & \underline{0.8900} & \underline{0.5864} & 0.5889 & 0.5884 & 0.5518 & 0.5204 & 0.5605 & 0.5337 & 0.5385 \\
        & LFD[0, 16) & 0.6081 & 0.8867 & 0.5864 & \underline{0.5919} & \underline{0.5908} & \underline{0.5551} & \underline{0.5218} & \underline{0.5650} & \underline{0.5402} & \underline{0.5429} \\
        & LFD[16, 32) & \textbf{0.6357} & \textbf{0.9367} & \textbf{0.6026} & \textbf{0.6093} & \textbf{0.6079} & \textbf{0.5813} & \textbf{0.5434} & \textbf{0.5982} & \textbf{0.5645} & \textbf{0.5680} \\
        \bottomrule
    \end{tabular}
    \label{tab:lower}
    \vspace{-2pt}
\end{table}

\paragraph{Effects of late-stage layer Integration}
We assess how different layer selection ranges affect performance using Llama2-7B and Mistral-7B across four benchmark datasets in Table~\ref{tab:lower}.
As Table~\ref{tab:lower} shows, our approach~(selecting from layers 16–32) consistently achieves higher accuracy than selecting from earlier layers (layers 0\-16), with an average gain of 3.01\%. 
Notably, our approach provides marginal accuracy gains ($\leq$0.1\%) over full-range selection (layers 0\-32), which indicates the advantage of the proposed layer selection strategy. 
These results demonstrate the greater efficacy of deeper layers for utilizing knowledge in RAG.
Results for DeepSeek-7B and Qwen3-8B are provided in Appendix~\ref{appendix-d}.

\begin{figure}[t]
    \centering
    \begin{subfigure}[b]{0.325\textwidth} 
        \includegraphics[width=\linewidth]{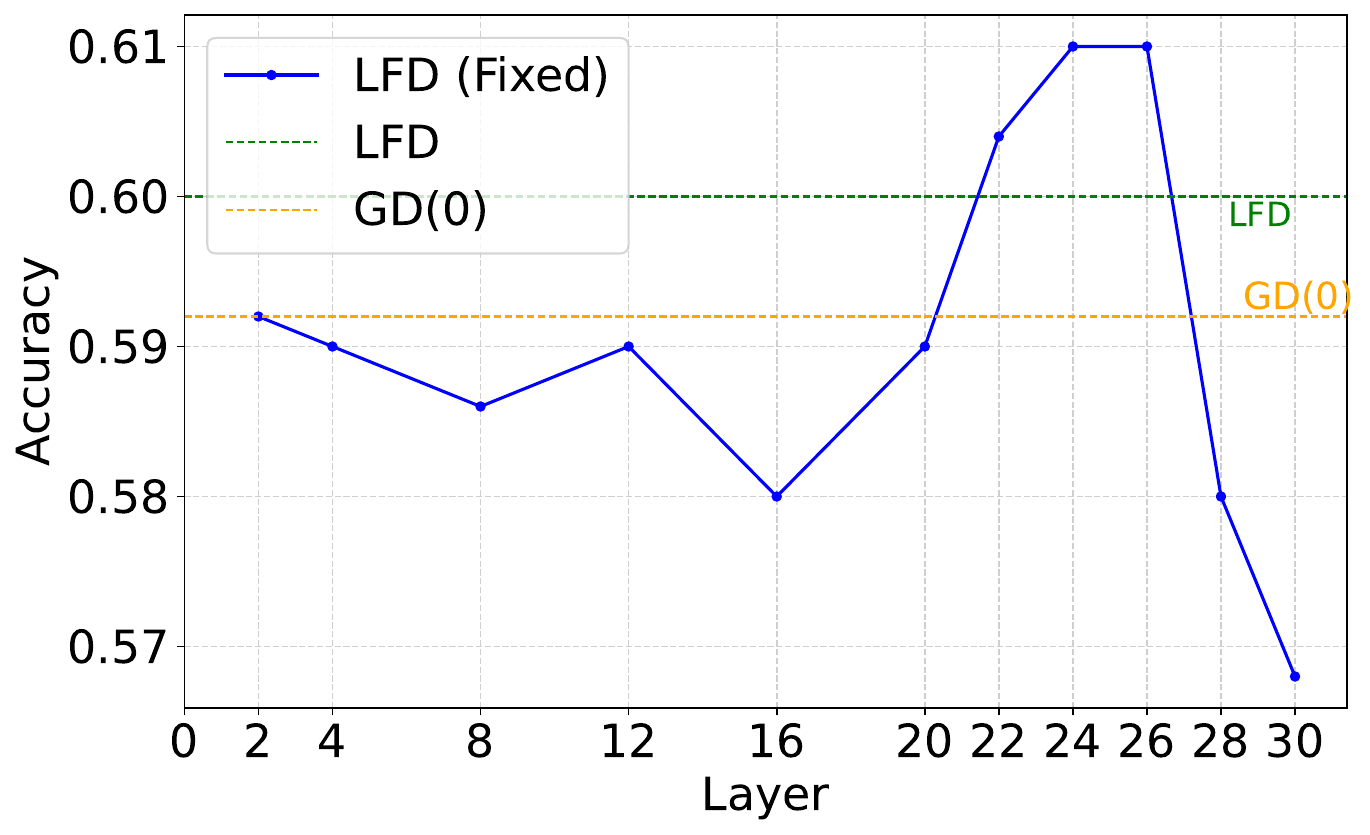}
        \vspace{-10pt}
        \caption{\tiny LFD vs. LFD~(Fixed) on NQ dataset.}
        \label{fig:nq-layer-compare}
    \end{subfigure}
    \hfill
    \begin{subfigure}[b]{0.325\textwidth}
        \includegraphics[width=\linewidth]{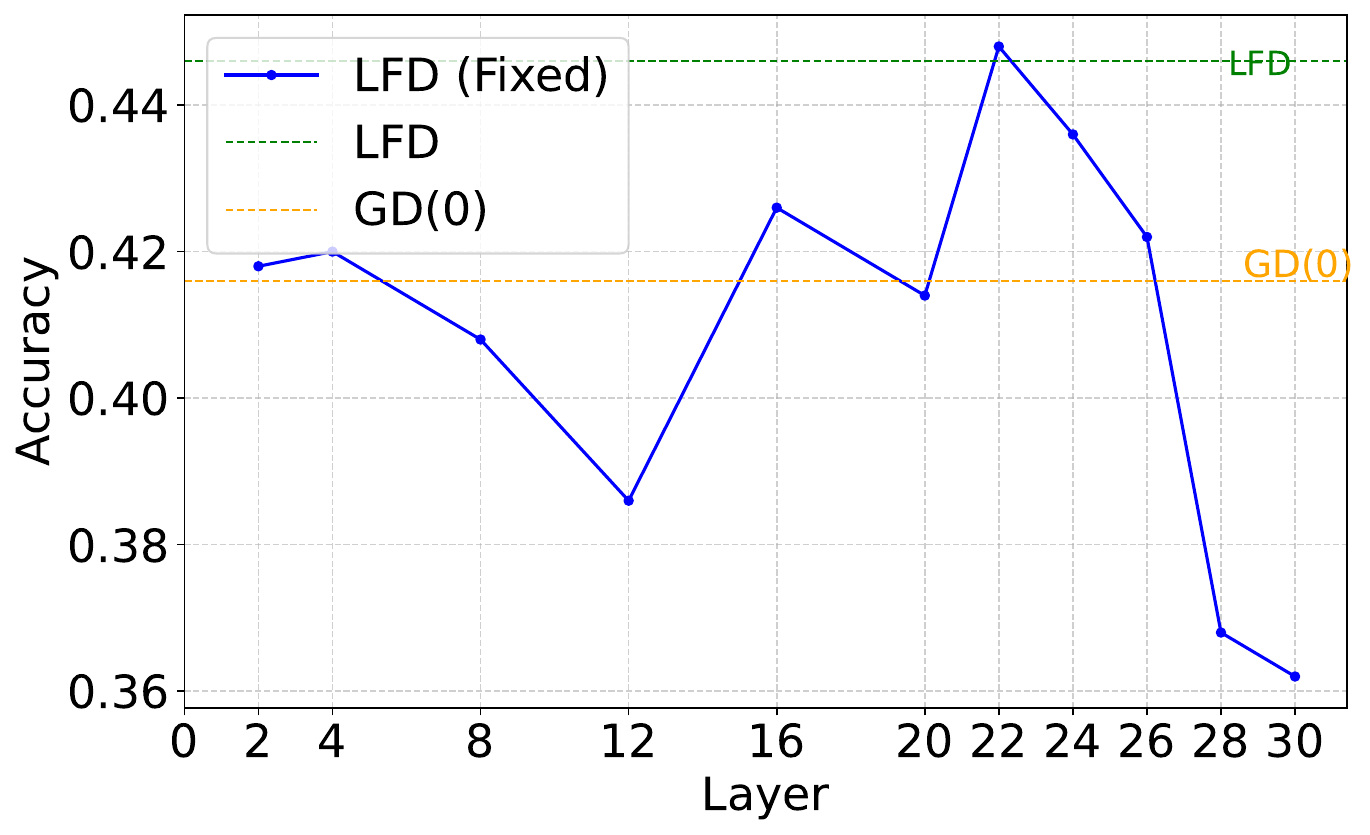}
        \vspace{-10pt}
        \caption{\tiny LFD vs. LFD~(Fixed) on HotpotQA dataset.}
        \label{fig:hotpot-layer-compare}
    \end{subfigure}
    \hfill
    \begin{subfigure}[b]{0.325\textwidth}
        \includegraphics[width=\linewidth]{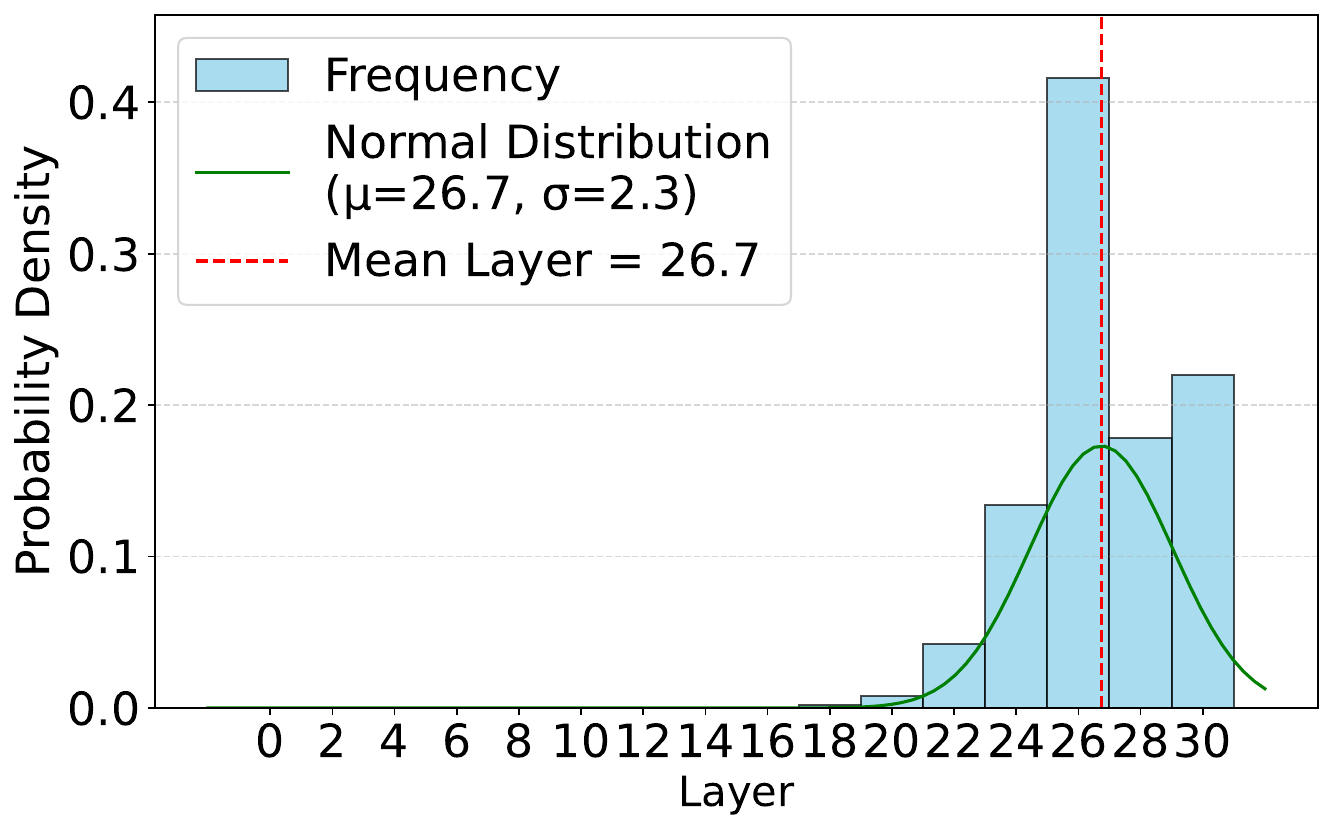}
        \vspace{-10pt}
        \caption{\tiny Histgram of layer selection in \name.}
        \label{fig:nq-layer-selected}
    \end{subfigure}
    \vspace{-5pt}
    \caption{Comparison between \name and \name~(Fixed) on the NQ (a) and HotpotQA (b) datasets. (c) illustrates the layer selection distribution in \name compared to the optimal fixed layer selection.}
    \vspace{-13pt}
    \label{fig:layer-seletion-strategy}
\end{figure}

\paragraph{Effects of the lowest IKS layer selection}
We evaluate the effectiveness of the lowest IKS layer selection
on the NQ and HotpotQA datasets. Extended results across models and datasets appear in Appendix~\ref{appendix-e}. 
First, we compare our dynamic strategy with fixed-layer selection \name~(Fixed) in Figure~\ref{fig:layer-seletion-strategy}(a-b). 
Second, we analyze the distribution of dynamical layer selections against the optimal fixed-layer baseline on NQ (Figure~\ref{fig:layer-seletion-strategy}(c)). 
Key findings emerge:  
\textbf{(1) Fixed-layer selection requires dataset-specific validation for optimality}.
While fixed-layer achieves peak performance at layer $24-26$ (NQ) and layer $22$ (HotpotQA), these layers differ across datasets, necessitating extra validating datasets.  
\textbf{(2) The lowest IKS tends to achieve near-optimal performance with small margins}. Compared to the best fixed-layer results, the lowest IKS exhibits performance gaps of only 1.8\% (NQ) and 0.2\% (HotpotQA), demonstrating robust generalization without dataset-specific tuning.
\textbf{(3) Dynamic layer selection concentrates near the optimal fixed-layer.} For NQ dataset, the lowest IKS selections cluster around layer $26$ (Figure~\ref{fig:layer-seletion-strategy}(c)), aligning closely with the optimal fixed layer despite stochasticity.

\begin{wraptable}{r}{0.5\textwidth}
\vspace{-10pt}
\caption{Input Length, Decoding Latency (ms), Throughput (tokens/s), and GPU Overhead (MB).}
\setlength{\tabcolsep}{2pt}
\scriptsize
\begin{tabular}{lrrrrr}
    \toprule
     & \multicolumn{1}{c}{Input}  & \multicolumn{1}{c}{Latency} & \multicolumn{1}{c}{Throughput} & \multicolumn{1}{c}{GPU Memory} \\
     & \multicolumn{1}{c}{Length} & \multicolumn{1}{c}{(ms)} & \multicolumn{1}{c}{(tokens/s)} & \multicolumn{1}{c}{(MB)} \\
     \hline\hline
     GD~(0) & 239\,\raisebox{-0.8ex}{\textsuperscript{($\times$\textbf{1.00})}} & 42.23\,\raisebox{-0.8ex}{\textsuperscript{($\times$\textbf{1.00})}} & 24.29\,\raisebox{-0.8ex}{\textsuperscript{($\times$\textbf{1.00})}} & 144.05\,\raisebox{-0.8ex}{\textsuperscript{($\times$\textbf{1.00})}} \\ 
     GD~(4) & 958\,\raisebox{-0.8ex}{\textsuperscript{($\times$\textbf{4.01})}} & 73.42\,\raisebox{-0.8ex}{\textsuperscript{($\times$\textbf{1.74})}} & 14.95\,\raisebox{-0.8ex}{\textsuperscript{($\times$\textbf{0.62})}} & 572.23\,\raisebox{-0.8ex}{\textsuperscript{($\times$\textbf{3.97})}} \\
     GD~(8) & 1675\,\raisebox{-0.8ex}{\textsuperscript{($\times$\textbf{7.01})}} & 97.99\,\raisebox{-0.8ex}{\textsuperscript{($\times$\textbf{2.32})}} & 11.92\,\raisebox{-0.8ex}{\textsuperscript{($\times$\textbf{0.49})}} & 997.81\,\raisebox{-0.8ex}{\textsuperscript{($\times$\textbf{6.93})}} \\
     GD~(12) & 2396\,\raisebox{-0.8ex}{\textsuperscript{($\times$\textbf{10.02})}} & 129.67\,\raisebox{-0.8ex}{\textsuperscript{($\times$\textbf{3.07})}} & 9.31\,\raisebox{-0.8ex}{\textsuperscript{($\times$\textbf{0.38})}} & 1426.87\,\raisebox{-0.8ex}{\textsuperscript{($\times$\textbf{9.91})}} \\
     DoLA & 239\,\raisebox{-0.8ex}{\textsuperscript{($\times$\textbf{1.00})}} & 49.66\,\raisebox{-0.8ex}{\textsuperscript{($\times$\textbf{1.18})}} & 20.58\,\raisebox{-0.8ex}{\textsuperscript{($\times$\textbf{0.85})}} & 201.00\,\raisebox{-0.8ex}{\textsuperscript{($\times$\textbf{1.39})}} \\
     LFD & 239\,\raisebox{-0.8ex}{\textsuperscript{($\times$\textbf{1.00})}} & 52.25\,\raisebox{-0.8ex}{\textsuperscript{($\times$\textbf{1.24})}} & 19.56\,\raisebox{-0.8ex}{\textsuperscript{($\times$\textbf{0.81})}} & 203.15\,\raisebox{-0.8ex}{\textsuperscript{($\times$\textbf{1.41})}} \\
     \bottomrule
\end{tabular}
\label{tab:latency}
\vspace{-10pt}
\end{wraptable}

\paragraph{Latency, Throughput \& Memory Usage}
We compare the decoding latency, throughput, and GPU overhead between \name and the greedy decoding method (with varying levels of noise), and the experimental results are illustrated in Table~\ref{tab:latency}.
The results demonstrate that, compared to the noise-injection baselines, \name exhibits advantage in terms of decoding time and memory overhead.
Furthermore, when compared to noise-free decoding baseline DoLA, our approach incurs just 1.05$\times$ the latency and 1.01$\times$ the memory usage, \textbf{keeping efficiency on par with state-of-the-art decoding methods}.

\paragraph{Qualitative Study}
\vspace{-5pt}
In Table~\ref{tab:case}, we analyze a case study from the NQ dataset using the Llama2-7B model, evaluating four decoding strategies: GD(0), CS, DoLA, and LFD. Despite access to ground-truth documents, both GD(0) and DoLA generate incorrect answers (e.g., ``18 minutes''), suggesting limited capacity to integrate contextual evidence. Similarly, while CS produces a partially relevant response (``Texas Revolution''), it exhibits reduced factual consistency with the source material. In contrast, LFD demonstrates superior utilization of retrieved context, synthesizing a precise and factually aligned answer. Additional case studies and analyses are provided in Appendix~\ref{appendix-f}.

\begin{table}[t]
    \caption{Qualitative study on GD(0), CS, DoLA, and LFD on the NQ dataset using LLaMA2-7B.}
    \centering
    \scriptsize
    \begin{tabular}{c|l|l|l|p{5.0cm}}
        \bottomrule
        Method & GD(0) & CS & DoLA & LFD \\
        \hline \hline
        \multirow{7}{*}{Prompt} & \multicolumn{4}{p{10.8cm}}{\parbox[t]{10.5cm}{
        
        You are given a question and you MUST respond by EXTRACTING the answer from one of the provided documents. If none of the documents contain the answer, respond with NO-RES. ... 
        \\
        Document [20983057](Title: Battle of San Jacinto) {\color{blue}The Battle of San Jacinto} , fought on April 21 , 1836 , in present - day Harris County , Texas , {\color{blue}was the decisive battle of the Texas Revolution} . Led by General Sam Houston , the Texian Army engaged and defeated General Antonio López de Santa Anna 's Mexican army in a fight that lasted just 18 minutes ...\\
        Question: Texans won their independence as a result of what battle? Answer:
        }
        } \\
        \hline
           Ground Truth& \multicolumn{2}{p{3cm}}{Battle of San Jacinto} \\
            \hline
        \multirow{2}{*}{Answer} & \multirow{2}{*}{18 minutes} & \multirow{2}{*}{Texas Revolution} & \multirow{2}{*}{18 minutes} & \textbf{Texans won their independence as a result of the Battle of San Jacinto.} \\
        \toprule
    \end{tabular}
    \vspace{-14pt}
    \label{tab:case}
\end{table}

\section{Related Work}
\vspace{-10pt}
\paragraph{Retrieval Augmented Generation}
Retrieval-Augmented Generation~(RAG) enhances model reasoning by integrating relevant external knowledge retrieved through user queries.
Recent advances focus on three directions: refined retrieval mechanisms~\cite{adaptive-rag, refine-query}, structured knowledge organization~\cite{graph-rag-survey, kag-zero-shot}, and optimized context embedding~\cite{prompt-optim-1, prompt-survey}.
Self-RAG~\cite{selfrag} and FLARE~\cite{flare} achieve adaptive retrieval through self-evaluation and uncertainty prediction respectively, dynamically optimizing knowledge acquisition.
GraphRAG~\cite{g-retriever, graphrag, kgqa} advances reasoning capabilities by constructing document-derived knowledge graphs that capture semantic relationships for multi-hop inference.
Embedding optimizations include noise injection~\cite{DBLP:conf/sigir/CuconasuTSFCMTS24, pandora-noise} to counter overfitting through strategic low-relevance document insertion, and context re-ranking~\cite{rankrag, G-rag} that prioritizes high-utility knowledge via learned document scoring.
While these methods enhance knowledge orchestration through pipeline improvements, they systematically neglect the internal mechanisms through which LLMs process external information during generation.
Our work bridges this fundamental gap by surfacing stratified knowledge integration patterns across LLM's layers through systematic layer-wise analysis.

\vspace{-10pt}
\paragraph{Decoding Strategy in LLMs}
Decoding strategies are pivotal in transforming raw model probabilities into coherent text outputs, critically influencing the quality and factual integrity of LLM generations~\cite{decoding-strategies-survey, yunfan, xia2024unlocking}.
While traditional methods like greedy decoding and beam search~\cite{best-first-beam-search, self-eval-beam-search} remain prevalent, recent work has introduced advanced techniques to address their limitations.
Contrastive search (CS)~\cite{contrastive_search} balances diversity and coherence by selecting tokens through a weighted combination of probability and semantic dissimilarity to preceding context, mitigating repetition while preserving fluency.
Simple Decoding (FSD)~\cite{frustratingly-decoding} suppresses redundant patterns by dynamically constructing an anti-language model to penalize overused token sequences.
DoLa~\cite{dola} addresses hallucinations by contrasting later-layer logit distributions with earlier ones, prioritizing factually consistent predictions.
Building on these advances, we propose a novel decoding strategy for RAG that dynamically balances external retrieved knowledge with the model's internal parametric knowledge, enhancing factual accuracy by mitigating interference from outdated or conflicting internal representations.

\vspace{-10pt}
\paragraph{Hallucinations in LLMs}
Hallucinations in LLMs refer to instances where the model generates false or unsupported information not grounded in its reference data~\cite{halluc-1}.
Existing mitigation strategies include multi-agent debating, where multiple LLM instances collaborate to detect inconsistencies through iterative debates~\cite{multi-agent-debate-1, multi-agent-debate-2}; self-consistency verification, which aggregates and reconciles multiple reasoning paths to reduce individual errors~\cite{self-consistency}; and model editing, which directly modifies neural network weights to correct systematic factual errors~\cite{model-editing-1, model-editing-2}.
While RAG systems aim to ground responses in retrieved external knowledge, recent studies show that they still exhibit hallucinations, especially those that contradict the retrieved content~\cite{redepe}. To address this limitation, our work conducts an empirical study analyzing how LLMs internally process external knowledge in RAG settings by controlling the noise from different granularity. Based on these findings, we propose a novel decoding method designed to improve answer accuracy and reduce hallucination by enhancing the integration of retrieved evidence.

\vspace{-6pt}
\section{Conclusion}
\vspace{-10pt}
By analyzing how noise injection amplifies external knowledge exploitation in LLMs, we establish a functional demarcation across LLMs' layers: shallow (local context), intermediate (external knowledge), and deep (internal parametric knowledge). 
Leveraging this, we propose \name, a training-free decoding strategy that fuses intermediate-layer representations to enhance external knowledge integration in final outputs via a the lowest internal knowledge score to pinpoint the ideal fusion layer. 
Experiments across diverse benchmarks demonstrate that \name enhances factual grounding in RAG systems while incurring minimal computational overhead.

\section*{Acknowledgment}
We express our sincere gratitude for the financial support provided by the National Natural Science Foundation of China (No. U23A20305, NO. {62302345}), the {Natural Science Foundation of Wuhan} (NO. {2024050702030136}) and the {Xiaomi Young Scholar Program}.

\bibliographystyle{plain}
\bibliography{reference}

@inproceedings{DBLP:conf/icml/BorgeaudMHCRM0L22,
  author       = {Sebastian Borgeaud and
                  Arthur Mensch and
                  Jordan Hoffmann and
                  Trevor Cai and
                  Eliza Rutherford and
                  Katie Millican and
                  George van den Driessche and
                  Jean{-}Baptiste Lespiau and
                  Bogdan Damoc and
                  Aidan Clark and
                  Diego de Las Casas and
                  Aurelia Guy and
                  Jacob Menick and
                  Roman Ring and
                  Tom Hennigan and
                  Saffron Huang and
                  Loren Maggiore and
                  Chris Jones and
                  Albin Cassirer and
                  Andy Brock and
                  Michela Paganini and
                  Geoffrey Irving and
                  Oriol Vinyals and
                  Simon Osindero and
                  Karen Simonyan and
                  Jack W. Rae and
                  Erich Elsen and
                  Laurent Sifre},
  title        = {Improving Language Models by Retrieving from Trillions of Tokens},
  booktitle    = {International Conference on Machine Learning, {ICML} 2022, 17-23 July
                  2022, Baltimore, Maryland, {USA}},
  series       = {Proceedings of Machine Learning Research},
  volume       = {162},
  pages        = {2206--2240},
  publisher    = {{PMLR}},
  year         = {2022},
}

@inproceedings{DBLP:conf/icml/GuuLTPC20,
  author       = {Kelvin Guu and
                  Kenton Lee and
                  Zora Tung and
                  Panupong Pasupat and
                  Ming{-}Wei Chang},
  title        = {Retrieval Augmented Language Model Pre-Training},
  booktitle    = {Proceedings of the 37th International Conference on Machine Learning,
                  {ICML} 2020, 13-18 July 2020, Virtual Event},
  series       = {Proceedings of Machine Learning Research},
  volume       = {119},
  pages        = {3929--3938},
  publisher    = {{PMLR}},
  year         = {2020},
}

@inproceedings{DBLP:conf/nips/LewisPPPKGKLYR020,
  author       = {Patrick Lewis and
                  Ethan Perez and
                  Aleksandra Piktus and
                  Fabio Petroni and
                  Vladimir Karpukhin and
                  Naman Goyal and
                  Heinrich K{\"{u}}ttler and
                  Mike Lewis and
                  Wen{-}tau Yih and
                  Tim Rockt{\"{a}}schel and
                  Sebastian Riedel and
                  Douwe Kiela},
  title        = {Retrieval-Augmented Generation for Knowledge-Intensive {NLP} Tasks},
  booktitle    = {Advances in Neural Information Processing Systems 33: Annual Conference
                  on Neural Information Processing Systems 2020, NeurIPS 2020, December
                  6-12, 2020, virtual},
  year         = {2020},
}

@inproceedings{DBLP:conf/sigir/CuconasuTSFCMTS24,
  author       = {Florin Cuconasu and
                  Giovanni Trappolini and
                  Federico Siciliano and
                  Simone Filice and
                  Cesare Campagnano and
                  Yoelle Maarek and
                  Nicola Tonellotto and
                  Fabrizio Silvestri},
  title        = {The Power of Noise: Redefining Retrieval for {RAG} Systems},
  booktitle    = {Proceedings of the 47th International {ACM} {SIGIR} Conference on
                  Research and Development in Information Retrieval, {SIGIR} 2024, Washington
                  DC, USA, July 14-18, 2024},
  pages        = {719--729},
  publisher    = {{ACM}},
  year         = {2024},
}

@article{DBLP:journals/corr/abs-2502-10634,
  author       = {Hao Sun and
                  Chenming Tang and
                  Gengyang Li and
                  Yunfang Wu},
  title        = {Lost in the Passage: Passage-level In-context Learning Does Not Necessarily
                  Need a "Passage"},
  journal      = {CoRR},
  volume       = {abs/2502.10634},
  year         = {2025},
}

@inproceedings{selfrag,
  title={Self-{RAG}: Learning to Retrieve, Generate, and Critique through Self-Reflection},
  author={Akari Asai and Zeqiu Wu and Yizhong Wang and Avirup Sil and Hannaneh Hajishirzi},
  booktitle={The Twelfth International Conference on Learning Representations},
  year={2024},
}

@inproceedings{flare,
    title = "Active Retrieval Augmented Generation",
    author = "Jiang, Zhengbao  and
      Xu, Frank  and
      Gao, Luyu  and
      Sun, Zhiqing  and
      Liu, Qian  and
      Dwivedi-Yu, Jane  and
      Yang, Yiming  and
      Callan, Jamie  and
      Neubig, Graham",
    editor = "Bouamor, Houda  and
      Pino, Juan  and
      Bali, Kalika",
    booktitle = "Proceedings of the 2023 Conference on Empirical Methods in Natural Language Processing",
    month = dec,
    year = "2023",
    address = "Singapore",
    publisher = "Association for Computational Linguistics",
    url = "https://aclanthology.org/2023.emnlp-main.495/",
    doi = "10.18653/v1/2023.emnlp-main.495",
    pages = "7969--7992",
}

@article{g-retriever,
  title={G-retriever: Retrieval-augmented generation for textual graph understanding and question answering},
  author={He, Xiaoxin and Tian, Yijun and Sun, Yifei and Chawla, Nitesh and Laurent, Thomas and LeCun, Yann and Bresson, Xavier and Hooi, Bryan},
  journal={Advances in Neural Information Processing Systems},
  volume={37},
  pages={132876--132907},
  year={2024}
}

@article{graphrag,
  title={From local to global: A graph rag approach to query-focused summarization},
  author={Edge, Darren and Trinh, Ha and Cheng, Newman and Bradley, Joshua and Chao, Alex and Mody, Apurva and Truitt, Steven and Metropolitansky, Dasha and Ness, Robert Osazuwa and Larson, Jonathan},
  journal={arXiv preprint arXiv:2404.16130},
  year={2024}
}

@inproceedings{kgqa,
  title={Uni{KGQA}: Unified Retrieval and Reasoning for Solving Multi-hop Question Answering Over Knowledge Graph},
  author={Jinhao Jiang and Kun Zhou and Xin Zhao and Ji-Rong Wen},
  booktitle={The Eleventh International Conference on Learning Representations },
  year={2023},
  url={https://openreview.net/forum?id=Z63RvyAZ2Vh}
}

@article{pandora-noise,
  title={Pandora's Box or Aladdin's Lamp: A Comprehensive Analysis Revealing the Role of RAG Noise in Large Language Models},
  author={Wu, Jinyang and Che, Feihu and Zhang, Chuyuan and Tao, Jianhua and Zhang, Shuai and Shao, Pengpeng},
  journal={arXiv preprint arXiv:2408.13533},
  year={2024}
}

@inproceedings{rankrag,
  author = {Yu, Yue and Ping, Wei and Liu, Zihan and Wang, Boxin and You, Jiaxuan and Zhang, Chao and Shoeybi, Mohammad and Catanzaro, Bryan},
  booktitle = {Advances in Neural Information Processing Systems},
  editor = {A. Globerson and L. Mackey and D. Belgrave and A. Fan and U. Paquet and J. Tomczak and C. Zhang},
  pages = {121156--121184},
  publisher = {Curran Associates, Inc.},
  title = {RankRAG: Unifying Context Ranking with Retrieval-Augmented Generation in LLMs},
  url = {https://proceedings.neurips.cc/paper_files/paper/2024/file/db93ccb6cf392f352570dd5af0a223d3-Paper-Conference.pdf},
  volume = {37},
  year = {2024}
}

@misc{G-rag,
  title={Don't Forget to Connect! Improving RAG with Graph-based Reranking}, 
  author={Jialin Dong and Bahare Fatemi and Bryan Perozzi and Lin F. Yang and Anton Tsitsulin},
  year={2024},
  eprint={2405.18414},
  archivePrefix={arXiv},
  primaryClass={cs.CL},
  url={https://arxiv.org/abs/2405.18414}, 
}

@inproceedings{redepe,
  title={ReDe{EP}: Detecting Hallucination in Retrieval-Augmented Generation via Mechanistic Interpretability},
  author={ZhongXiang Sun and Xiaoxue Zang and Kai Zheng and Jun Xu and Xiao Zhang and Weijie Yu and Yang Song and Han Li},
  booktitle={The Thirteenth International Conference on Learning Representations},
  year={2025},
  url={https://openreview.net/forum?id=ztzZDzgfrh}
}

@inproceedings{ragtruth,
  title={RAGTruth: A Hallucination Corpus for Developing Trustworthy Retrieval-Augmented Language Models},
  author={Niu, Cheng and Wu, Yuanhao and Zhu, Juno and Xu, Siliang and Shum, Kashun and Zhong, Randy and Song, Juntong and Zhang, Tong},
  booktitle={Proceedings of the 62nd Annual Meeting of the Association for Computational Linguistics (Volume 1: Long Papers)},
  pages={10862--10878},
  year={2024}
}

@article{lora,
  title={Lora: Low-rank adaptation of large language models.},
  author={Hu, Edward J and Shen, Yelong and Wallis, Phillip and Allen-Zhu, Zeyuan and Li, Yuanzhi and Wang, Shean and Wang, Lu and Chen, Weizhu and others},
  journal={ICLR},
  volume={1},
  number={2},
  pages={3},
  year={2022}
}

@inproceedings{lora+,
  title={LoRA+ efficient low rank adaptation of large models},
  author={Hayou, Soufiane and Ghosh, Nikhil and Yu, Bin},
  booktitle={Proceedings of the 41st International Conference on Machine Learning},
  pages={17783--17806},
  year={2024}
}

@article{prompt-survey,
  title={A systematic survey of prompt engineering in large language models: Techniques and applications},
  author={Sahoo, Pranab and Singh, Ayush Kumar and Saha, Sriparna and Jain, Vinija and Mondal, Samrat and Chadha, Aman},
  journal={arXiv preprint arXiv:2402.07927},
  year={2024}
}

@article{contrastive_search,
  title={A contrastive framework for neural text generation},
  author={Su, Yixuan and Lan, Tian and Wang, Yan and Yogatama, Dani and Kong, Lingpeng and Collier, Nigel},
  journal={Advances in Neural Information Processing Systems},
  volume={35},
  pages={21548--21561},
  year={2022}
}

@inproceedings{frustratingly-decoding,
    title = "A Frustratingly Simple Decoding Method for Neural Text Generation",
    author = "Yang, Haoran  and
      Cai, Deng  and
      Li, Huayang  and
      Bi, Wei  and
      Lam, Wai  and
      Shi, Shuming",
    editor = "Calzolari, Nicoletta  and
      Kan, Min-Yen  and
      Hoste, Veronique  and
      Lenci, Alessandro  and
      Sakti, Sakriani  and
      Xue, Nianwen",
    booktitle = "Proceedings of the 2024 Joint International Conference on Computational Linguistics, Language Resources and Evaluation (LREC-COLING 2024)",
    month = may,
    year = "2024",
    address = "Torino, Italia",
    publisher = "ELRA and ICCL",
    url = "https://aclanthology.org/2024.lrec-main.47/",
    pages = "536--557",
}

@inproceedings{dola,
  title={DoLa: Decoding by Contrasting Layers Improves Factuality in Large Language Models},
  author={Yung-Sung Chuang and Yujia Xie and Hongyin Luo and Yoon Kim and James R. Glass and Pengcheng He},
  booktitle={The Twelfth International Conference on Learning Representations},
  year={2024},
  url={https://openreview.net/forum?id=Th6NyL07na}
}

@inproceedings{decoding-strategies-survey,
  title={A Thorough Examination of Decoding Methods in the Era of LLMs},
  author={Shi, Chufan and Yang, Haoran and Cai, Deng and Zhang, Zhisong and Wang, Yifan and Yang, Yujiu and Lam, Wai},
  booktitle={Proceedings of the 2024 Conference on Empirical Methods in Natural Language Processing},
  pages={8601--8629},
  year={2024}
}

@article{best-first-beam-search,
  title={Best-first beam search},
  author={Meister, Clara and Vieira, Tim and Cotterell, Ryan},
  journal={Transactions of the Association for Computational Linguistics},
  volume={8},
  pages={795--809},
  year={2020},
  publisher={MIT Press One Rogers Street, Cambridge, MA 02142-1209, USA journals-info~…}
}

@article{self-eval-beam-search,
  title={Self-evaluation guided beam search for reasoning},
  author={Xie, Yuxi and Kawaguchi, Kenji and Zhao, Yiran and Zhao, James Xu and Kan, Min-Yen and He, Junxian and Xie, Michael},
  journal={Advances in Neural Information Processing Systems},
  volume={36},
  pages={41618--41650},
  year={2023}
}

@article{nq-dataset,
  title={Natural questions: a benchmark for question answering research},
  author={Kwiatkowski, Tom and Palomaki, Jennimaria and Redfield, Olivia and Collins, Michael and Parikh, Ankur and Alberti, Chris and Epstein, Danielle and Polosukhin, Illia and Devlin, Jacob and Lee, Kenton and others},
  journal={Transactions of the Association for Computational Linguistics},
  volume={7},
  pages={453--466},
  year={2019},
  publisher={MIT Press One Rogers Street, Cambridge, MA 02142-1209, USA journals-info~…}
}

@inproceedings{rgb-dataset,
  title={Benchmarking large language models in retrieval-augmented generation},
  author={Chen, Jiawei and Lin, Hongyu and Han, Xianpei and Sun, Le},
  booktitle={Proceedings of the AAAI Conference on Artificial Intelligence},
  volume={38},
  number={16},
  pages={17754--17762},
  year={2024}
}

@inproceedings{hotpotqa-dataset,
  title={HotpotQA: A Dataset for Diverse, Explainable Multi-hop Question Answering},
  author={Yang, Zhilin and Qi, Peng and Zhang, Saizheng and Bengio, Yoshua and Cohen, William and Salakhutdinov, Ruslan and Manning, Christopher D},
  booktitle={Proceedings of the 2018 Conference on Empirical Methods in Natural Language Processing},
  pages={2369--2380},
  year={2018}
}

@inproceedings{2wiki-dataset,
    title = "Constructing A Multi-hop {QA} Dataset for Comprehensive Evaluation of Reasoning Steps",
    author = "Ho, Xanh  and
      Duong Nguyen, Anh-Khoa  and
      Sugawara, Saku  and
      Aizawa, Akiko",
    editor = "Scott, Donia  and
      Bel, Nuria  and
      Zong, Chengqing",
    booktitle = "Proceedings of the 28th International Conference on Computational Linguistics",
    month = dec,
    year = "2020",
    address = "Barcelona, Spain (Online)",
    publisher = "International Committee on Computational Linguistics",
    url = "https://aclanthology.org/2020.coling-main.580/",
    doi = "10.18653/v1/2020.coling-main.580",
    pages = "6609--6625",
}

@article{llama2,
  title={Llama 2: Open foundation and fine-tuned chat models},
  author={Touvron, Hugo and Martin, Louis and Stone, Kevin and Albert, Peter and Almahairi, Amjad and Babaei, Yasmine and Bashlykov, Nikolay and Batra, Soumya and Bhargava, Prajjwal and Bhosale, Shruti and others},
  journal={arXiv preprint arXiv:2307.09288},
  year={2023}
}

@misc{mistral7b,
      title={Mistral 7B}, 
      author={Albert Q. Jiang and Alexandre Sablayrolles and Arthur Mensch and Chris Bamford and Devendra Singh Chaplot and Diego de las Casas and Florian Bressand and Gianna Lengyel and Guillaume Lample and Lucile Saulnier and Lélio Renard Lavaud and Marie-Anne Lachaux and Pierre Stock and Teven Le Scao and Thibaut Lavril and Thomas Wang and Timothée Lacroix and William El Sayed},
      year={2023},
      eprint={2310.06825},
      archivePrefix={arXiv},
      primaryClass={cs.CL},
      url={https://arxiv.org/abs/2310.06825}, 
}

@article{deepseek,
  title={Deepseek llm: Scaling open-source language models with longtermism},
  author={Bi, Xiao and Chen, Deli and Chen, Guanting and Chen, Shanhuang and Dai, Damai and Deng, Chengqi and Ding, Honghui and Dong, Kai and Du, Qiushi and Fu, Zhe and others},
  journal={arXiv preprint arXiv:2401.02954},
  year={2024}
}

@misc{qwen3,
    title  = {Qwen3},
    url    = {https://qwenlm.github.io/blog/qwen3/},
    author = {Qwen Team},
    month  = {April},
    year   = {2025}
}

@inproceedings{fnn-are-key-value-memories,
    title = "Transformer Feed-Forward Layers Are Key-Value Memories",
    author = "Geva, Mor  and
      Schuster, Roei  and
      Berant, Jonathan  and
      Levy, Omer",
    editor = "Moens, Marie-Francine  and
      Huang, Xuanjing  and
      Specia, Lucia  and
      Yih, Scott Wen-tau",
    booktitle = "Proceedings of the 2021 Conference on Empirical Methods in Natural Language Processing",
    month = nov,
    year = "2021",
    address = "Online and Punta Cana, Dominican Republic",
    publisher = "Association for Computational Linguistics",
    url = "https://aclanthology.org/2021.emnlp-main.446/",
    doi = "10.18653/v1/2021.emnlp-main.446",
    pages = "5484--5495",
}

@inproceedings{knowledge-neurons,
    title = "Knowledge Neurons in Pretrained Transformers",
    author = "Dai, Damai  and
      Dong, Li  and
      Hao, Yaru  and
      Sui, Zhifang  and
      Chang, Baobao  and
      Wei, Furu",
    editor = "Muresan, Smaranda  and
      Nakov, Preslav  and
      Villavicencio, Aline",
    booktitle = "Proceedings of the 60th Annual Meeting of the Association for Computational Linguistics (Volume 1: Long Papers)",
    month = may,
    year = "2022",
    address = "Dublin, Ireland",
    publisher = "Association for Computational Linguistics",
    url = "https://aclanthology.org/2022.acl-long.581/",
    doi = "10.18653/v1/2022.acl-long.581",
    pages = "8493--8502",
}

@inproceedings{metric-eval-ref-1,
  title={Large language models struggle to learn long-tail knowledge},
  author={Kandpal, Nikhil and Deng, Haikang and Roberts, Adam and Wallace, Eric and Raffel, Colin},
  booktitle={International Conference on Machine Learning},
  pages={15696--15707},
  year={2023},
  organization={PMLR}
}

@article{metric-eval-ref-2,
  title={Lost in the Middle: How Language Models Use Long Contexts},
  author={Liu, Nelson F and Lin, Kevin and Hewitt, John and Paranjape, Ashwin and Bevilacqua, Michele and Petroni, Fabio and Liang, Percy},
  journal={Transactions of the Association for Computational Linguistics},
  volume={12},
  year={2024}
}

@inproceedings{recom-1,
  title={A survey on rag meeting llms: Towards retrieval-augmented large language models},
  author={Fan, Wenqi and Ding, Yujuan and Ning, Liangbo and Wang, Shijie and Li, Hengyun and Yin, Dawei and Chua, Tat-Seng and Li, Qing},
  booktitle={Proceedings of the 30th ACM SIGKDD Conference on Knowledge Discovery and Data Mining},
  pages={6491--6501},
  year={2024}
}

@inproceedings{recom-2,
  title={Retrieval-augmented recommender system: Enhancing recommender systems with large language models},
  author={Di Palma, Dario},
  booktitle={Proceedings of the 17th ACM Conference on Recommender Systems},
  pages={1369--1373},
  year={2023}
}

@inproceedings{recom-3,
  title={Genrec: Large language model for generative recommendation},
  author={Ji, Jianchao and Li, Zelong and Xu, Shuyuan and Hua, Wenyue and Ge, Yingqiang and Tan, Juntao and Zhang, Yongfeng},
  booktitle={European Conference on Information Retrieval},
  pages={494--502},
  year={2024},
  organization={Springer}
}

@inproceedings{search-engine-1,
  title={Towards a search engine for machines: Unified ranking for multiple retrieval-augmented large language models},
  author={Salemi, Alireza and Zamani, Hamed},
  booktitle={Proceedings of the 47th International ACM SIGIR Conference on Research and Development in Information Retrieval},
  pages={741--751},
  year={2024}
}

@article{search-engine-2,
  title={When search engine services meet large language models: visions and challenges},
  author={Xiong, Haoyi and Bian, Jiang and Li, Yuchen and Li, Xuhong and Du, Mengnan and Wang, Shuaiqiang and Yin, Dawei and Helal, Sumi},
  journal={IEEE Transactions on Services Computing},
  year={2024},
  publisher={IEEE}
}

@INPROCEEDINGS{halluc-1,
  author={Perković, Gabrijela and Drobnjak, Antun and Botički, Ivica},
  booktitle={2024 47th MIPRO ICT and Electronics Convention (MIPRO)}, 
  title={Hallucinations in LLMs: Understanding and Addressing Challenges}, 
  year={2024},
  volume={},
  number={},
  pages={2084-2088},
  doi={10.1109/MIPRO60963.2024.10569238}
}

@inproceedings{multi-agent-debate-1,
  title={ReConcile: Round-Table Conference Improves Reasoning via Consensus among Diverse LLMs},
  author={Chen, Justin and Saha, Swarnadeep and Bansal, Mohit},
  booktitle={Proceedings of the 62nd Annual Meeting of the Association for Computational Linguistics (Volume 1: Long Papers)},
  pages={7066--7085},
  year={2024}
}

@inproceedings{multi-agent-debate-2,
  title={Improving factuality and reasoning in language models through multiagent debate},
  author={Du, Yilun and Li, Shuang and Torralba, Antonio and Tenenbaum, Joshua B and Mordatch, Igor},
  booktitle={Forty-first International Conference on Machine Learning},
  year={2023}
}

@article{self-consistency,
  title={Self-consistency improves chain of thought reasoning in language models},
  author={Wang, Xuezhi and Wei, Jason and Schuurmans, Dale and Le, Quoc and Chi, Ed and Narang, Sharan and Chowdhery, Aakanksha and Zhou, Denny},
  journal={arXiv preprint arXiv:2203.11171},
  year={2022}
}

@article{model-editing-1,
  title={Truthx: Alleviating hallucinations by editing large language models in truthful space},
  author={Zhang, Shaolei and Yu, Tian and Feng, Yang},
  journal={arXiv preprint arXiv:2402.17811},
  year={2024}
}

@article{model-editing-2,
  title={Aging with grace: Lifelong model editing with discrete key-value adaptors},
  author={Hartvigsen, Tom and Sankaranarayanan, Swami and Palangi, Hamid and Kim, Yoon and Ghassemi, Marzyeh},
  journal={Advances in Neural Information Processing Systems},
  volume={36},
  pages={47934--47959},
  year={2023}
}

@article{JSD,
  title={Divergence measures based on the Shannon entropy},
  author={Lin, Jianhua},
  journal={IEEE Transactions on Information theory},
  volume={37},
  number={1},
  pages={145--151},
  year={2002},
}

@article{DBLP:journals/corr/abs-2303-08112,
  author       = {Nora Belrose and
                  Zach Furman and
                  Logan Smith and
                  Danny Halawi and
                  Igor Ostrovsky and
                  Lev McKinney and
                  Stella Biderman and
                  Jacob Steinhardt},
  title        = {Eliciting Latent Predictions from Transformers with the Tuned Lens},
  journal      = {CoRR},
  volume       = {abs/2303.08112},
  year         = {2023},
}

@inproceedings{adaptive-rag,
    title = "Adaptive-{RAG}: Learning to Adapt Retrieval-Augmented Large Language Models through Question Complexity",
    author = "Jeong, Soyeong  and
      Baek, Jinheon  and
      Cho, Sukmin  and
      Hwang, Sung Ju  and
      Park, Jong",
    editor = "Duh, Kevin  and
      Gomez, Helena  and
      Bethard, Steven",
    booktitle = "Proceedings of the 2024 Conference of the North American Chapter of the Association for Computational Linguistics: Human Language Technologies (Volume 1: Long Papers)",
    month = jun,
    year = "2024",
    address = "Mexico City, Mexico",
    publisher = "Association for Computational Linguistics",
    url = "https://aclanthology.org/2024.naacl-long.389/",
    doi = "10.18653/v1/2024.naacl-long.389",
    pages = "7036--7050",
}

@article{refine-query,
  title={Rq-rag: Learning to refine queries for retrieval augmented generation},
  author={Chan, Chi-Min and Xu, Chunpu and Yuan, Ruibin and Luo, Hongyin and Xue, Wei and Guo, Yike and Fu, Jie},
  journal={arXiv preprint arXiv:2404.00610},
  year={2024}
}

@INPROCEEDINGS{graph-rag-survey,
  author={Procko, Tyler Thomas and Ochoa, Omar},
  booktitle={2024 Conference on AI, Science, Engineering, and Technology (AIxSET)}, 
  title={Graph Retrieval-Augmented Generation for Large Language Models: A Survey}, 
  year={2024},
  volume={},
  number={},
  pages={166-169},
  doi={10.1109/AIxSET62544.2024.00030}
}

@inproceedings{kag-zero-shot,
    title = "Knowledge-Augmented Language Model Prompting for Zero-Shot Knowledge Graph Question Answering",
    author = "Baek, Jinheon  and
      Aji, Alham Fikri  and
      Saffari, Amir",
    editor = "Dalvi Mishra, Bhavana  and
      Durrett, Greg  and
      Jansen, Peter  and
      Neves Ribeiro, Danilo  and
      Wei, Jason",
    booktitle = "Proceedings of the 1st Workshop on Natural Language Reasoning and Structured Explanations (NLRSE)",
    month = jun,
    year = "2023",
    address = "Toronto, Canada",
    publisher = "Association for Computational Linguistics",
    url = "https://aclanthology.org/2023.nlrse-1.7/",
    doi = "10.18653/v1/2023.nlrse-1.7",
    pages = "78--106",
}

@inproceedings{prompt-optim-1,
  title={Prompt perturbation in retrieval-augmented generation based large language models},
  author={Hu, Zhibo and Wang, Chen and Shu, Yanfeng and Paik, Hye-Young and Zhu, Liming},
  booktitle={Proceedings of the 30th ACM SIGKDD Conference on Knowledge Discovery and Data Mining},
  pages={1119--1130},
  year={2024}
}

@article{rag-nlp-tasks,
  title={Retrieval-augmented generation for knowledge-intensive nlp tasks},
  author={Lewis, Patrick and Perez, Ethan and Piktus, Aleksandra and Petroni, Fabio and Karpukhin, Vladimir and Goyal, Naman and K{\"u}ttler, Heinrich and Lewis, Mike and Yih, Wen-tau and Rockt{\"a}schel, Tim and others},
  journal={Advances in neural information processing systems},
  volume={33},
  pages={9459--9474},
  year={2020}
}

@inproceedings{frames,
    title = "Fact, Fetch, and Reason: A Unified Evaluation of Retrieval-Augmented Generation",
    author = "Krishna, Satyapriya  and
      Krishna, Kalpesh  and
      Mohananey, Anhad  and
      Schwarcz, Steven  and
      Stambler, Adam  and
      Upadhyay, Shyam  and
      Faruqui, Manaal",
    editor = "Chiruzzo, Luis  and
      Ritter, Alan  and
      Wang, Lu",
    booktitle = "Proceedings of the 2025 Conference of the Nations of the Americas Chapter of the Association for Computational Linguistics: Human Language Technologies (Volume 1: Long Papers)",
    month = apr,
    year = "2025",
    address = "Albuquerque, New Mexico",
    publisher = "Association for Computational Linguistics",
    url = "https://aclanthology.org/2025.naacl-long.243/",
    doi = "10.18653/v1/2025.naacl-long.243",
    pages = "4745--4759",
    ISBN = "979-8-89176-189-6",
}

@inproceedings{mirage,
    title = "{MIRAGE}: A Metric-Intensive Benchmark for Retrieval-Augmented Generation Evaluation",
    author = "Park, Chanhee  and
      Moon, Hyeonseok  and
      Park, Chanjun  and
      Lim, Heuiseok",
    editor = "Chiruzzo, Luis  and
      Ritter, Alan  and
      Wang, Lu",
    booktitle = "Findings of the Association for Computational Linguistics: NAACL 2025",
    month = apr,
    year = "2025",
    address = "Albuquerque, New Mexico",
    publisher = "Association for Computational Linguistics",
    url = "https://aclanthology.org/2025.findings-naacl.157/",
    doi = "10.18653/v1/2025.findings-naacl.157",
    pages = "2883--2900",
    ISBN = "979-8-89176-195-7",
}

@inproceedings{yunfan,
  title={Mitigating language confusion through inference-time intervention},
  author={Xie, Yunfan and Zou, Lixin and Luo, Dan and Tang, Min and Li, Chenliang and Dong, Liming and Luo, Xiangyang}
}

@inproceedings{meta-lora,
  title={META-LORA: Memory-efficient sample reweighting for fine-tuning large language models},
  author={Li, Weicheng and Zou, Lixin and Tang, Min and Yu, Qing and Li, Wanli and Li, Chenliang},
  booktitle={Proceedings of the 31st International Conference on Computational Linguistics},
  pages={8504--8517},
  year={2025}
}

@inproceedings{li2025token,
  title={Token-level Preference Self-Alignment Optimization for Multi-style Outline Controllable Generation},
  author={Li, Zihao and Xu, Xuekong and Chen, Ziyao and Zou, Lixin and Ethanhjwu, Ethanhjwu and Chen, Qiang and Li, Chenliang},
  booktitle={Findings of the Association for Computational Linguistics: ACL 2025},
  pages={15974--16007},
  year={2025}
}

@article{dpo,
  title={Direct preference optimization: Your language model is secretly a reward model},
  author={Rafailov, Rafael and Sharma, Archit and Mitchell, Eric and Manning, Christopher D and Ermon, Stefano and Finn, Chelsea},
  journal={Advances in neural information processing systems},
  volume={36},
  pages={53728--53741},
  year={2023}
}

@article{xia2024unlocking,
  title={Unlocking efficiency in large language model inference: A comprehensive survey of speculative decoding},
  author={Xia, Heming and Yang, Zhe and Dong, Qingxiu and Wang, Peiyi and Li, Yongqi and Ge, Tao and Liu, Tianyu and Li, Wenjie and Sui, Zhifang},
  journal={arXiv preprint arXiv:2401.07851},
  year={2024}
}
\newpage
\appendix
\section{Limitations}\label{limitations}
While our approach demonstrates promising results in improving model outputs, several inherent constraints should be acknowledged.
The methodology primarily focuses on factuality, without incorporating broader alignment techniques like reinforcement learning from human feedback (RLHF)~\cite{dpo, li2025token}, which adapts outputs to human preference styles.
Furthermore, the current implementation operates directly on existing pretrained models without additional fine-tuning strategies~\cite{lora, lora+, meta-lora}, which may constrain potential performance gains.
These considerations suggest that while the current approach shows initial success, future work could explore integration with human preference alignment and fine-tuning strategies to further enhance model performance.

\section{Comprehensive Analysis of External Knowledge Intervention in RAG}\label{appendix-b}

\begin{figure}[h]
    \centering
    \begin{subfigure}[b]{0.49\textwidth}
        \includegraphics[width=\linewidth]{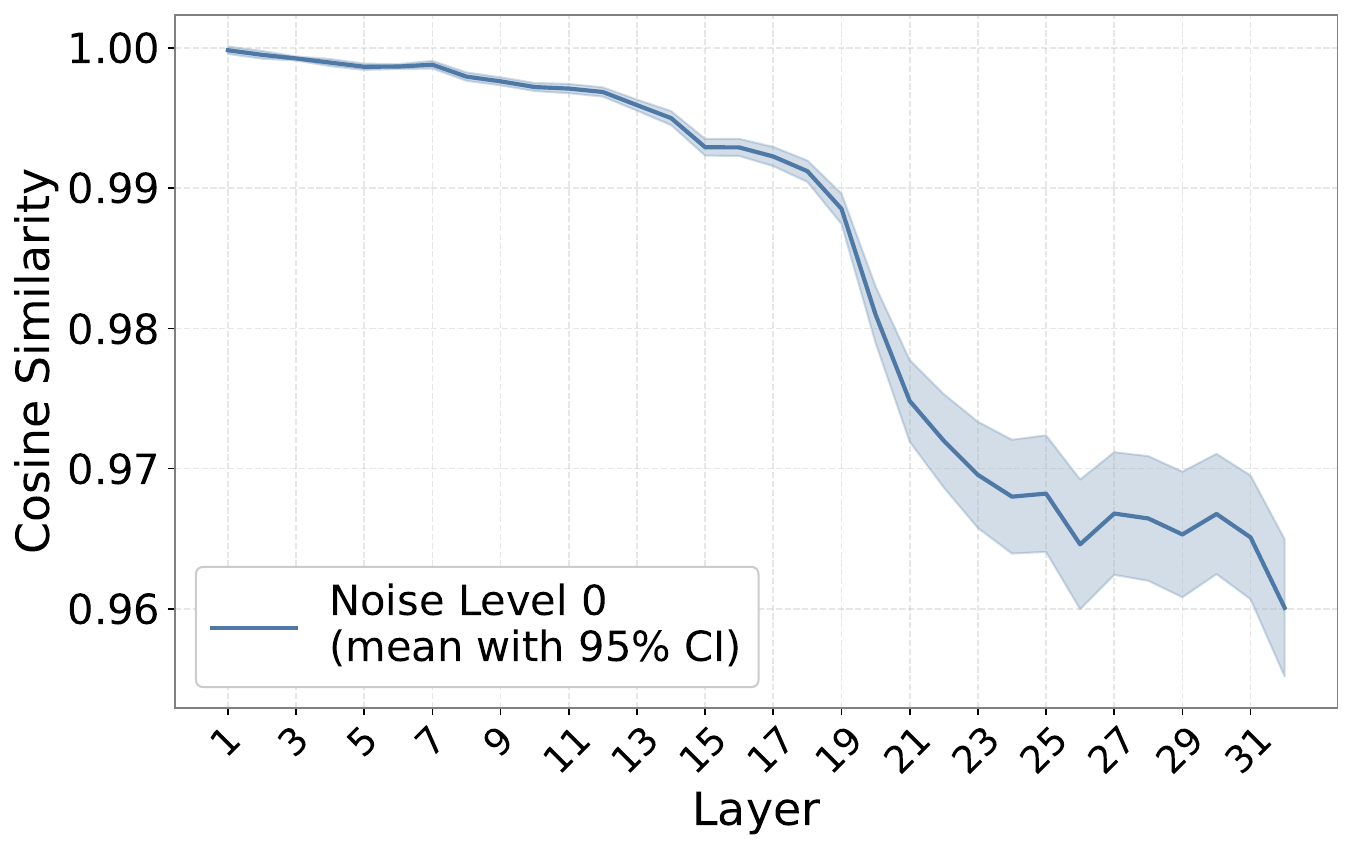}
        \caption{\small SimHidden (Smaller is better) on HotpotQA dataset.}
    \end{subfigure}
    \hfill
    \begin{subfigure}[b]{0.49\textwidth}
        \includegraphics[width=\linewidth]{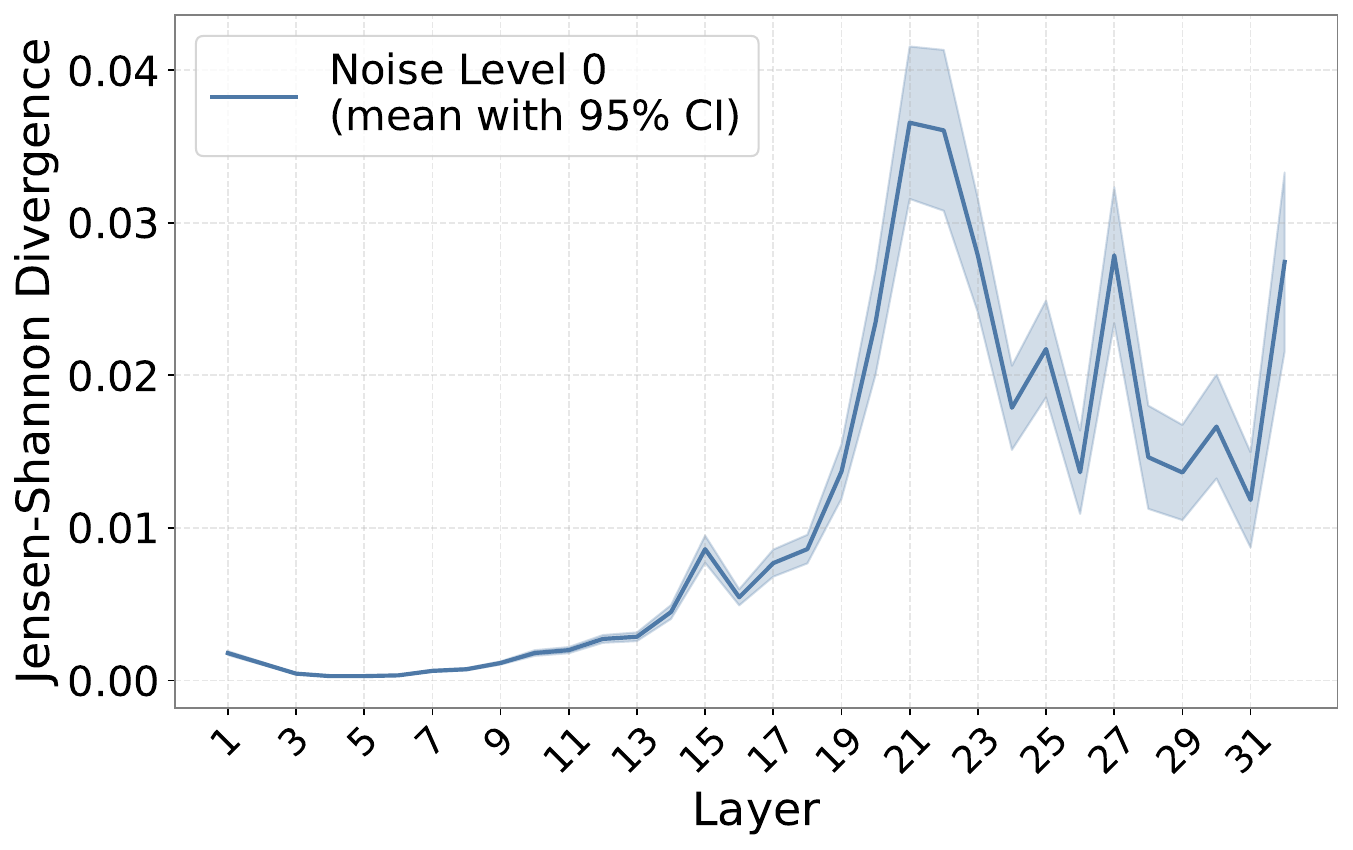}
        \caption{\small DiffAttn (Larger is better) on HotpotQA dataset.}
    \end{subfigure}
    
    \caption{(a)~Average SimHidden scores (with 95\% confidence intervals) across layers when noise level = 0;~(b)~Average DiffAttn scores (with 95\% confidence intervals) across layers when noise level = 0. Results are from Llama2-7B on HotpotQA dataset.}
    \label{fig:analysis-llama}
\end{figure}

In this section, we expand our analysis by incorporating a multi-hop question answering dataset HotpotQA to further quantify the impact of external knowledge.
As shown in Figure~\ref{fig:analysis-llama}, LLaMA2-7B exhibits a consistent three-stage pattern of knowledge utilization across layers, as reflected by the SimHidden scores: early layers (1--14), middle layers (15--26), and deeper layers (27--32). This stratification is further supported by the DiffAttn scores, which peak at layer 21 and remain lower in both ealier and latter layers, reinforcing the validity of the three-stage division.

\begin{figure}[h]
    \centering
    \begin{subfigure}[b]{0.49\textwidth}
        \includegraphics[width=\linewidth]{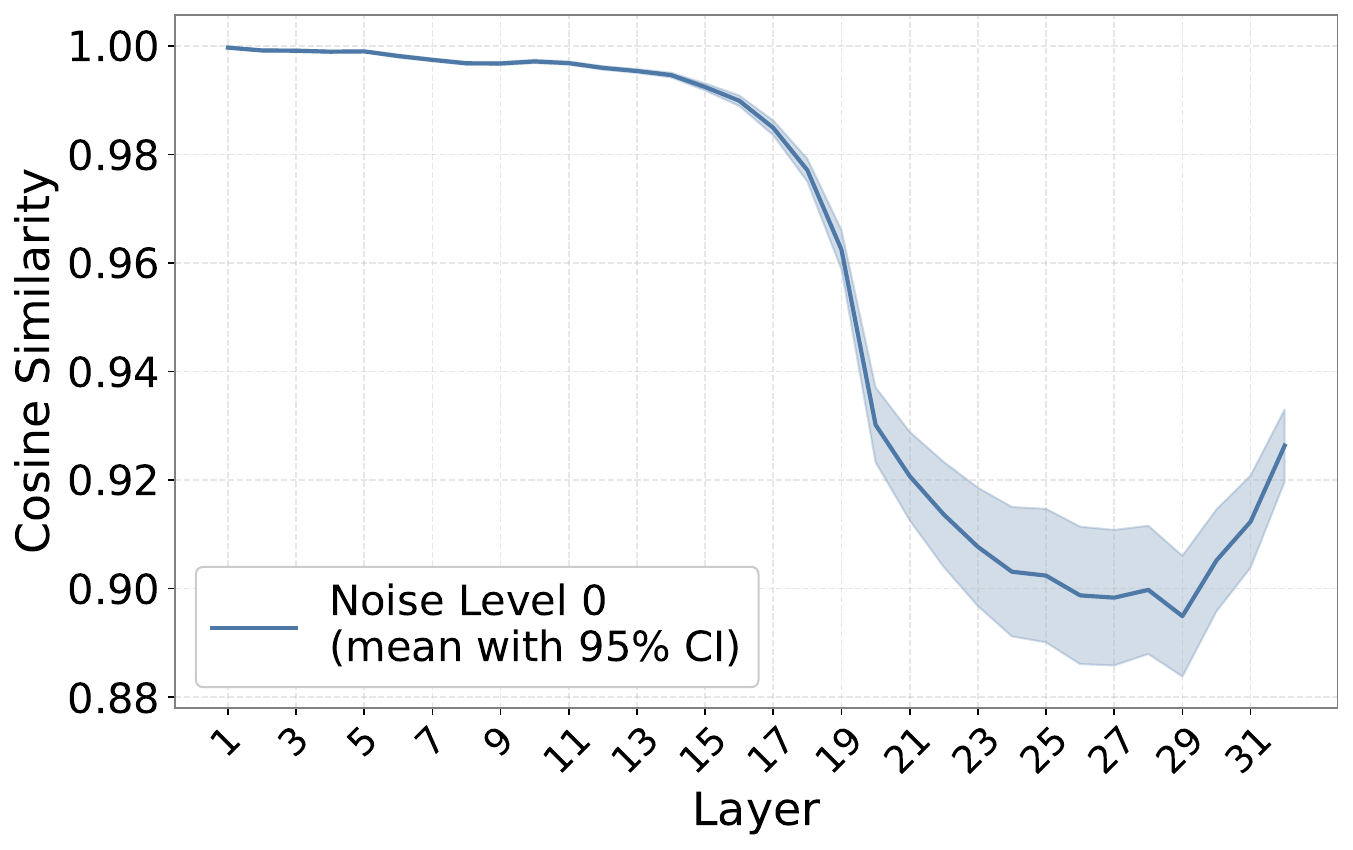}
        \caption{\small SimHidden (Smaller is better) on NQ dataset.}
    \end{subfigure}
    \hfill
    \begin{subfigure}[b]{0.49\textwidth}
        \includegraphics[width=\linewidth]{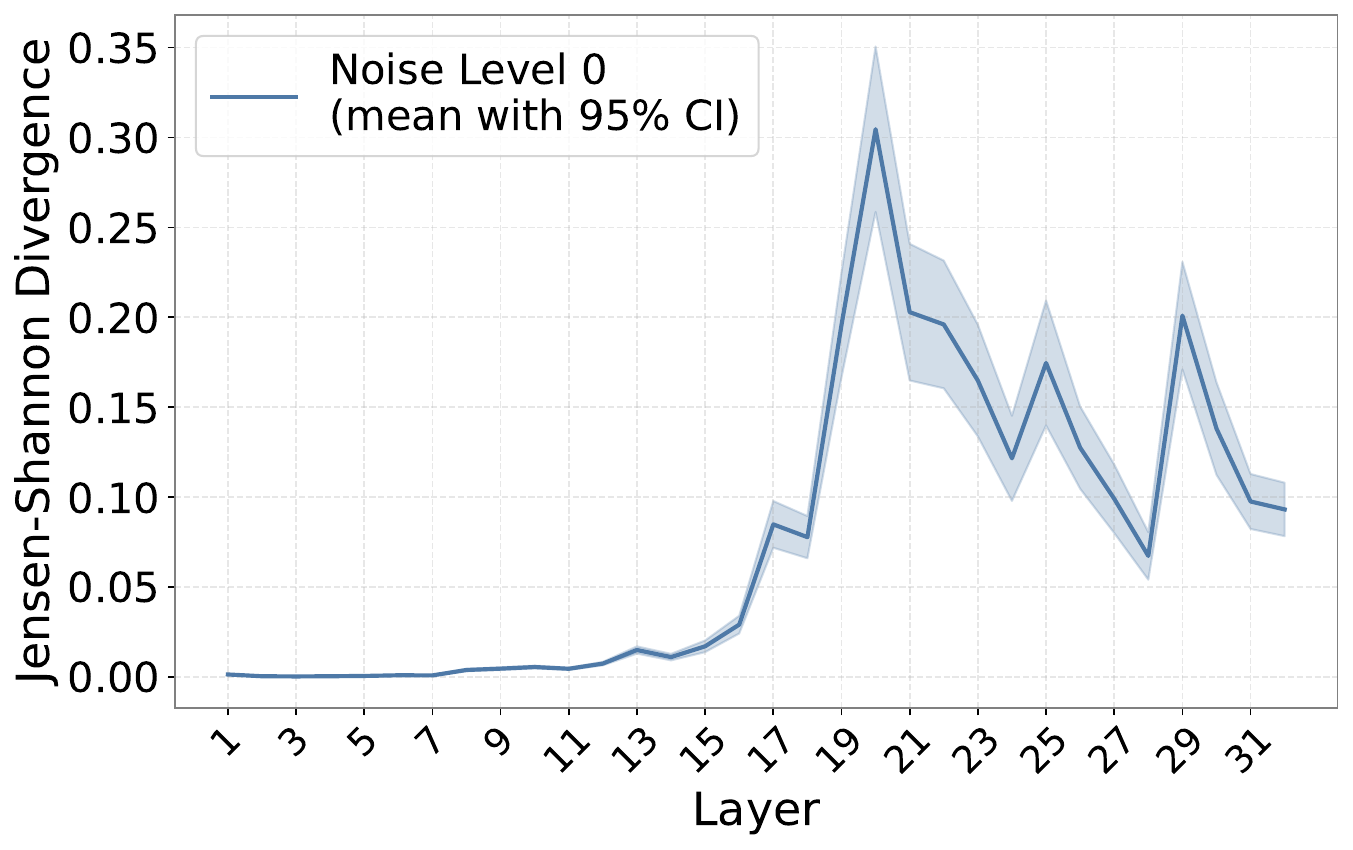}
        \caption{\small DiffAttn (Larger is better) on NQ dataset.}
    \end{subfigure}

    \vspace{\baselineskip}

    \begin{subfigure}[b]{0.49\textwidth}
        \includegraphics[width=\linewidth]{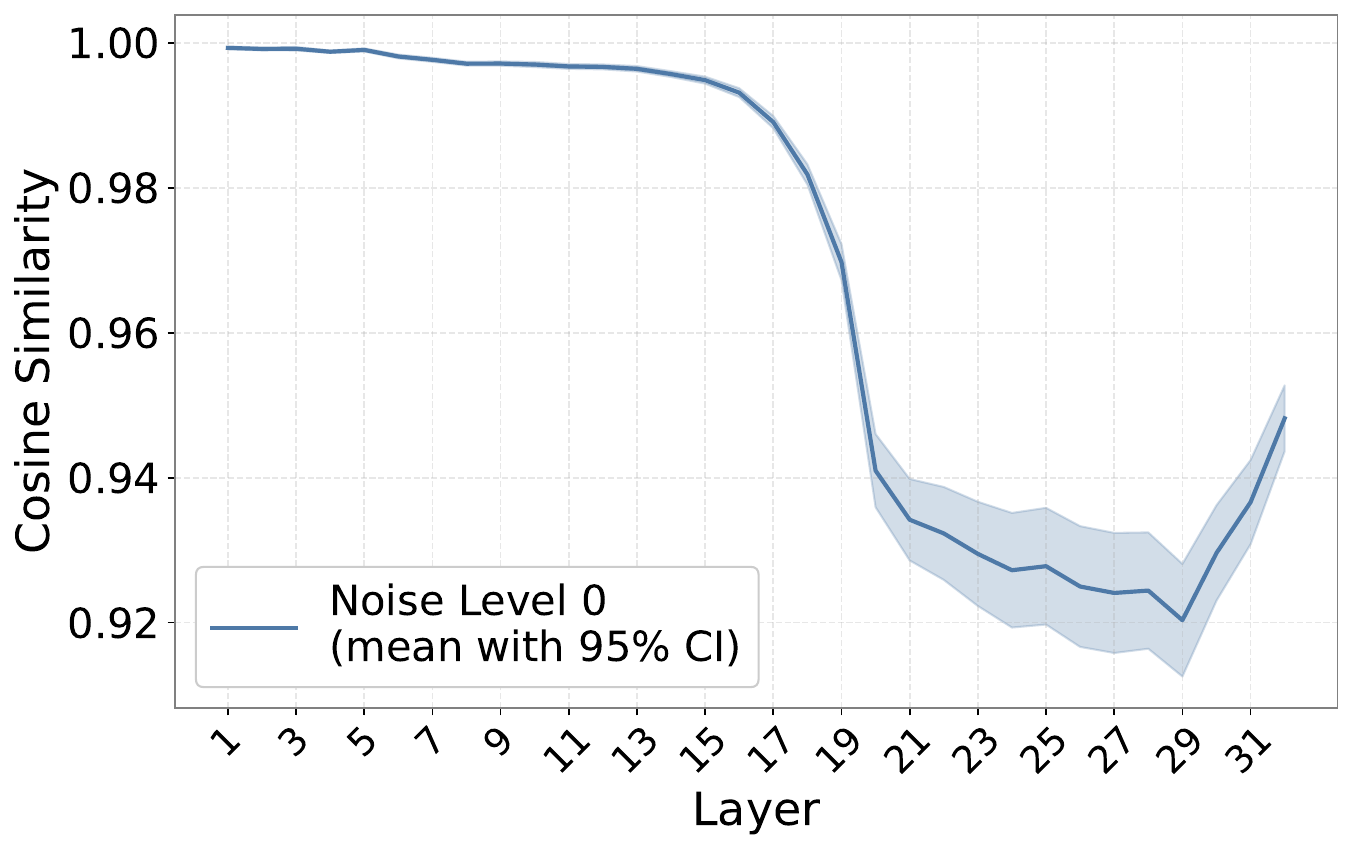}
        \caption{\small SimHidden (Smaller is better) on HotpotQA dataset.}
    \end{subfigure}
    \hfill
    \begin{subfigure}[b]{0.49\textwidth}
        \includegraphics[width=\linewidth]{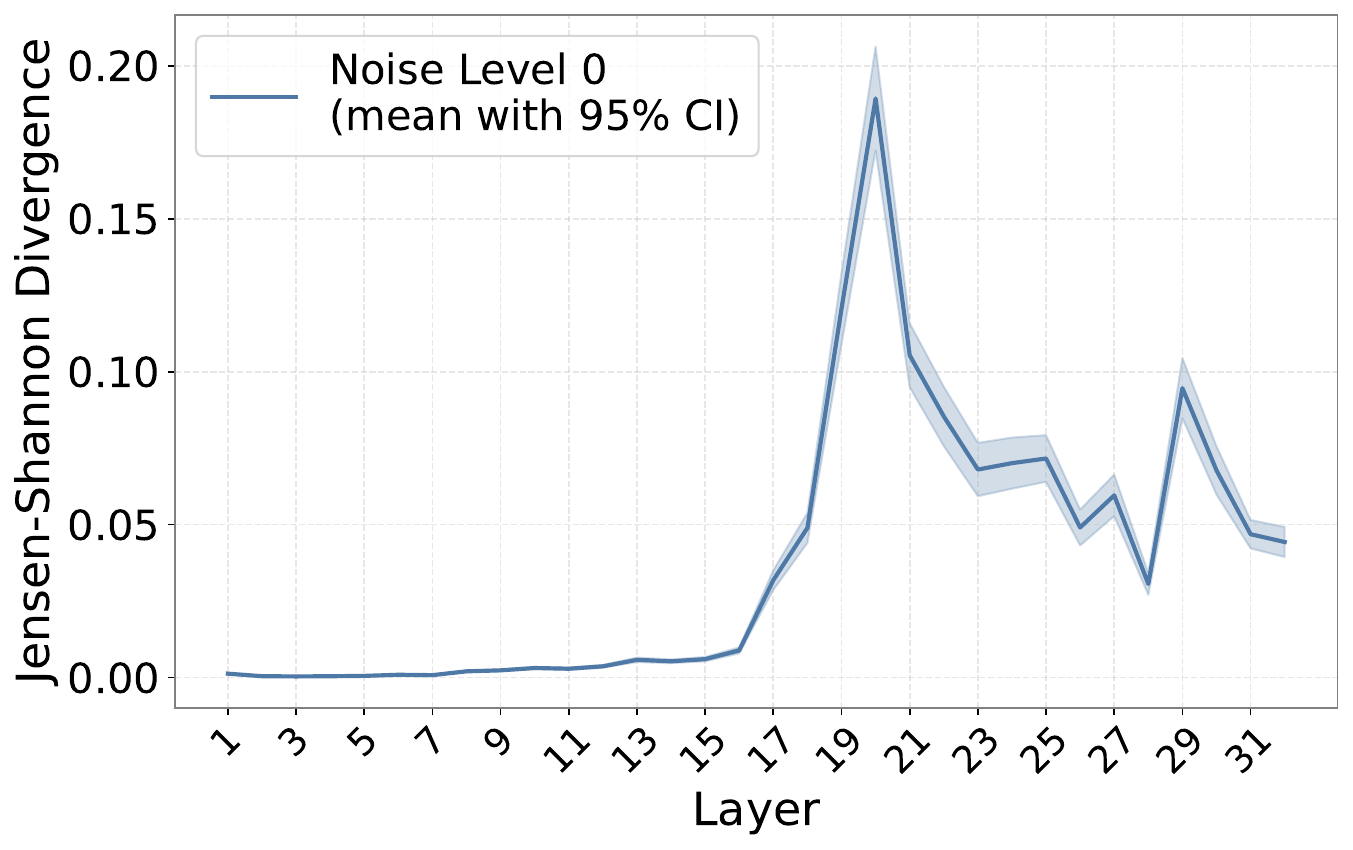}
        \caption{\small DiffAttn (Larger is better) on HotpotQA dataset.}
    \end{subfigure}
    
    \caption{(a)~Average SimHidden scores (with 95\% confidence intervals) across layers when noise level = 0;~(b)~Average DiffAttn scores (with 95\% confidence intervals) across layers when noise level = 0. Results are from Mistral-7B on NQ dataset and HotpotQA dataset.}
    \label{fig:analysis-mistral}
\end{figure}

\begin{figure}[h]
    \centering
    \begin{subfigure}[b]{0.49\textwidth}
        \includegraphics[width=\linewidth]{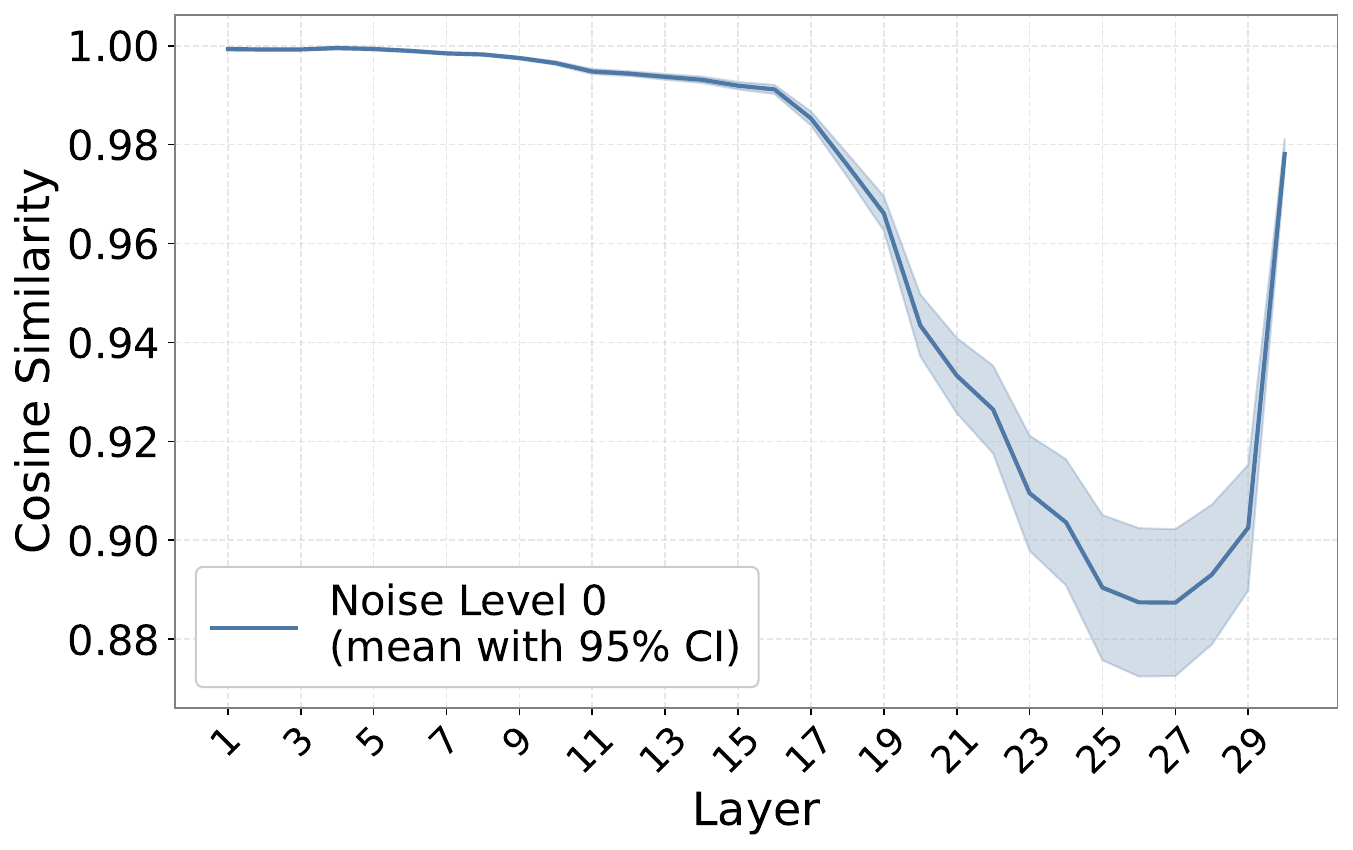}
        \caption{\small SimHidden (Smaller is better) on NQ dataset.}
    \end{subfigure}
    \hfill
    \begin{subfigure}[b]{0.49\textwidth}
        \includegraphics[width=\linewidth]{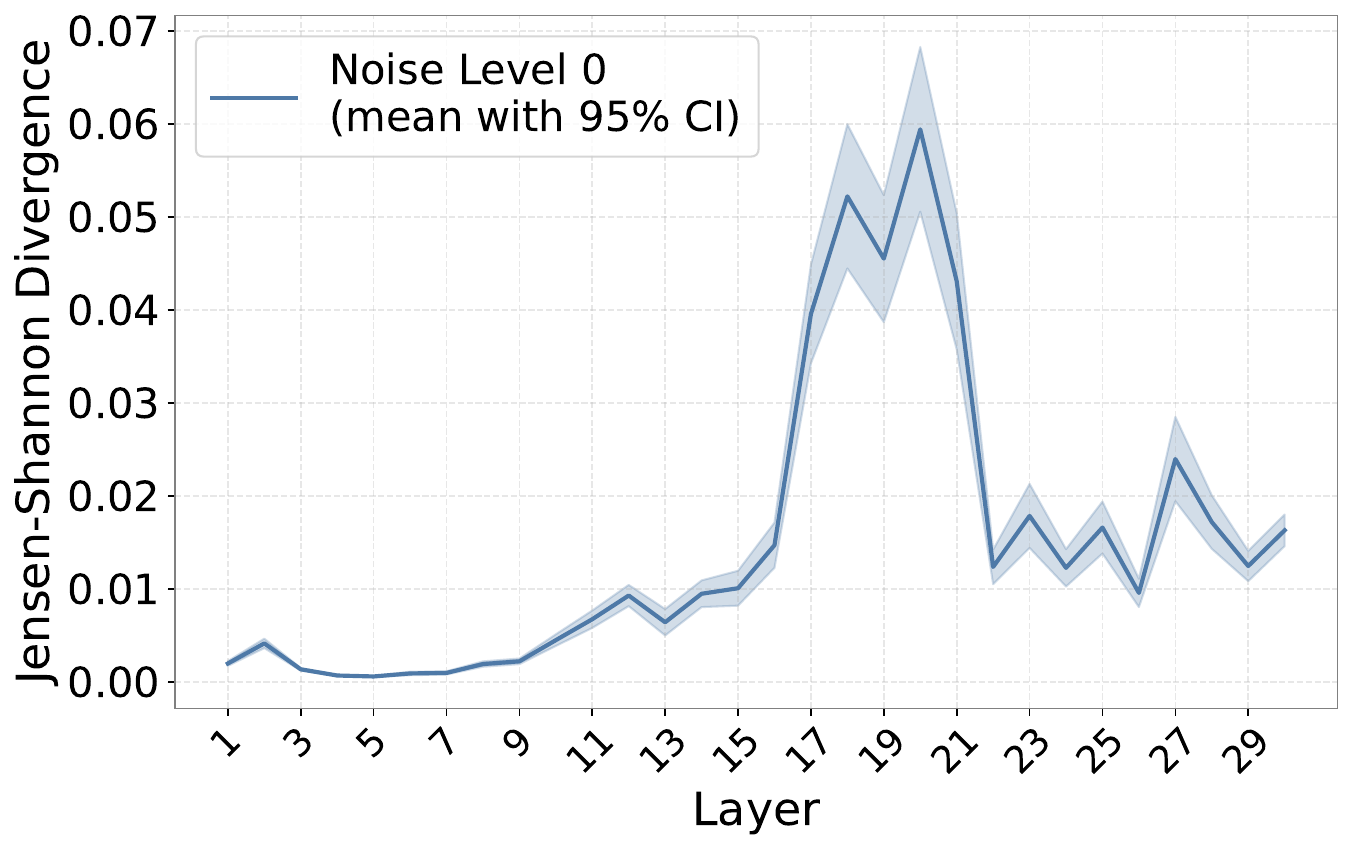}
        \caption{\small DiffAttn (Larger is better) on NQ dataset.}
    \end{subfigure}

    \vspace{\baselineskip}

    \begin{subfigure}[b]{0.49\textwidth}
        \includegraphics[width=\linewidth]{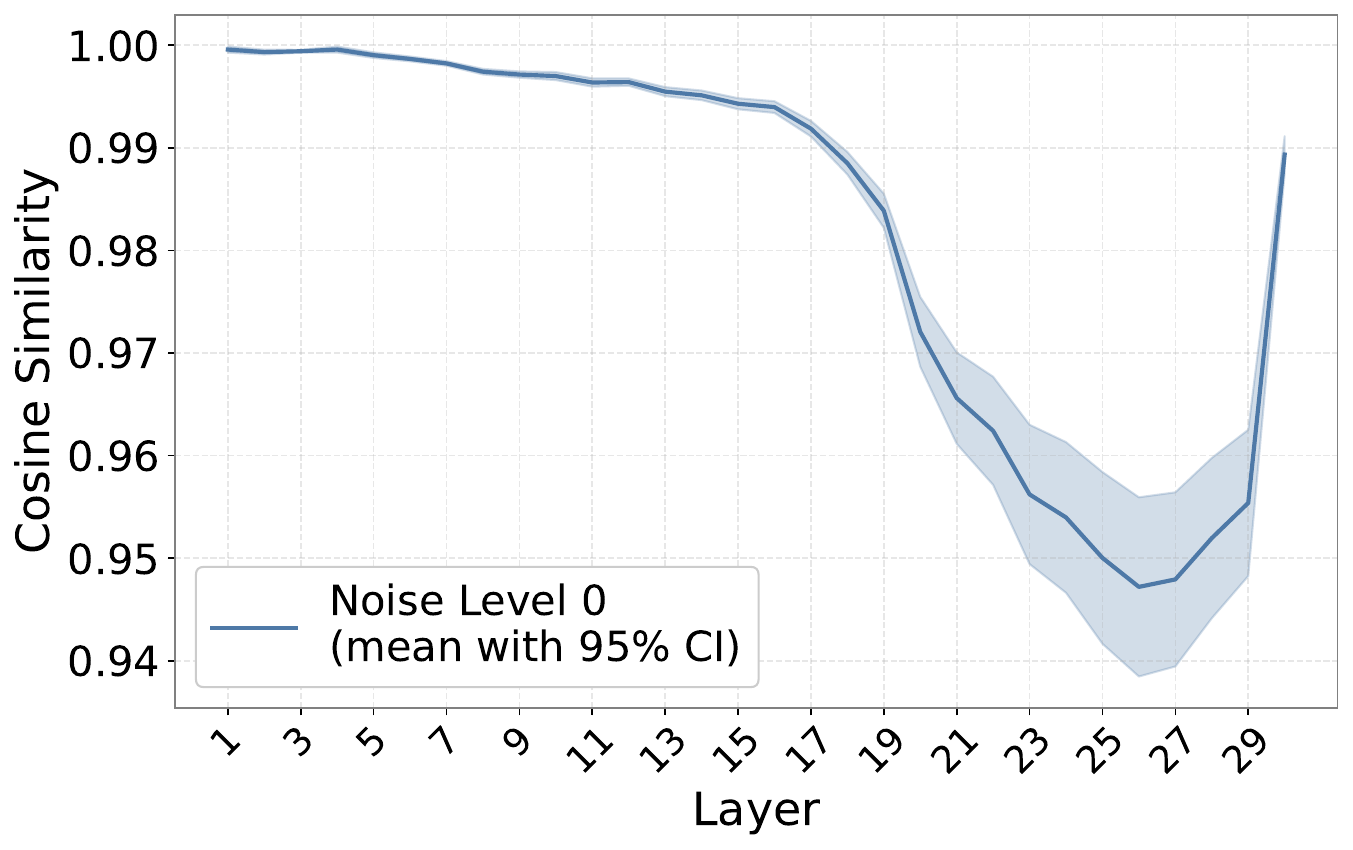}
        \caption{\small SimHidden (Smaller is better) on HotpotQA dataset.}
    \end{subfigure}
    \hfill
    \begin{subfigure}[b]{0.49\textwidth}
        \includegraphics[width=\linewidth]{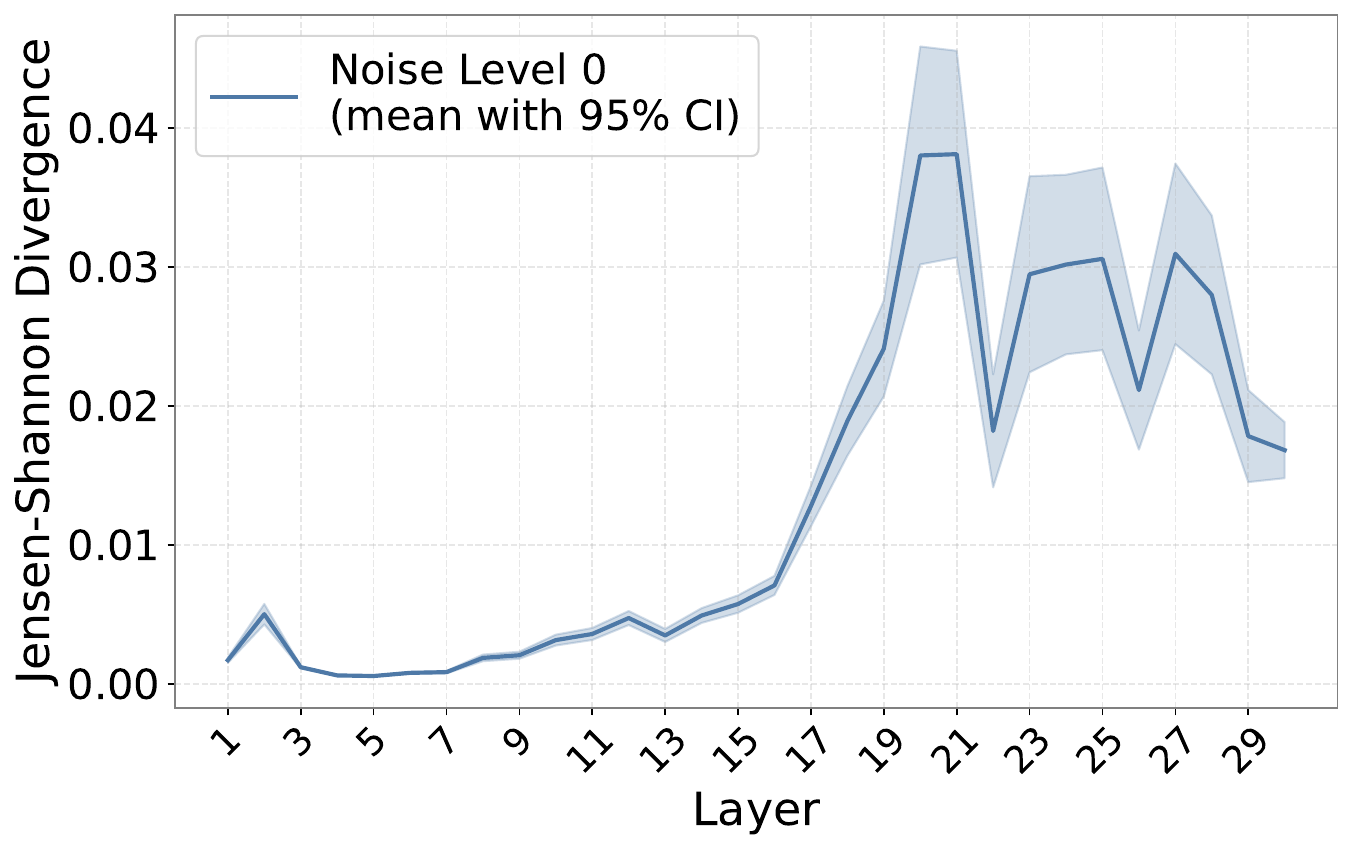}
        \caption{\small DiffAttn (Larger is better) on HotpotQA dataset.}
    \end{subfigure}
    
    \caption{(a)~Average SimHidden scores (with 95\% confidence intervals) across layers when noise level = 0;~(b)~Average DiffAttn scores (with 95\% confidence intervals) across layers when noise level = 0. Results are from DeepSeek-7B on NQ dataset and HotpotQA dataset.}
    \label{fig:analysis-deepseek}
\end{figure}

\begin{figure}[h]
    \centering
    \begin{subfigure}[b]{0.49\textwidth}
        \includegraphics[width=\linewidth]{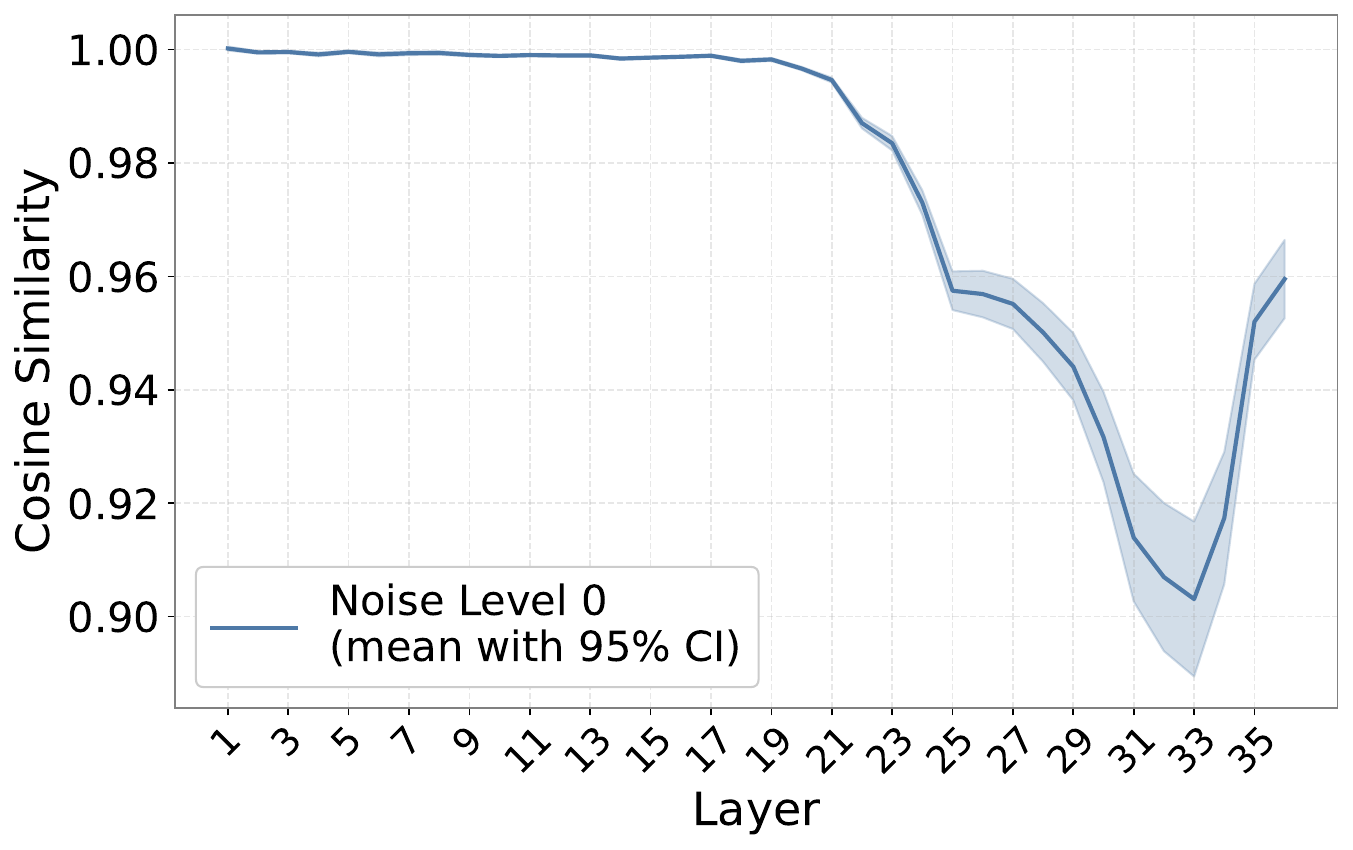}
        \caption{\small SimHidden (Smaller is better) on NQ dataset.}
    \end{subfigure}
    \hfill
    \begin{subfigure}[b]{0.49\textwidth}
        \includegraphics[width=\linewidth]{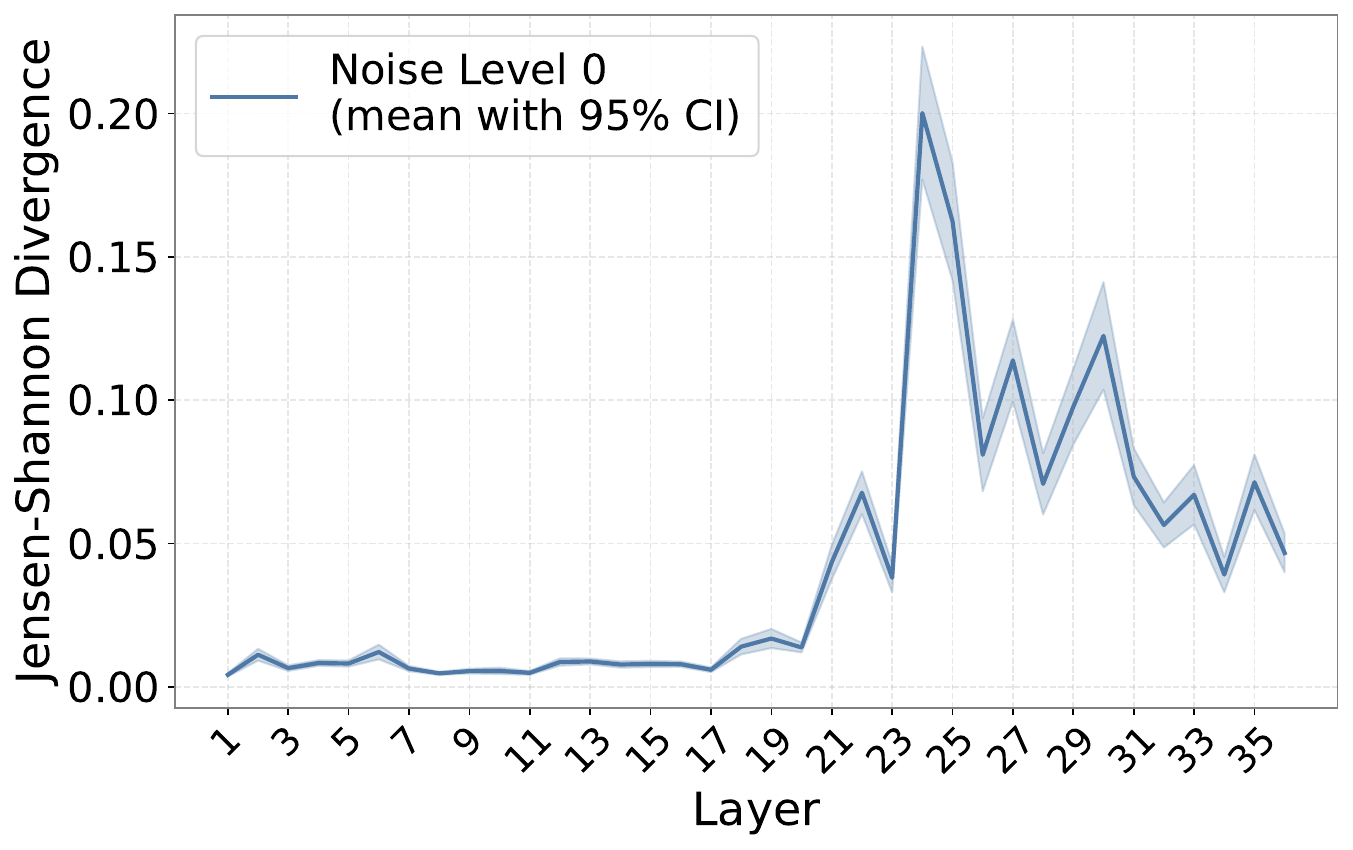}
        \caption{\small DiffAttn (Larger is better) on NQ dataset.}
    \end{subfigure}

    \vspace{\baselineskip}

    \begin{subfigure}[b]{0.49\textwidth}
        \includegraphics[width=\linewidth]{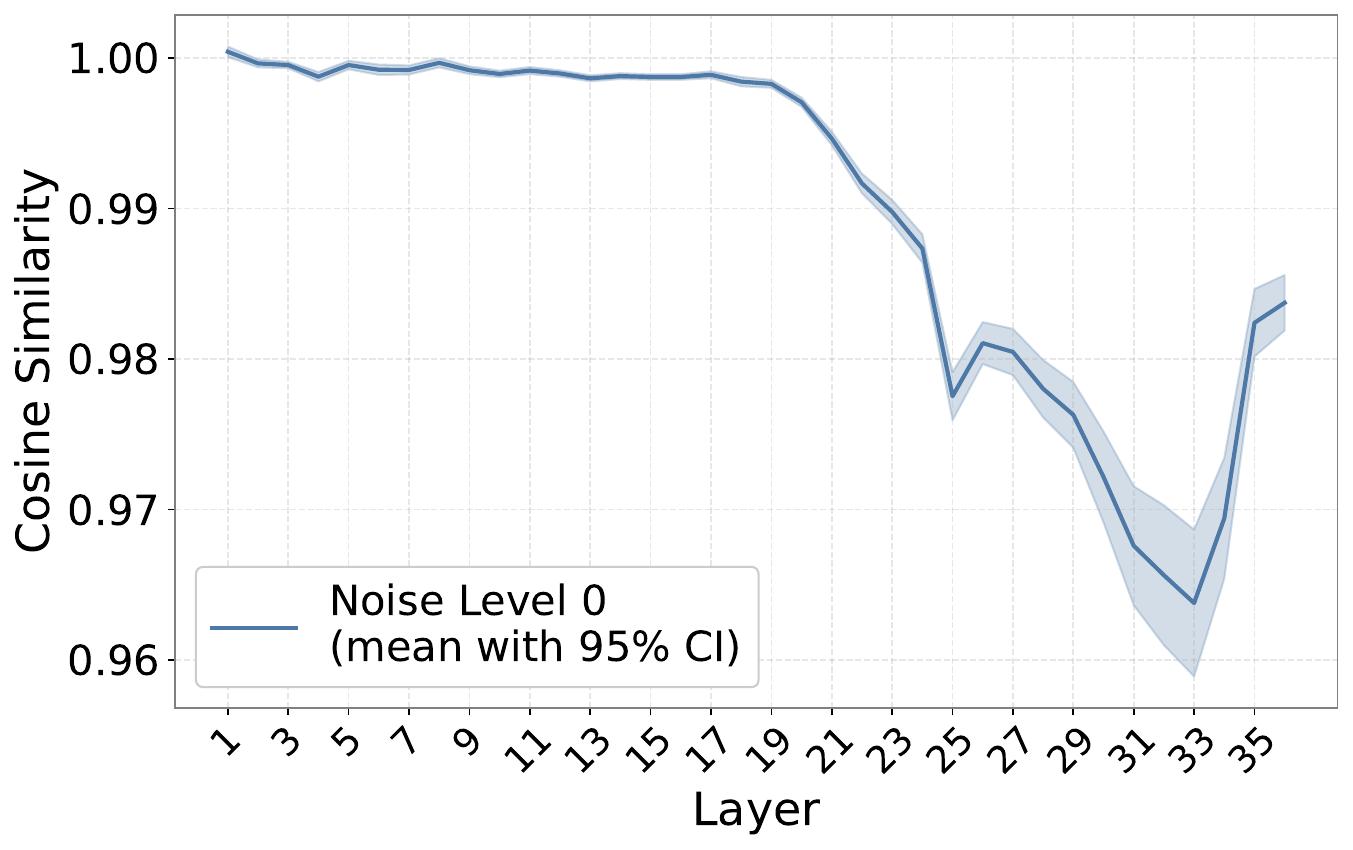}
        \caption{\small SimHidden (Smaller is better) on HotpotQA dataset.}
    \end{subfigure}
    \hfill
    \begin{subfigure}[b]{0.49\textwidth}
        \includegraphics[width=\linewidth]{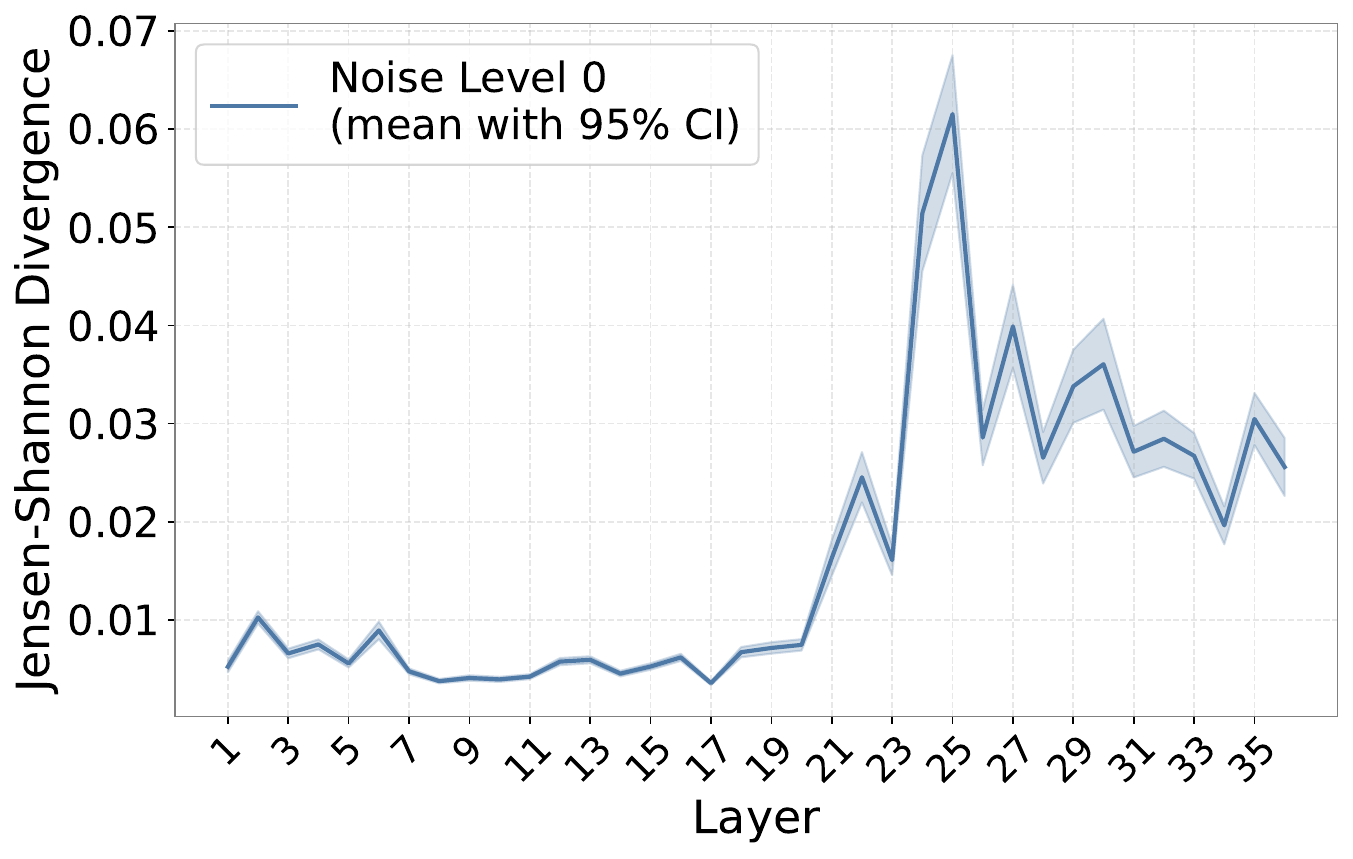}
        \caption{\small DiffAttn (Larger is better) on HotpotQA dataset.}
    \end{subfigure}
    
    \caption{(a)~Average SimHidden scores (with 95\% confidence intervals) across layers when noise level = 0;~(b)~Average DiffAttn scores (with 95\% confidence intervals) across layers when noise level = 0. Results are from Qwen3-8B on NQ dataset and HotpotQA dataset.}
    \label{fig:analysis-qwen}
\end{figure}

To assess the generality of this phenomenon, we evaluate three additional models: Mistral-7B, DeepSeek-7B, and Qwen3-8B, on both the NQ and HotpotQA datasets.
Results are shown in Figures~\ref{fig:analysis-mistral}--\ref{fig:analysis-qwen}. Despite architectural differences, all models exhibit similar three-phase trends in SimHidden scores. Specifically, the boundaries of the early, middle, and deeper layers are as follows: Mistral-7B (1--14, 15--29, 30--32), DeepSeek-7B (1--16, 17--27, 28--30), and Qwen3-8B (1--19, 20--33, 34--36).Correspondingly, the peak DiffAttn scores occur in the middle layers, at layer 20 for both Mistral-7B and DeepSeek-7B, and at layer 24 for Qwen3-8B.

\section{Noise Injection Analysis for CS, DoLA, and LFD}\label{appendix-c}

To investigate the performance of different decoding strategies under varying noise conditions, we introduce controlled noise levels (4, 8, and 12) to CS, DoLA, and LFD. Experiments are conducted using LLaMA2-7B and Mistral-7B on the NQ and HotpotQA datasets, respectively.
As shown in Table~\ref{tab:main-noise-llama} and Table~\ref{tab:main-noise-mistral}, LFD generally achieves higher accuracy than other decoding methods under various noise levels, indicating comparatively stronger robustness to noise.
Notably, a significant improvement in LFD's accuracy is observed under moderate noise conditions. For example, on the NQ dataset using LLaMA2-7B, applying noise level 12 yields a 6.7\%accuracy gain compared to the noise-free setting. This suggests that controlled noise exposure can further improve the performance of LFD.

\begin{table}[t]
    \centering
    \setlength{\tabcolsep}{12pt}
    \scriptsize
    \begin{tabular}{lcccc}
        \toprule
        & \multirow{2}{*}{NQ} & \multicolumn{3}{c}{HotpotQA} \\
        \cmidrule(lr){3-5}
        & & Compare & Bridge & Total \\
        \hline \hline
         GD~(0) & 0.5745 &  \underline{0.4755} &  \underline{0.4108} &  \underline{0.4237} \\
         CS~(0) & 0.5775 & 0.4284 & 0.4049 & 0.4096 \\
         DoLA~(0) & \underline{0.5809} & 0.4438 & 0.3883 & 0.3995 \\
         LFD~(0) & \textbf{0.5949} & \textbf{0.5453} & \textbf{0.4295} & \textbf{0.4528} \\
        \midrule
         GD~(4) &  0.5559 & \textbf{0.5084} & 0.4049 & 0.4257\\
         CS~(4) &  \textbf{0.5657} & 0.4546 & \underline{0.4199} & \underline{0.4269} \\
         DoLA~(4) & 0.5574 & 0.4775 & 0.4111 & 0.4244 \\
         LFD~(4) & \underline{0.5637} & \underline{0.5071} & \textbf{0.4304} & \textbf{0.4458} \\
        \midrule
         GD~(8) & 0.5965 & \underline{0.5091} & 0.4096 & 0.4317 \\
         CS~(8) & \underline{0.6054} & 0.4694 & \underline{0.4236} & 0.4328 \\
         DoLA~(8) & 0.6023 & 0.4956 & 0.4199 & \underline{0.4351} \\
         LFD~(8) & \textbf{0.6216} & \textbf{0.5400} & \textbf{0.4400} & \textbf{0.4601} \\
        \midrule
         GD~(12) & 0.6383 & \underline{0.5286} & 0.4236 & 0.4447 \\
         CS~(12) & 0.6410 & 0.4781 & \underline{0.4419} & \underline{0.4492} \\
         DoLA~(12) & \underline{0.6458} & 0.5158 & 0.4306 & 0.4477 \\
         LFD~(12) & \textbf{0.6622} & \textbf{0.5521} & \textbf{0.4532} & \textbf{0.4731} \\
        \bottomrule
    \end{tabular}
    \caption{Accuracy performance comparison of different decoding methods with varying levels of noise, evaluated on the NQ and HotpotQA datasets using the LLaMA2-7B. \textbf{Bold} values indicate the best performance, while \underline{underlined} values represent the second-best.}
    \label{tab:main-noise-llama}
\end{table}

\begin{table}[t]
    \centering
    \setlength{\tabcolsep}{12pt}
    \scriptsize
    \begin{tabular}{lcccc}
        \toprule
        & \multirow{2}{*}{NQ} & \multicolumn{3}{c}{HotpotQA} \\
        \cmidrule(lr){3-5}
        & & Compare & Bridge & Total \\
        \hline \hline
         GD~(0) & 0.6130 & 0.5864 & \underline{0.5889} & \underline{0.5884} \\
         CS~(0) & 0.5598 & 0.4149 & 0.5117 & 0.4922 \\
         DoLA~(0) & \underline{0.6142} & \underline{0.5931} & 0.5870 & 0.5883 \\
         LFD~(0) & \textbf{0.6357} & \textbf{0.6026} & \textbf{0.6093} & \textbf{0.6079} \\
        \midrule
         GD~(4) & \underline{0.5667} & \textbf{0.5494} & \underline{0.5348} & \underline{0.5377} \\
         CS~(4) & 0.5014 & 0.2065 & 0.2778 & 0.2635 \\
         DoLA~(4) & 0.5625 & 0.5373 & 0.5331 & 0.5340 \\
         LFD~(4) & \textbf{0.6036} & \underline{0.5467} & \textbf{0.5502} & \textbf{0.5495} \\
        \midrule
         GD~(8) & \underline{0.5678} & \textbf{0.5326} & \underline{0.5260} & \underline{0.5273} \\
         CS~(8) & 0.5201 & 0.1137 & 0.1926 & 0.1768 \\
         DoLA~(8) & 0.5659 & 0.5111 & 0.5252 & 0.5223 \\
         LFD~(8) & \textbf{0.5966} & \underline{0.5259} & \textbf{0.5446} & \textbf{0.5409} \\
        \midrule
         GD~(12) & 0.5814 & \textbf{0.5440} & 0.5461 & \underline{0.5457} \\
         CS~(12) & 0.5334 & 0.1432 & 0.2328 & 0.2149 \\
         DoLA~(12) & \underline{0.5845} & 0.5293 & \underline{0.5492} & 0.5452 \\
         LFD~(12) & \textbf{0.5943} & \underline{0.5346} & \textbf{0.5640} & \textbf{0.5581} \\
        \bottomrule
    \end{tabular}
    \caption{Accuracy performance comparison of different decoding methods with varying levels of noise, evaluated on the NQ and HotpotQA datasets using the Mistral-7B. \textbf{Bold} values indicate the best performance, while \underline{underlined} values represent the second-best.}
    \label{tab:main-noise-mistral}
\end{table}

\section{Ablation Study on Layer Selection Range with DeepSeek-7B and Qwen3-8B}\label{appendix-d}

\begin{table}[h]
    \caption{Comparison of accuracy between different layer selection ranges under dynamic layer selection strategy. \textbf{Bold} indicate the best performance, while \underline{underline} represent the second-best.}
    \centering
    \setlength{\tabcolsep}{4pt}
    \scriptsize
    \begin{tabular}{llcccccccccc}
        \toprule
        & & \multirow{3}{*}{NQ} & \multirow{2}{*}{RGB} & \multicolumn{3}{c}{HotpotQA} & \multicolumn{5}{c}{2WikiMultihopQA} \\
        \cmidrule(lr){5-7} \cmidrule(lr){8-12}
        & & & & Compare & Bridge & Total & Compare & Bridge & Inf & Compose & Total \\
        \hline\hline
        \multirow{4}{*}{DeepSeek-7B}
        & GD~(0) & \underline{0.5250} & \underline{0.8233} & \underline{0.3282} & \underline{0.4033} & \underline{0.3883} & \underline{0.2625} & \underline{0.2414} & \underline{0.2663} & \underline{0.2609} & \underline{0.2609} \\
        & LFD[0, 15) & 0.3865 & 0.5000 & 0.2313 & 0.3366 & 0.3154 & 0.2155 & 0.2316 & 0.2334 & 0.2152 & 0.2211 \\
        & LFD[0, 30) & 0.5389 & 0.8233 & 0.4768 & 0.4395 & 0.4470 & 0.4432 & \textbf{0.4277} & 0.4135 & 0.4159 & 0.4248 \\
        & LFD[15, 30) & \textbf{0.5412} & \textbf{0.8267} & \textbf{0.4801} & \textbf{0.4466} & \textbf{0.4533} & \textbf{0.4492} & 0.4227 & \textbf{0.4194} & \textbf{0.4223} & \textbf{0.4285} \\
        \midrule
        \multirow{4}{*}{Qwen3-8B}
        & GD~(0) & \underline{0.7318} & \underline{0.9571} & 0.6960 & 0.6708 & 0.6759 & 0.6637 & \underline{0.6335} & 0.5897 & 0.6085 & 0.6250 \\
        & LFD[0, 18) & 0.7304 & 0.9567 & \underline{0.7068} & \underline{0.6845} & \underline{0.6889} & \underline{0.6640} & 0.6334 & \underline{0.5903} & \underline{0.6097} & \underline{0.6256} \\
        & LFD[0, 36) & 0.7372 & 0.9567 & \textbf{0.7196} & 0.6949 & 0.6999 & 0.6660 & \textbf{0.6349} & 0.6021 & 0.6146 & 0.6299 \\
        & LFD[18, 36) & \textbf{0.7380} & \textbf{0.9600} & 0.7182 & \textbf{0.6974} & \textbf{0.7016} & \textbf{0.6663} & 0.6342 & \textbf{0.6034} & \textbf{0.6148} & \textbf{0.6301} \\
        \bottomrule
    \end{tabular}
    \label{tab:lower-extend}
    \vspace{-2pt}
\end{table}

To complement the layer selection range analysis in Section~\ref{ablation}, we extend our experiments to include DeepSeek-7B and Qwen3-8B models across four datasets: NQ, RGB, HotpotQA, and 2WikiMultihopQA. As shown in Table~\ref{tab:lower-extend}, our findings remain consistent with those presented in Table~\ref{tab:lower}, demonstrating that selecting layers from the latter half of the model consistently yields superior performance compared to earlier ranges.

\section{Evaluating IKS Score Effectiveness Across Different Models and Datasets}\label{appendix-e}
\begin{figure*}[h]
    \centering
    \begin{subfigure}[b]{0.48\textwidth}
        \includegraphics[width=\linewidth]{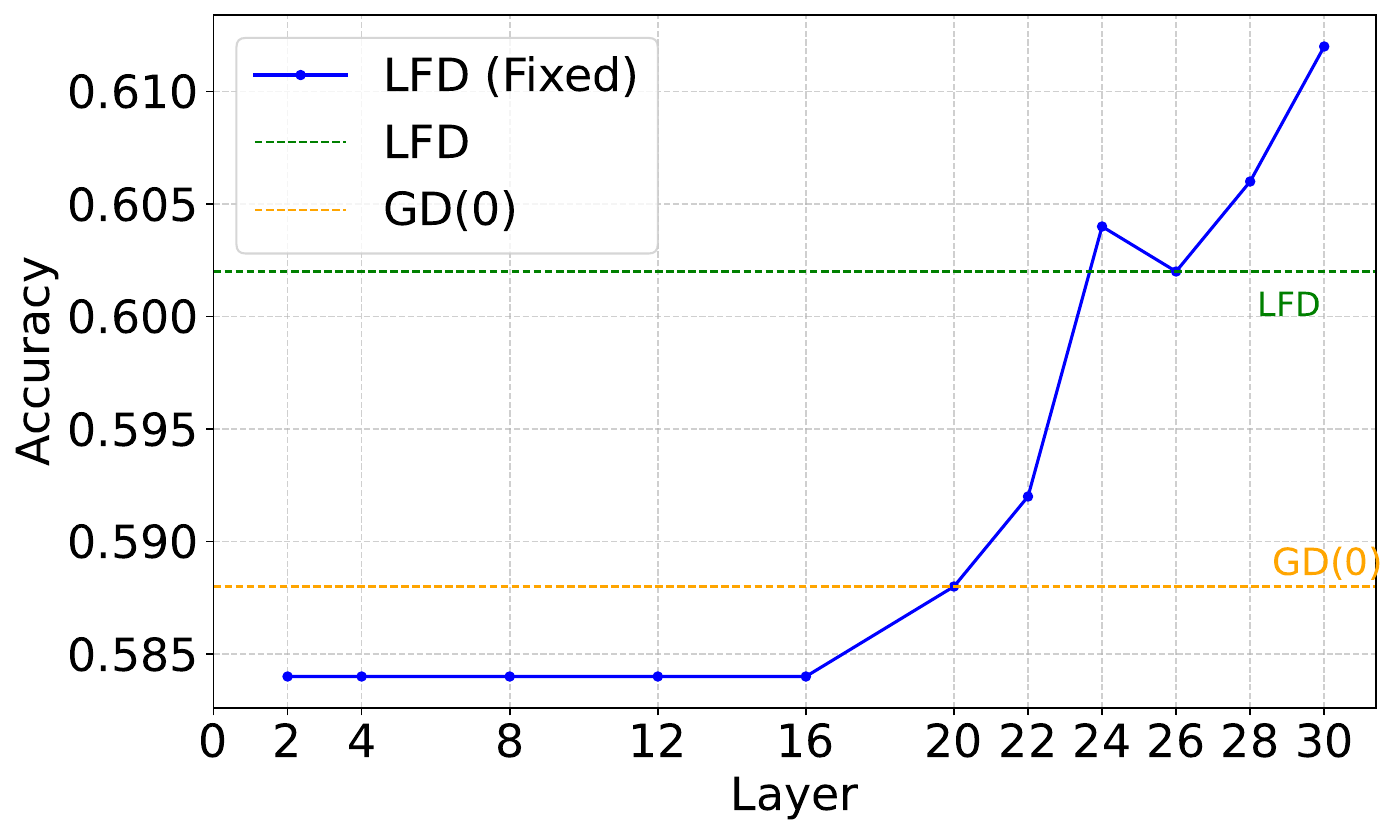}
        \caption{\small LFD vs. LFD~(Fixed) on NQ dataset.}
    \end{subfigure}
    \hfill
    \begin{subfigure}[b]{0.48\textwidth}
        \includegraphics[width=\linewidth]{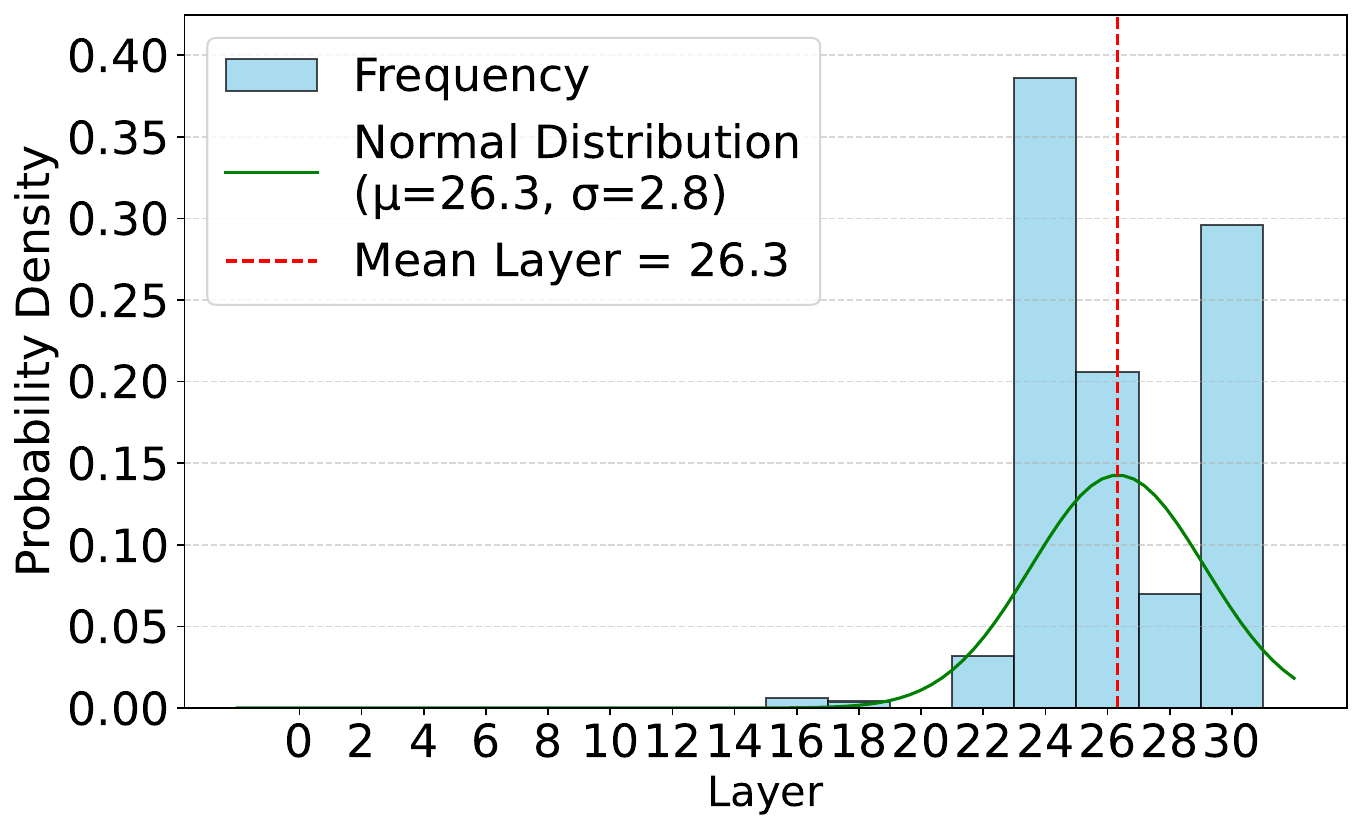}
        \caption{\small Histgram of layer selection in LFD on NQ dataset.}
    \end{subfigure}

    \vspace{\baselineskip}

    \begin{subfigure}[b]{0.48\textwidth}
        \includegraphics[width=\linewidth]{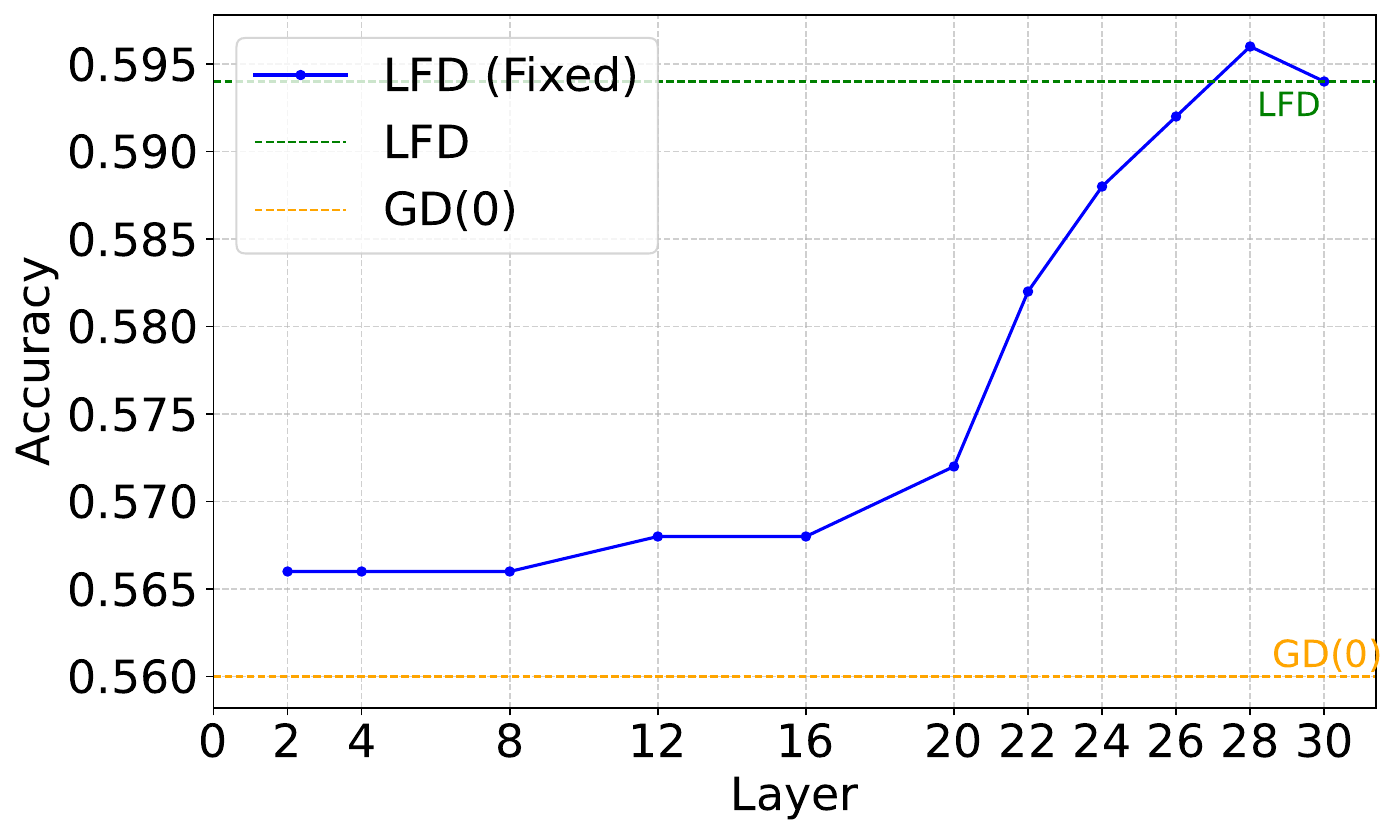}
        \caption{\small LFD vs. LFD~(Fixed) on HotpotQA dataset.}
    \end{subfigure}
    \hfill
    \begin{subfigure}[b]{0.48\textwidth}
        \includegraphics[width=\linewidth]{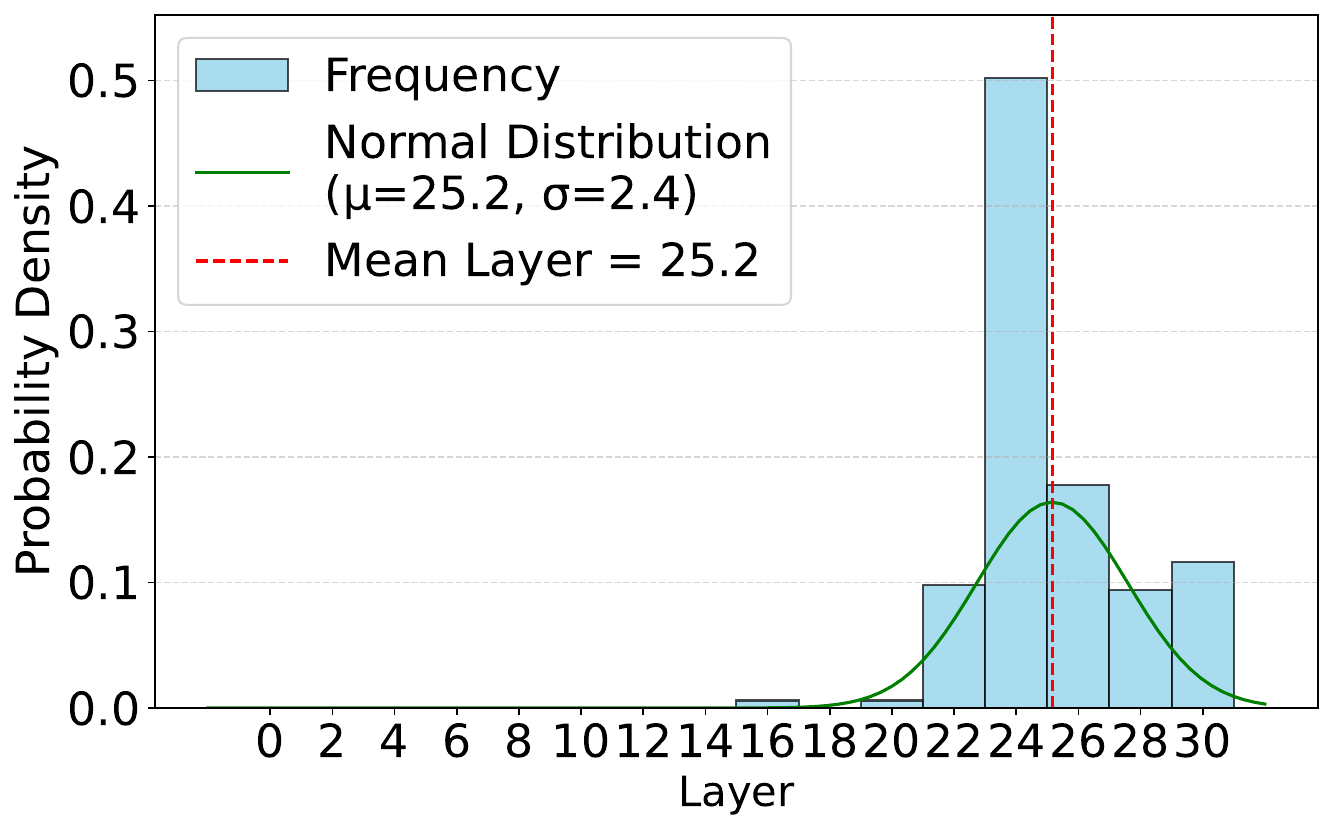}
        \caption{\small Histgram of layer selection in LFD on HotpotQA dataset.}
    \end{subfigure}
    
    \caption{Comparison between LFD and LFD (Fixed) using the Mistral-7B model on the NQ and HotpotQA datasets.}
    \label{fig:fixed-mistral}
\end{figure*}

\begin{figure*}[h]
    \centering
    \begin{subfigure}[b]{0.48\textwidth}
        \includegraphics[width=\linewidth]{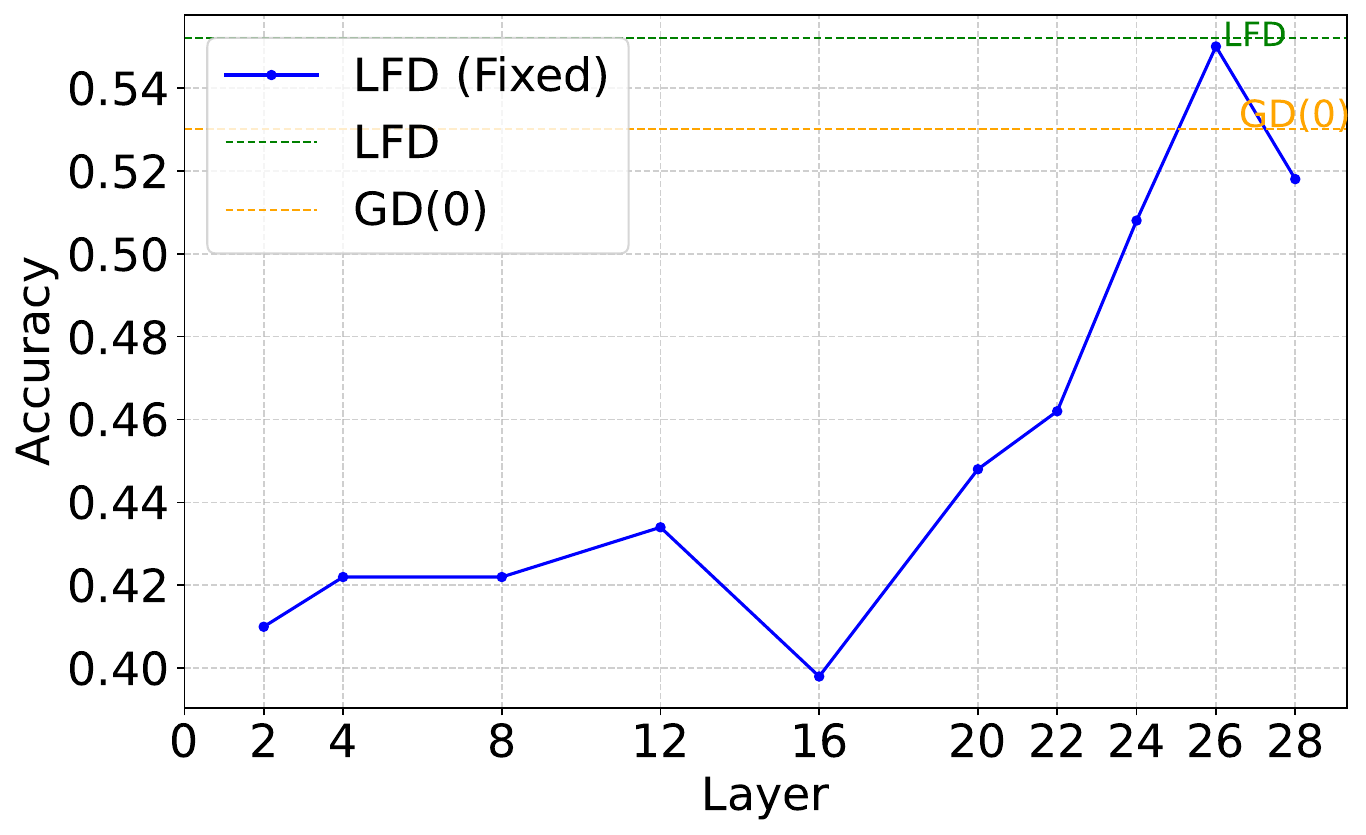}
        \caption{\small LFD vs. LFD~(Fixed) on NQ dataset.}
    \end{subfigure}
    \hfill
    \begin{subfigure}[b]{0.48\textwidth}
        \includegraphics[width=\linewidth]{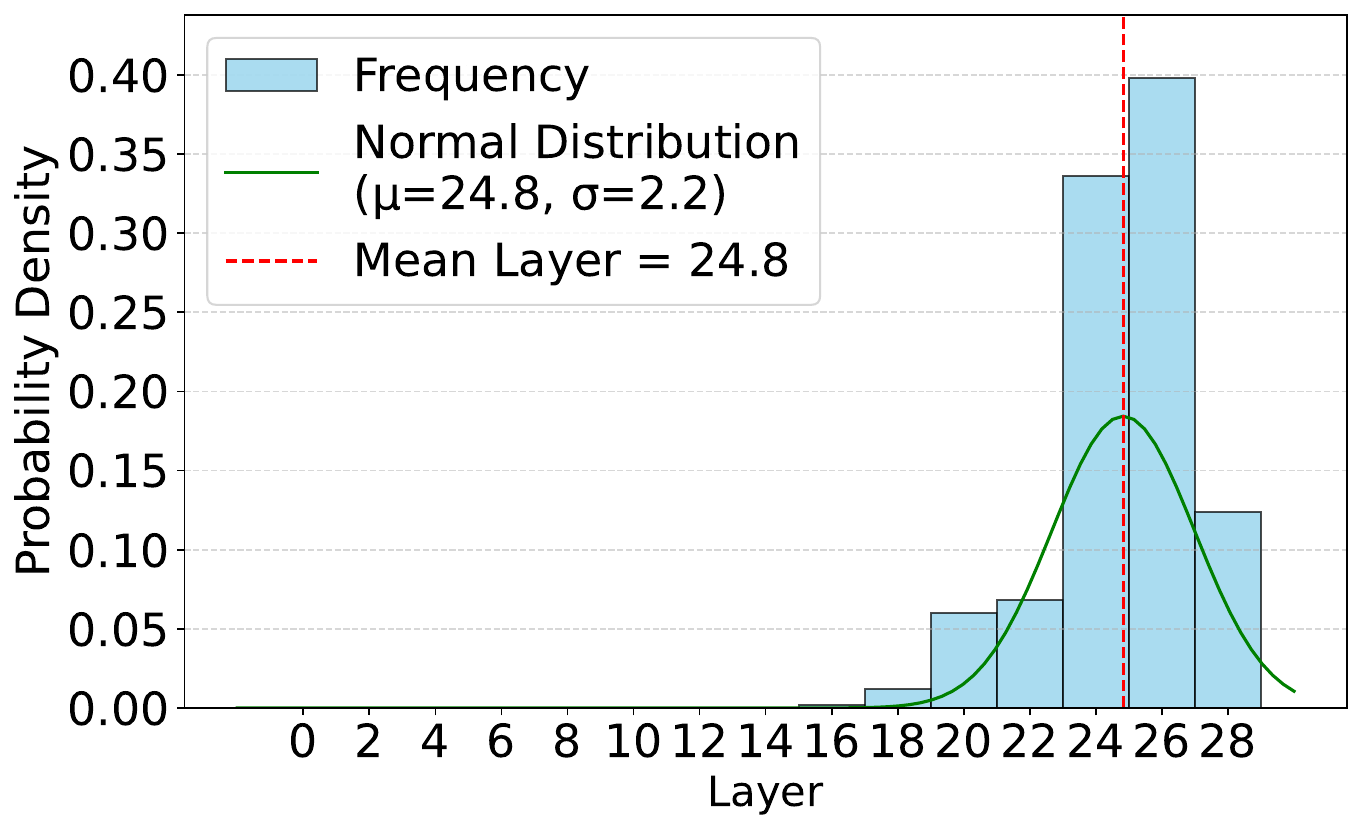}
        \caption{\small Histgram of layer selection in LFD on NQ dataset.}
    \end{subfigure}

    \vspace{\baselineskip}

    \begin{subfigure}[b]{0.48\textwidth}
        \includegraphics[width=\linewidth]{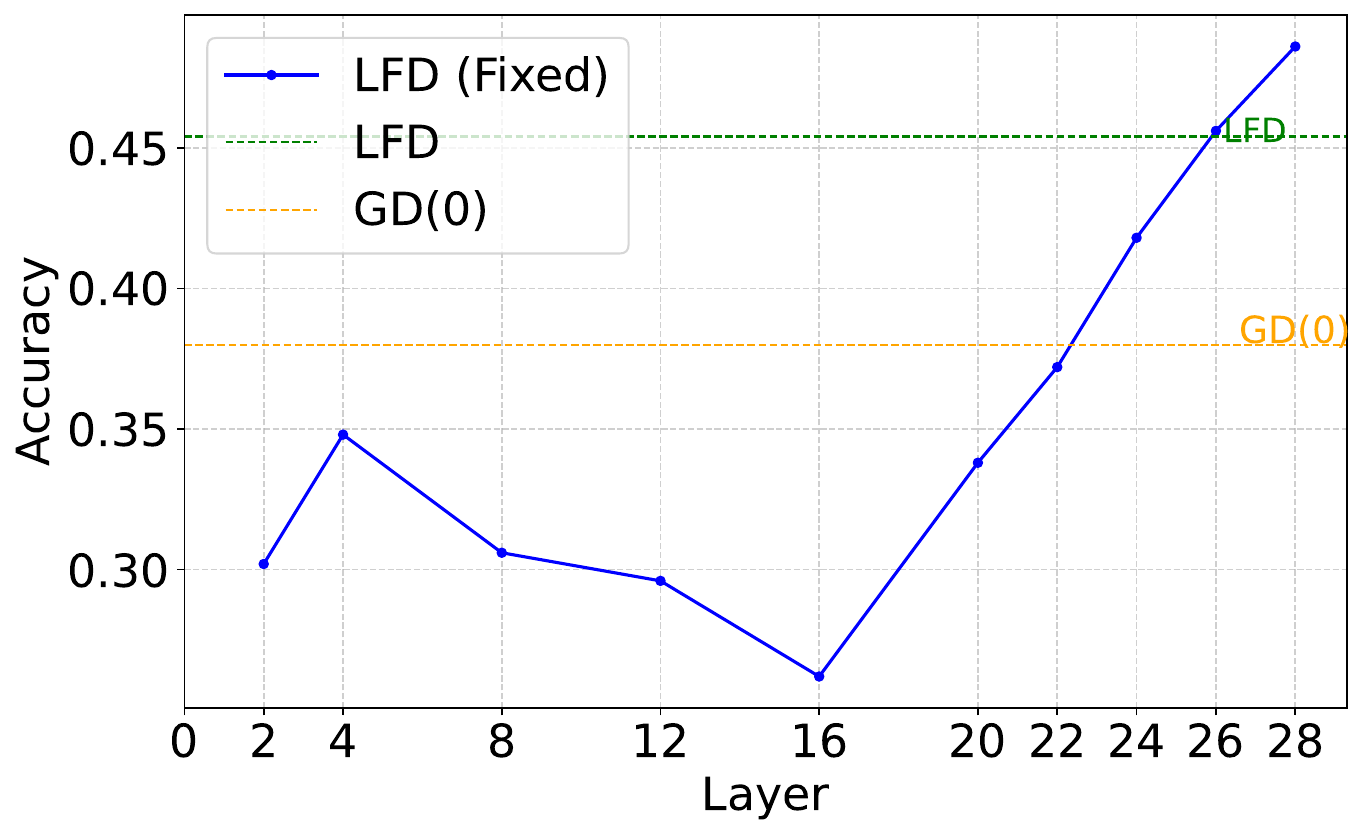}
        \caption{\small LFD vs. LFD~(Fixed) on HotpotQA dataset.}
    \end{subfigure}
    \hfill
    \begin{subfigure}[b]{0.48\textwidth}
        \includegraphics[width=\linewidth]{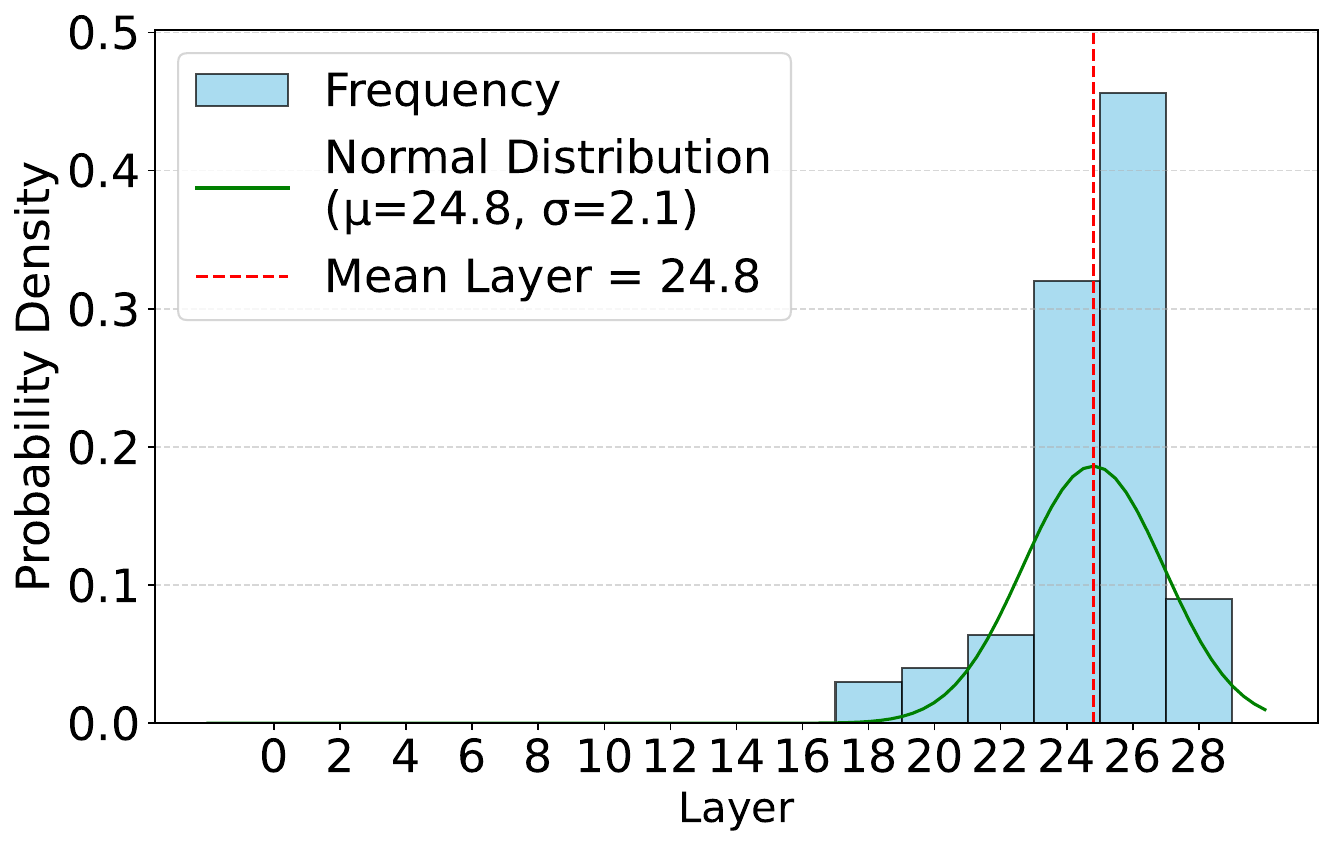}
        \caption{\small Histgram of layer selection in LFD on HotpotQA dataset.}
    \end{subfigure}
    
    \caption{Comparison between LFD and LFD (Fixed) using the DeepSeek-7B model on the NQ and HotpotQA datasets.}
    \label{fig:fixed-deepseek}
\end{figure*}

\begin{figure*}[h]
    \centering
    \begin{subfigure}[b]{0.48\textwidth}
        \includegraphics[width=\linewidth]{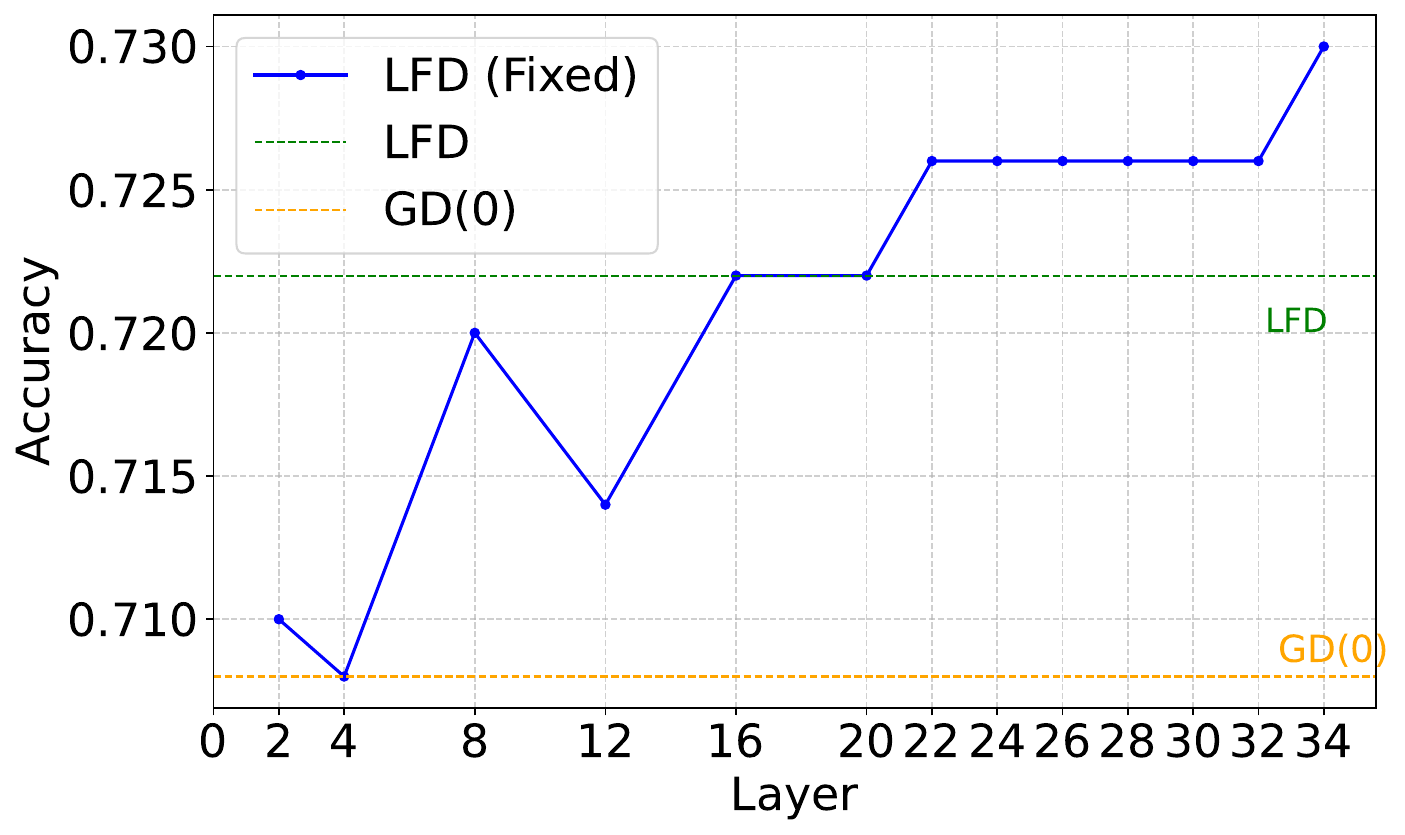}
        \caption{\small LFD vs. LFD~(Fixed) on NQ dataset.}
    \end{subfigure}
    \hfill
    \begin{subfigure}[b]{0.48\textwidth}
        \includegraphics[width=\linewidth]{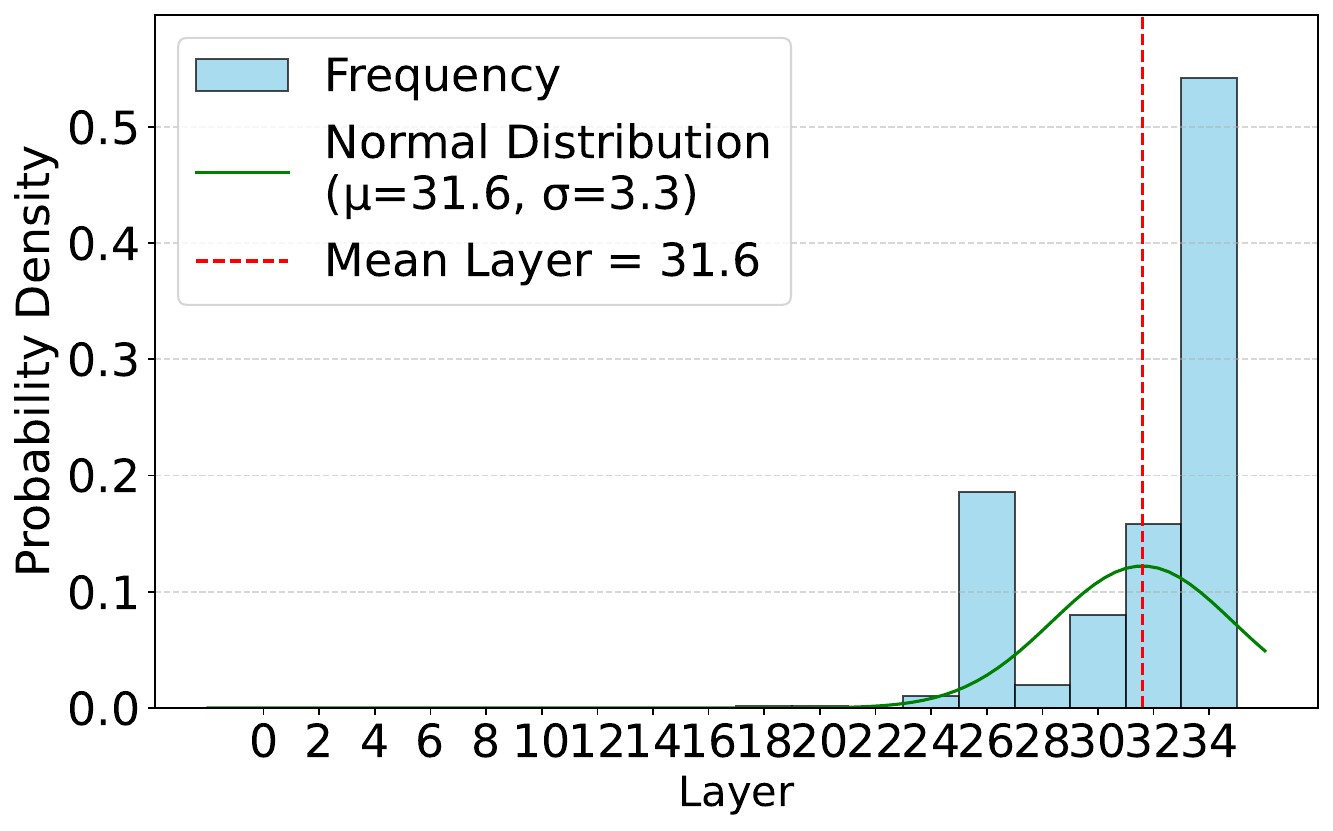}
        \caption{\small Histgram of layer selection in LFD on NQ dataset.}
    \end{subfigure}

    \vspace{\baselineskip}

    \begin{subfigure}[b]{0.48\textwidth}
        \includegraphics[width=\linewidth]{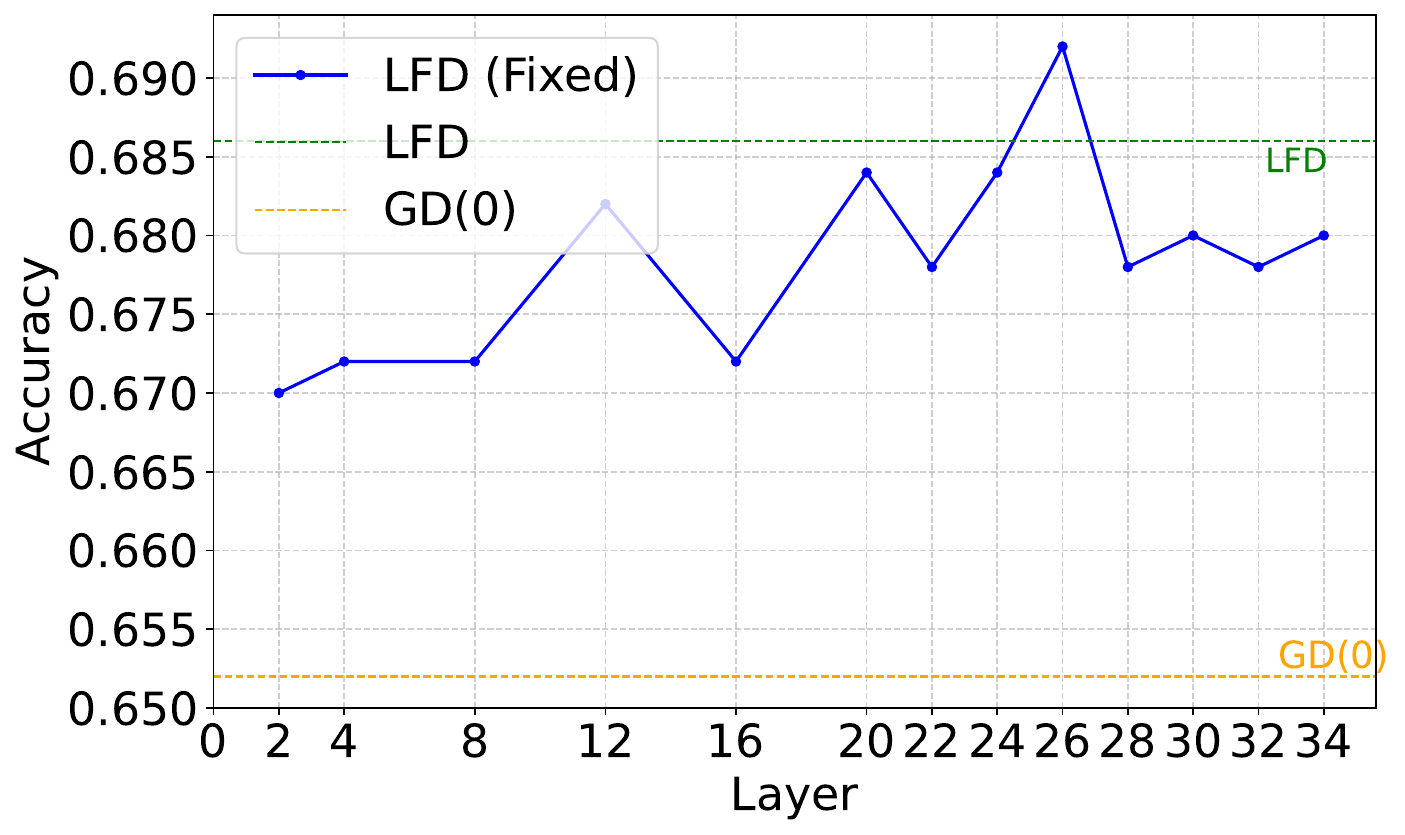}
        \caption{\small LFD vs. LFD~(Fixed) on HotpotQA dataset.}
    \end{subfigure}
    \hfill
    \begin{subfigure}[b]{0.48\textwidth}
        \includegraphics[width=\linewidth]{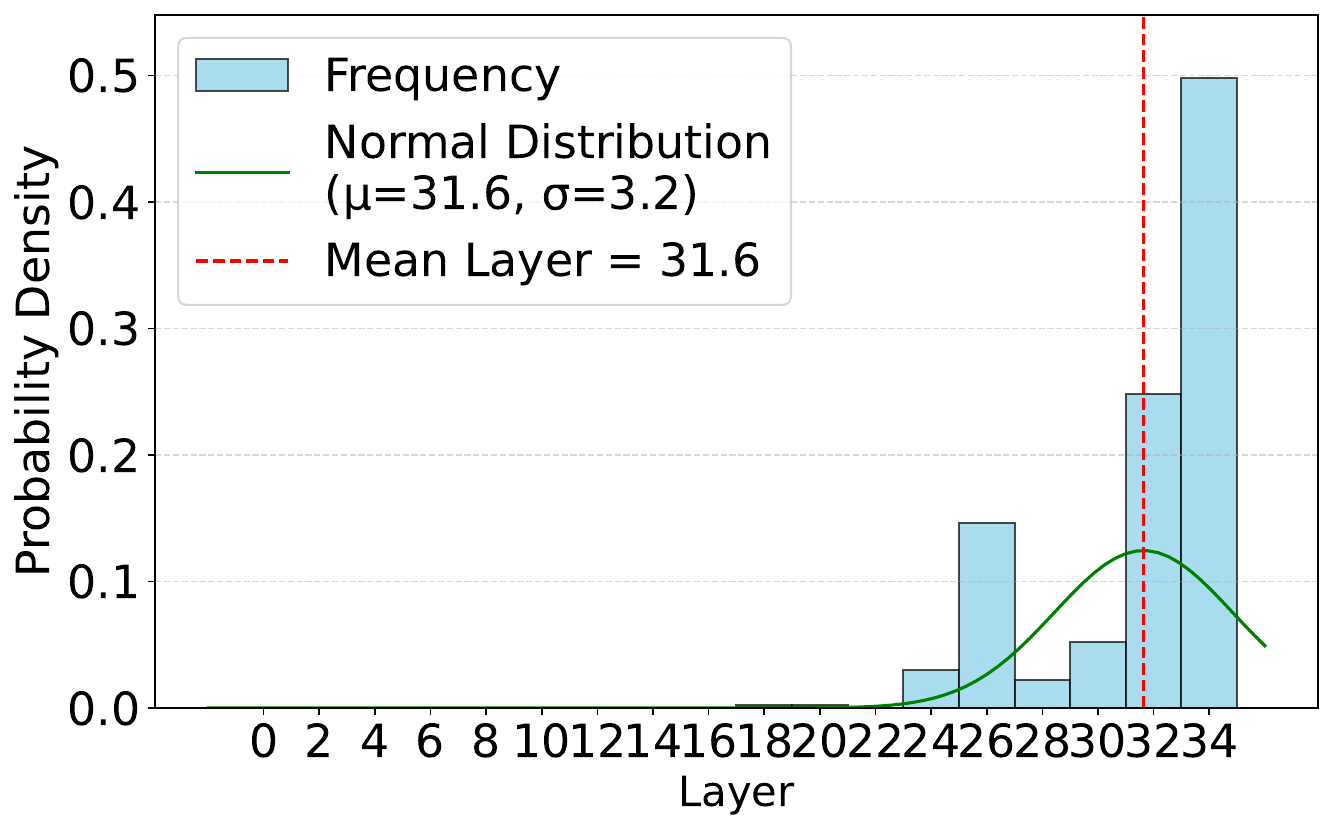}
        \caption{\small Histgram of layer selection in LFD on HotpotQA dataset.}
    \end{subfigure}
    
    \caption{Comparison between LFD and LFD (Fixed) using the Qwen3-8B model on the NQ and HotpotQA datasets.}
    \label{fig:fixed-qwen}
\end{figure*}

To further validate the effectiveness of the lowest IKS layer selection discussed in Section~\ref{ablation}, we conduct additional evaluations using models: Mistral-7B, DeepSeek-7B, and Qwen3-8B. Our experiments encompass both a single-hop QA dataset (NQ) and a multi-hop QA dataset (HotpotQA). The complete results are presented in Figures~\ref{fig:fixed-mistral}--\ref{fig:fixed-qwen}.

\section{Case Studies}\label{appendix-f}
We provide some case studies using the Llama2-7B model across four benchmark datasets: NQ, RGB, HotpotQA, and 2WikiMultihopQA. As shown in Tables~\ref{tab:case-nq}--\ref{tab:case-2wiki}, our method enables the model to better adhere to the provided context and correctly identify answers within the given information.

\begin{table}[h]
\vspace{-5pt}
    \caption{Case study on GD(0), CS, DoLA, and LFD on the NQ dataset using LLaMA2-7B.}
    \centering
    \scriptsize
    \begin{tabular}{c|p{2.5cm}|p{2.5cm}|p{2.5cm}|p{2.5cm}}
        \bottomrule
        Method & GD(0) & CS & DoLA & LFD \\
        \hline \hline
        \multirow{8}{*}{Prompt} & \multicolumn{4}{p{10.8cm}}{\parbox[t]{10.5cm}{
        You are given a question and you MUST respond by EXTRACTING the answer from one of the provided documents. If none of the documents contain the answer, respond with NO-RES ...
        \\
        Document [20971901](Title: Pirates of the Caribbean (film series)) The film series started in 2003 with Pirates of the Caribbean : {\color{blue}The Curse of the Black Pearl} ... The franchise 's second film , subtitled {\color{blue}Dead Man 's Chest} , was released three years later in 2006 ... The third film in the series , subtitled {\color{blue}At World 's End} , ... and Disney released a fourth film , subtitled {\color{blue}On Stranger Tides} , in 2011 ...
        \\
        Question: All the names of the pirate of the caribbean movies. Answer:
        }
        } \\
        \hline
        Ground Truth& \multicolumn{2}{p{5.5cm}}{[At World 's End, Dead Man 's Chest, On Stranger Tides]} \\
        \hline
        \multirow{5}{*}{Answer} & Captain Jack Sparrow, Will Turner, Elizabeth Swann, James Norrington, Barbossa, and Blackbeard. & Captain Jack Sparrow, Barbossa, Will Turner, Elizabeth Swann & Captain Jack Sparrow, Will Turner, Elizabeth Swann, James Norrington, Barbossa, and Blackbeard. & \textbf{Pirates of the Caribbean : The Curse of the Black Pearl, Dead Man's Chest, At World's End, and On Stranger Tides.} \\
        \toprule
    \end{tabular}
    \vspace{-14pt}
    \label{tab:case-nq}
\end{table}

\begin{table}[h]
\vspace{-5pt}
    \caption{Case study on GD(0), CS, DoLA, and LFD on the RGB dataset using LLaMA2-7B.}
    \centering
    \scriptsize
    \begin{tabular}{c|l|l|l|p{5.0cm}}
        \bottomrule
        Method & GD(0) & CS & DoLA & LFD \\
        \hline \hline
        \multirow{6}{*}{Prompt} & \multicolumn{4}{p{10.8cm}}{\parbox[t]{10.5cm}{
        
        You are given a question and you MUST respond by EXTRACTING the answer from one of the provided documents. If none of the documents contain the answer, respond with NO-RES. ... 
        \\
        Document: Riot Games Singapore ... In October 2022, Riot acquired {\color{blue}Wargaming Sydney}—a subsidiary of Cyprus-based Wargaming that had originally developed the MMO middleware BigWorld—for an undisclosed amount, and renamed it Riot Sydney ...
        \\
        Question: What gaming software development studio did Riot Games acquire? Answer:
        }
        } \\
        \hline
        Ground Truth& \multicolumn{2}{p{3cm}}{Wargaming Sydney} \\
        \hline
        \multirow{5}{*}{Answer} & \multirow{5}{*}{Riot Sydney} & \multirow{5}{*}{RIOT SYDNEY} & \multirow{5}{*}{Riot Sydney} & \textbf{Riot Games acquired Wargaming Sydney—a subsidiary of Cyprus-based Wargaming that had originally developed the MMO middleware BigWorld—for an undisclosed amount, and renamed it Riot Sydney.} \\
        \toprule
    \end{tabular}
    \vspace{-14pt}
    \label{tab:case-rgb}
\end{table}

\begin{table}[h]
\vspace{-5pt}
    \caption{Case study on GD(0), CS, DoLA, and LFD on the HotpotQA dataset using LLaMA2-7B.}
    \centering
    \scriptsize
    \begin{tabular}{c|l|l|l|p{5.0cm}}
        \bottomrule
        Method & GD(0) & CS & DoLA & LFD \\
        \hline \hline
        \multirow{11}{*}{Prompt} & \multicolumn{4}{p{10.8cm}}{\parbox[t]{10.5cm}{
        You are given a question and you MUST respond by EXTRACTING or DERIVING the answer from the provided documents. If the answer cannot be logically inferred from the documents, respond with NO-RES ...
        \\
        Document(Title: Terry Norris (actor)) ... As an actor, he has starred in TV Shows such as "Bellbird" \& "Cop Shop", and in films like {\color{blue}"Romulus, My Father"} and "Paper Planes" ...
        \\
        Document(Title: Romulus, My Father (film)) ...
        {\color{blue}Romulus, My Father} is a 2007 Australian drama film directed by Richard Roxburgh ... Based on the memoir by {\color{blue}Raimond Gaita}, the film tells the story of Romulus (Eric Bana) and his wife Christine (Franka Potente), and their struggle in the face of great adversity to raise their son, Raimond (Kodi Smit-McPhee) ...
        \\
        Question: Terence Richard "Terry" Norris starred in a 2007 Australian drama film that was directed by Richard Roxburgh, and was based on a memoir by who? Answer:
        }
        } \\
        \hline 
        Ground Truth& \multicolumn{2}{p{3cm}}{Raimond Gaita} \\
        \hline
        \multirow{4}{*}{Answer} & \multirow{4}{*}{ERIC BANA} & \multirow{4}{*}{ERIC BANA} & \multirow{4}{*}{NO-RES} & \textbf{Based on the documents provided, Terry Norris starred in the film "Romulus, My Father" directed by Richard Roxburgh, based on the memoir by Raimond Gaita.} \\
        \toprule
    \end{tabular}
    \vspace{-14pt}
    \label{tab:case-hotpot}
\end{table}

\begin{table}[h]
\vspace{-5pt}
    \caption{Case study on GD(0), CS, DoLA, and LFD on the 2WikiMultihopQA dataset using LLaMA2-7B.}
    \centering
    \scriptsize
    \begin{tabular}{c|p{2.5cm}|l|p{3cm}|p{3.5cm}}
        \bottomrule
        Method & GD(0) & CS & DoLA & LFD \\
        \hline \hline
        \multirow{8}{*}{Prompt} & \multicolumn{4}{p{10.8cm}}{\parbox[t]{10.5cm}{
        You are given a question and you MUST respond by EXTRACTING or DERIVING the answer from the provided documents. If the answer cannot be logically inferred from the documents, respond with NO-RES ...
        \\
        Document(Title: A Tale of Winter) A Tale of Winter is a 1992 French drama film directed by {\color{blue}Éric Rohmer}, and starring Charlotte Véry, Frédéric van den Driessche and Michael Voletti ...
        \\
        Document(Title: Éric Rohmer) ... He edited the influential film journal, {\color{blue}"Cahiers du cinéma"}, from 1957 to 1963 ...
        \\
        Question: Where does the director of film A Tale Of Winter work at? Answer:
        }
        } \\
        \hline 
        Ground Truth& \multicolumn{2}{p{3cm}}{Cahiers du cinéma} \\
        \hline
        \multirow{3}{*}{Answer} & Éric Rohmer works at a film production company. & \multirow{3}{*}{Éric Rohmer} & NO-RES: The answer cannot be logically inferred from the provided documents. & \textbf{Éric Rohmer works at a film journal called "Cahiers du cinéma".} \\
        \toprule
    \end{tabular}
    \vspace{-14pt}
    \label{tab:case-2wiki}
\end{table}

\section{Computational Details}\label{appendix-g}
All experiments are performed on a GPU-accelerated computing system equipped with NVIDIA GeForce RTX 3090 graphics processors (24GB GDDR6X VRAM each), supported by dual Intel Xeon Gold 6271C CPUs (2.6GHz base frequency, 48 cores total) and 251GB of system memory.
\newpage






\newpage

\end{document}